\documentclass[10pt,twocolumn,letterpaper]{article}

\usepackage[pagenumbers]{cvpr} % To force page numbers, e.g. for an arXiv version

\usepackage{graphicx}
\usepackage{amsmath}
\usepackage{amssymb}
\usepackage{booktabs}
\usepackage{multirow}

\usepackage{bbding}

\usepackage[ruled]{algorithm2e}
\usepackage{caption}
\usepackage[pagebackref,breaklinks,colorlinks]{hyperref}

\usepackage[capitalize]{cleveref}
\crefname{section}{Sec.}{Secs.}
\Crefname{section}{Section}{Sections}
\Crefname{table}{Table}{Tables}
\crefname{table}{Tab.}{Tabs.}

\def\cvprPaperID{5555} % *** Enter the CVPR Paper ID here
\def\confName{CVPR}
\def\confYear{2023}

\newcommand{\GDP}[1]{GDP #1}

\begin{document}

\title{Generative Diffusion Prior for Unified Image Restoration and Enhancement}

\author{Ben Fei$^{1,2,*}$, Zhaoyang Lyu$^{2,*}$, Liang Pan$^3$, Junzhe Zhang$^3$,\\ Weidong Yang$^{1,\dagger}$, Tianyue Luo$^1$, Bo Zhang$^2$, Bo Dai$^{2,\dagger}$\\
$^1$ Fudan University, $^2$Shanghai AI Laboratory, $^3$S-Lab, Nanyang Technological University\\
{\tt\small {bfei21@m.fudan.edu.cn,  wdyang@fudan.edu.cn, (lvzhaoyang,daibo)@pjlab.org.cn}}
% {\tt\small {(bfei21, tianyueluo21)@m.fudan.edu.cn,  wdyang@m.fudan.edu.cn, junzhe001@e.ntu.edu.sg}}\\
% {\tt\small liang.pan@ntu.edu.sg, bo.zhangzx@gmail.com, (lvzhaoyang,daibo)@pjlab.org.cn}
% \and
% Zhaoyang Lyu\\
}

%%% all sections
% !TEX root = ../PaperForReview.tex

\twocolumn[{%
\renewcommand\twocolumn[1][]{#1}%
\maketitle

\vspace{-1cm}

\begin{center}
    \centering
    \vspace{-8pt}
    \captionsetup{type=figure}
    \includegraphics[width=\textwidth]{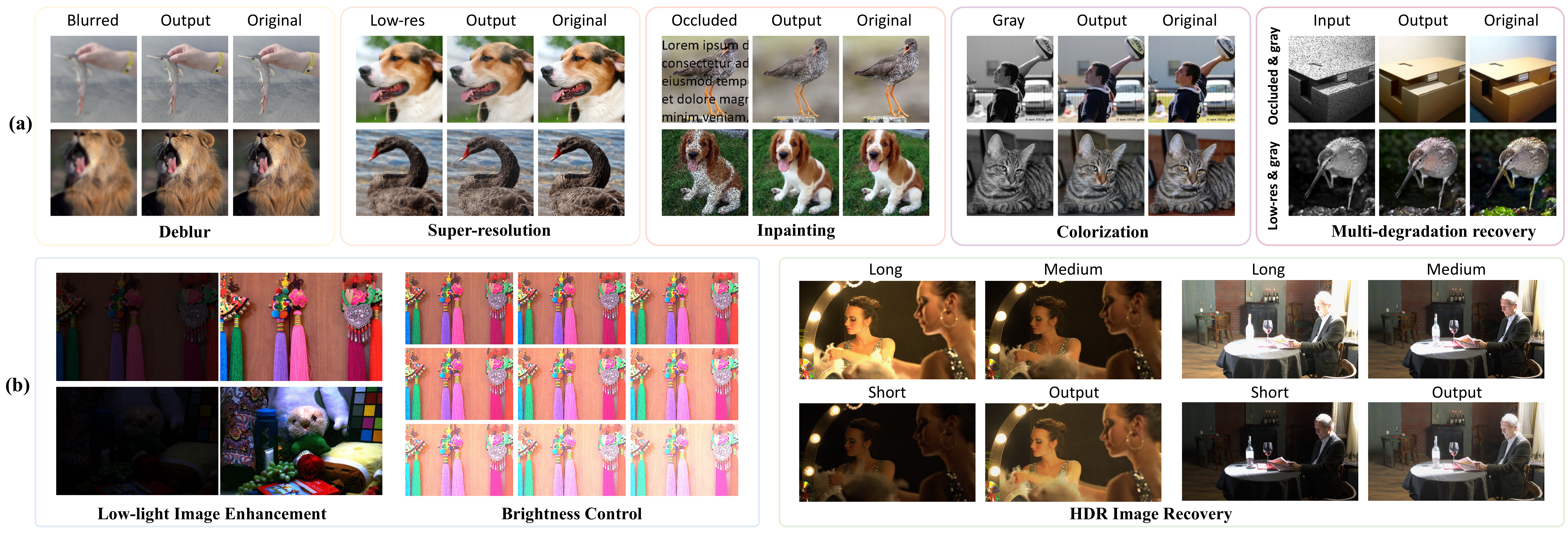}
    \vspace{-2em}
    \captionof{figure}{\textbf{Generative Diffusion Prior (GDP) is capable of generating high-fidelity restoration across various tasks.} \GDP gives  faithful image recovery on \textbf{(a) linear and multi-linear restoration.}. In addition, \GDP also enables novel applications of \textbf{(b) blind, non-linear, multiple-guidance, or any-size image}, including low-light enhancement and HDR recovery.}
    % \junzhe{I think (a) Linear ... and (b) Blind ... can be bold.}
    \label{teaser}
    \vspace{-1em}
\end{center}%
}]

\newcommand\blfootnote[1]{%
\begingroup
\renewcommand\thefootnote{}\footnote{#1}%
\addtocounter{footnote}{-1}%
\endgroup
}
\blfootnote{{$*$}Equal contribution, {$\dagger$}Corresponding author.}

%  without task-specific customization.\junzhe{To verify this statement, \ie, if we have any more changes other than additional losses?}
% !TEX root = ../PaperForReview.tex
\begin{abstract}
% \lyu{generative diffusion/denoising prior}.
   % The current \textbf{modus operandi} for tackling the image restoration problems leverages stochastic algorithms that sample from the posterior distribution of natural images given the measurements. 
   Existing image restoration methods mostly leverage the posterior distribution of natural images. 
   However, they often assume known degradation and also require supervised training, which restricts their adaptation to complex real applications.
%   However, efficient methods often require     problem-specific supervised training to model the posterior, while unsupervised solutions that are not problem-specific rely on inefficient iterative methods. 
   % This contribution copes with these challenges by presenting Generative Diffusion Prior (GDP), an efficient and unsupervised posterior sampling method. 
   In this work, we propose the Generative Diffusion Prior (GDP) to effectively model the posterior distributions in an unsupervised sampling manner.
   \GDP utilizes a pre-train denoising diffusion generative model (DDPM) for solving linear inverse, non-linear, or blind problems.
   Specifically, \GDP systematically explores a protocol of conditional guidance, which is verified more practical than the commonly used guidance way.
   Furthermore, \GDP is strength at optimizing the parameters of degradation model during the denoising process, achieving blind image restoration. 
%   Inspired by variational inference, \GDP fully utilizes a pre-trained denoising diffusion generative model (DDPM) for solving any linear inverse, non-linear, or blind problems. 
%   Specifically, a protocol of conditional guidance is systematically explored, which is proven more practical than the commonly used guidance way. 
%   And \GDP is strength at optimizing the parameters of the degradation model during the denoise process, achieving blind image restoration.
   Besides, we devise hierarchical guidance and patch-based methods, enabling the \GDP to generate images of arbitrary resolutions. 
   Experimentally, we demonstrate \GDP’s versatility on several image datasets for linear problems, such as super-resolution, deblurring, inpainting, and colorization, as well as non-linear and blind issues, such as low-light enhancement and HDR image recovery. 
   \GDP outperforms the current leading unsupervised methods on the diverse benchmarks in reconstruction quality and perceptual quality. 
   Moreover, \GDP also generalizes well for natural images or synthesized images with arbitrary sizes from various tasks out of the distribution of the ImageNet training set.
    The project page is available at \hyperlink{https://generativediffusionprior.github.io/}{https://generativediffusionprior.github.io/}
\end{abstract}
\vspace{-0.3cm}
% !TEX root = ../PaperForReview.tex

\section{Introduction}
\label{sec:intro}

% \begin{figure*}[t]
%   \centering
%   \includegraphics[width=\linewidth]{cvpr2023-author_kit-v1_1-1/Manuscript/teaser-1.pdf}
%   \caption{Guided diffusion models are capable to generate high-fidelity restoration across various tasks without task-specific customization.}
% \end{figure*}
% \lyu{In the sampling process of a pre-trained DDPM, the pre-trained DDPM usually first predicts a clean image $x^0$ from the noisy image $x_{t}$ by estimating the noise in $x_{t}$. Then we use the predicted $x^0$ together with $x_{t}$ to sample the next step latent $x_{t-1}$. We can add guidance on this intermediate variable $x^0$ to control the generation process of the DDPM.}
Image quality often degrades during capture, storage, transmission, and rendering. Image restoration and enhancement~\cite{liang2021swinir} aim to inverse the degradation and improve the image quality.
Typically, restoration and enhancement tasks can be divided into two main categories:
1) \textbf{Linear inverse problems}, such as image super-resolution (SR)~\cite{haris2018deep, ledig2017photo}, deblurring~\cite{kupyn2019deblurgan,suin2020spatially}, inpainting~\cite{yeh2017semantic}, colorization~\cite{larsson2016learning, zhang2016colorful}, where the degradation model is usually linear and known;
2) \textbf{Non-linear or blind problems}~\cite{asim2020blind}, such as image low-light enhancement~\cite{li2021low} and HDR image recovery~\cite{chen2021hdrunet,wang2021deep}, where the degradation model is non-linear and unknown.
% \junzhe{in fact, SR also got blind type, so we may make the statement more soft; double confirm the classifcation}
For a specific linear degradation model, image restoration can be tackled through end-to-end supervised training of neural networks~\cite{dong2015image,zhang2016colorful}.
Nonetheless, corrupted images in the real world often have multiple complex degradations~\cite{ongie2020deep}, where fully supervised approaches suffer to generalize.

There is a surge of interest to seek for more general image priors through generative models~\cite{shaham2019singan,gu2020image,asim2020blind}, and tackle image restoration in an unsupervised setting~\cite{chen2018image,el2022bigprior}, where multiple restoration tasks of different degradation models can be addressed during inference without re-training. For instance, Generative Adversarial Networks (GANs)~\cite{goodfellow2020generative} that are trained on a large dataset of clean images learn rich knowledge of the real-world scenes have succeeded in various linear inverse problems through GAN inversion~\cite{pan2021exploiting, menon2020pulse, gu2020image}. 
In parallel, Denoising Diffusion Probabilistic Models (DDPMs)~\cite{austin2021structured,cai2020learning,kingma2021variational,vahdat2021score,saharia2022palette,song2020denoising} have demonstrated impressive generative capabilities, level of details, and diversity on top of GAN~\cite{ho2020denoising,sohl2015deep,song2020improved,ulhaq2022efficient,ramesh2021zero,ramesh2022hierarchical}.
As an early attempt, Kawar~\etal~\cite{kawar2022denoising} explore pre-trained DDPMs with variational inference, and achieve satisfactory results on multiple restoration tasks, but their Denoising Diffusion Restoration Model (DDRM) leverages the singular value decomposition (SVD) on a known linear degradation matrix, making it still limited to linear inverse problems.

In this study, we take a step further and propose an efficient approach named Generative Diffusion Prior (GDP). It exploits a well-trained DDPM as effective prior for general-purpose image restoration and enhancement, using degraded image as guidance.
% and gradually and stochastically denoises samples to desired outputs\junzhe{can refine}, conditioned on the degraded images.
As a unified framework, \GDP not only works on various linear inverse problems, but also generalizes to non-linear, and blind image restoration and enhancement tasks for the first time.
% \junzhe{would need to concisely describe tech novelties, and merits with DDPM and the proposed framework.}
% Different from DDRM~\cite{kawar2022denoising}, our \GDP utilizes well-trained DDPMs as effective prior to recover images with competitive results on various linear inverse, 
% In this study, we take a step further to explore how DDPM priors can be used in a wider variety of restoration and enhancement tasks, including non-linear and blind inverse problems.
% Therefore, inspired by previous work \cite{r1,r6, r57}, we introduce an efficient approach named \GDP that can achieve competitive results on various linear inverse, non-linear, and blind problems. 
% Solving these problems is not trivial due to there is the main challenge that previous methods~\cite{mataev2019deepred,lempitsky2018deep,wang2021towards,yang2021gan,xia2022gan} can not simultaneously estimate the degradation models and recover images with high fidelity. 
% To this end, \GDP exploits 
However, solving the blind inverse problem is not trivial, as one would need to concurrently estimate the degradation model and recover the clean image with high fidelity. Thanks to the generative prior in a pre-trained DDPM, denoising within the DDPM manifold naturally regularizes the realness and fidelity of the recovered image. Therefore, we adopt a blind degradation estimation strategy, where the degradation model parameters of \GDP are randomly initialized and optimized during the denoising process.
% To this end, we exploit \GDP to solve this challenge\junzhe{GDP is our main framework and alr mentioned earlier, so I think this statement is not appropriate, should be GDP exploit sth else, like generative prior}, where the degradation model parameters of \GDP are randomly initialized and optimized during the denoising process. 
Moreover, to further improve the photorealism and image quality, we systematically investigate an effective way to guide the diffusion models. 
Specifically, in the sampling process, the pre-trained DDPM first predicts a clean image $\boldsymbol{\tilde{x}}_0$ from the noisy image $\boldsymbol{x}_{t}$ by estimating the noise in $\boldsymbol{x}_{t}$. 
We can add guidance on this intermediate variable $\boldsymbol{\tilde{x}}_0$ to control the generation process of the DDPMs.
In addition, with the help of the proposed hierarchical guidance and patch-based generation strategy, \GDP is able to recover images of arbitrary resolutions, where low-resolution images and degradation models are first predicted to guide the generation of high-resolution images.
%\junzhe{might need to slightly elaborate how it is possible.}. 
% Then we use the predicted $\boldsymbol{\hat{x}}_0$ together with $\boldsymbol{x}_{t}$ to sample the next step latent $\boldsymbol{x}_{t-1}$. 

We demonstrate the empirical effectiveness of \GDP by comparing it with various competitive unsupervised methods under the linear or multi-linear inverse problem on ImageNet~\cite{deng2009imagenet}, LSUN~\cite{yu2015lsun}, and CelebA~\cite{karras2017progressive} datasets in terms of consistency and FID. 
Over the low-light~\cite{li2021low} and NTIRE~\cite{perez2021ntire} datasets, we further show \GDP results on non-linear and blind issues, including low-light enhancement and HDR recovery, superior to other zero-shot baselines both qualitatively and quantitively, manifesting that \GDP trained on ImageNet also works on images out of its training set distribution.

% \lyu{find a place in intro to emphasize that we use a single unconditional DDPM pretrained on ImageNet provide by ... in all of our experiments.}
Our contributions are fourfold: 
% Our main \textbf{contributions}: can be summarized as follows: 
(1) 
% We introduce GDP, an effective and unsupervised posterior sampling method. 
To our best knowledge, \GDP is the first unified problem solver that can effectively use \textbf{a single unconditional DDPM pre-trained on ImageNet} provide by~\cite{dhariwal2021diffusion} to produce diverse and high-fidelity outputs for unified image restoration and enhancement in an unsupervised manner. 
% ~\junzhe{I guess this is Chinglish} 
(2) GDP is capable of optimizing randomly initiated parameters of degradation that are unknown, resulting in a powerful framework that can tackle any blind image restoration. (3) Further, to achieve arbitrary size image generation, we propose hierarchical guidance and patch-based methods, greatly promoting \GDP on natural image enhancement. (4) Moreover, the comprehensive experiments are carried out, different from the conventional guidance way, where \GDP directly predicts the temporary output given the noisy image in every step, which will be leveraged to guide the generation of images in the next step. 

% \begin{itemize}
% \setlength{\itemsep}{0pt}
% \setlength{\parsep}{0pt}
% \setlength{\parskip}{0pt}
%     \item 
%     \item 
%     \item 
%     \item 
% \end{itemize}

%teaser 修改，细化我们方法的优势在intro。画一张figure2总结的。teaser可以加入光照调整的实验。
%-------------------------------------------------------------------------
% !TEX root = ../PaperForReview.tex

\section{Related works}
\label{sec:Related works}

% !TEX root = ../PaperForReview.tex

\begin{figure*}[t]
   \centering
   % \setlength{\abovecaptionskip}{0.15cm}
   % \vspace{-6pt}
   % \includegraphics[width=\linewidth]{cvpr2023-author_kit-v1_1-1/Manuscript/figure-overview-7.pdf}
   \includegraphics[width=\linewidth]{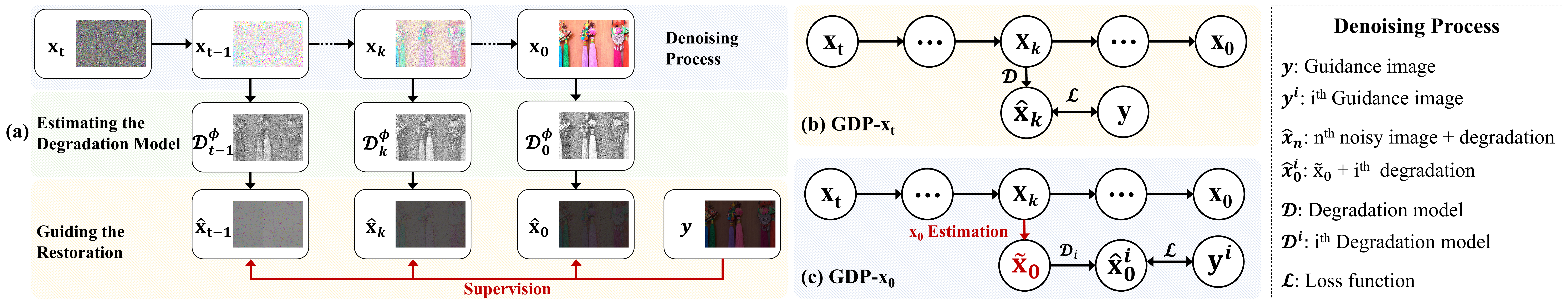}
   \vspace{-20pt}
   \caption{
   % \lyu{change ``learning'' to ``estimating''. Do not use $x_n$, since $n$ is used as the number of reference images.} 
   \textbf{Overview of our \GDP for unified image recovery.} (a) Given a corrupted image $\boldsymbol{y}$ during inference, \GDP systematically studies the reverse process from $\boldsymbol{x}_T$ to $\boldsymbol{x}_0$ guided by the $\boldsymbol{y}$. The guidance can be added on $\boldsymbol{\hat{x}}_0$ or $\boldsymbol{\hat{x}}_t$, leading to two variants of GDP. And $\boldsymbol{\hat{x}}_0$ and $\boldsymbol{\hat{x}}_t$ can be collectively expressed as $\boldsymbol{\hat{x}_t}$. The supervision signal (Sec. \ref{sec:loss}) is applied between $\boldsymbol{\hat{x}_t}$ and $\boldsymbol{y}$. \GDP looks for an intermediate variable $\boldsymbol{x}_t$ and optimizes the degradation model $\{\mathcal{D}^i_\phi \mid i = 1, 2, \ldots, n\}$ that best reconstruct the image corresponding to $\boldsymbol{y}$ via gradient descent. Note that \GDP is a generic image restoration method. We illustrate it with the low light enhancement example. 
   (b) GDP-$\boldsymbol{x}_t$ adds guidance $\boldsymbol{x}_t$ in every time step (Algo. ~\ref{algo5}), while (c) GDP-$\boldsymbol{x}_0$ estimates the $\boldsymbol{\tilde{x}}_0$ given $\boldsymbol{x}_t$, then adds guidance on $\boldsymbol{\tilde{x}}_0$ to obtain $\boldsymbol{\hat{x}}_0$(Algo. ~\ref{algo2}). The number of guidance images $\{\boldsymbol{y}^i \mid i = 1, 2, \ldots, n\}$ and the degradation models $\{\mathcal{D}^i_\phi \mid i = 1, 2, \ldots, n\}$ are dependent on the tasks. For instance, $n = 3$ for HDR recovery, while $n = 1$ for other tasks.}
\vspace{-1.5em}
\label{fig:comparison}
\end{figure*}

% \begin{equation}
%     \mathbf{y}=\boldsymbol{H} \mathbf{x}+\mathbf{z},
% \label{eq1}
% \end{equation}

% A general linear inverse problem is posed as $\mathbf{y}=\boldsymbol{H} \mathbf{x}+\mathbf{z}$, where the target is to recover the signal $\mathbf{x} \in \mathbb{R}^{n}$ from measurements $\mathbf{y} \in \mathbb{R}^{m}$, as well as $\boldsymbol{H} \in \mathbb{R}^{m \times n}$ is a known linear degradation matrix, and $\mathbf{z} \sim \mathcal{N}\left(0, \sigma_{\mathbf{y}}^{2} \boldsymbol{I}\right)$ is an i.i.d. additive Gaussian noise with known variance. The underlying structure of $\mathbf{x}$ can be represented via a generative model, denoted as $p_{\theta}(\mathbf{x})$. Given $\mathbf{y}$ and $\boldsymbol{H}$, a posterior over the signal can be posed as: $p_{\theta}(\mathbf{x}|\mathbf{y}) \propto p_{\theta}(\mathbf{x}) p(\mathbf{y}|\mathbf{x})$, where the "likelihood" term $p(\mathbf{y}|\mathbf{x})$ is defined. Recovering $\mathbf{x}$ can be done by sampling from this posterior, which may require many iterations to produce a good sample. Alternatively, one can also approximate this posterior by learning a model via amortized inference (i.e., supervised learning); the model learns to predict $\mathbf{x}$ given $\mathbf{y}$, generated from $\mathbf{x}$ and a specific $\boldsymbol{H}$. While this can be more efficient than sampling-based methods, it may generalize poorly to inverse problems that have not been trained on. 

\noindent\textbf{Linear Inverse Image Restoration.} Most diffusion models toward linear inverse problems have employed unconditional models for the conditional tasks\cite{meng2021sdedit,song2020score}, where only one model needs to be trained. However, unconditional tasks tend to be more difficult than conditional tasks. Moreover, the multi-linear task is also a relatively under-explored subject in image restoration. For instance, \cite{qian2019trinity, yu2018crafting} train simultaneously on multiple tasks, but they mainly concentrate on the enhancement tasks like deblurring and so on. Some works have also handled the multi-scale super-resolution by simultaneous training over multiple degradations \cite{kim2016deeply}. Here, we propose \GDP as a single model for dealing with single linear inverse or multiple linear inverse tasks. 

% With \GDP, we take a first step toward building an image restoration diffusion model for unified linear inverse and multi-linear tasks.

% The non-linear image degradation model for a spatially invariant system is generally formulated as $\mathbf{y}=s\left(\boldsymbol{H} \mathbf{x}\right)+\mathbf{z}$. Here the true image $\mathbf{x} \in \mathbb{R}^{n}$ is corrupted by a component-wise non-linearity $s(\cdot)$.

\noindent\textbf{Non-linear Image Restoration.} The non-linear image formation model provides an accurate description of several imaging systems, including camera response functions in high-dynamic-range imaging\cite{Rameshan2011high}. The non-linear image restoration model is more accurate but is often more computationally intractable. Recently, great attention has been paid to non-linear image restoration problems. For example, HDR-GAN~\cite{niu2021hdr} is proposed for synthesizing HDR images from multi-exposed LDR images, while EnlightenGAN \cite{jiang2021enlightengan} is devised as an unsupervised GAN to generalize very well on various real-world test images. The diffusion models are rarely studied for non-linear image restoration.

% Blind image restoration handles this issue by assuming that the degradation matrix $\boldsymbol{H}$ and the additive Gaussian noise $\mathbf{z}$ are unknown.

\noindent\textbf{Blind Image Restoration.} Early supervised attempts~\cite{begin2004blind,he2009soft} tend to estimate the unknown point spread function. As an example,~\cite{ke2021unsupervised} designs a class of structured denoisers, and ~\cite{shocher2018zero} employs a fixed down-sampling operation to generate synthetic pairs during testing. However, these methods often are incapable of obtaining the parameters or distribution of the observed data due to the complicated degradation types. Another way to solve the blind image restoration is to utilize unsupervised learning methods~\cite{zhu2017unpaired,du2020learning}. Following the CycleGAN~\cite{zhu2017unpaired}, CinCGAN \cite{yuan2018unsupervised} and MCinCGAN \cite{zhang2019multiple} employ a pre-trained SR model together with cycle consistency loss to learn a mapping from the input image to high-quality image space. However, it still remains a challenge to exploit a unified architecture for blind image restoration. By the merits of the powerful GDP, these blind problems could also be solved by simultaneously estimating the recovered image and a specific degradation model.

% !TEX root = ../PaperForReview.tex

\section{Preliminary}

% \input{cvpr2023-author_kit-v1_1-1/latex/figures/overview.tex}

% \junzhe{some warm-up needed, or a brief intro of this section in front; besides, the motivation of using DDPM should also be discussed}

% The diffusion process gradually corrupts a clean image to Gaussian noise, while the reverse process iteratively denoises a sampled Gaussian noise to a clean image.

% We formulate $\boldsymbol{x}_0 \sim p_{\text {data}}$ is the distribution of all images $\boldsymbol{x}$ in the dataset, while $p_{\text {latent }}=\mathcal{N}\left(\mathbf{0}, \boldsymbol{I}\right)$ stands for the latent distribution and $\mathcal{N}$ is the Gaussian distribution. 

% \lyu{also explain $N$. Explain every notation when they first appear.}

Diffusion models~\cite{ryu2022pyramidal,wu2022guided,gu2022vector,austin2021structured,kingma2021variational} transform complex data distribution $\boldsymbol{x}_0 \sim p_{\text {data}}$ into simple noise distribution $\boldsymbol{x}_T \sim p_{\text {latent }}=\mathcal{N}\left(\mathbf{0}, \boldsymbol{I}\right)$ and recover data from noise, where $\mathcal{N}$ is the Gaussian distribution. DDPMs mainly comprise the diffusion process and the reverse process.

% The diffusion process from clean data $\boldsymbol{x}_0$ to $\boldsymbol{x}_T$ is written as:

% \lyu{always use subscript or superscript. be consistent!}

\noindent\textit{\underline{The Diffusion Process}} is a Markov chain that gradually corrupts data $\boldsymbol{x}_0$ until it approaches Gaussian noise $p_{\text {latent }}$ at $T$ diffusion time steps. Corrupted data $\boldsymbol{x}_1, \ldots, \boldsymbol{x}_T$ are sampled from data $p_{\text {data}}$, with a diffusion process, which is defined as Gaussian transition:
\begin{equation}
\vspace{-0.1cm}
     \!\!q\left(\boldsymbol{x}_1, \cdots, \boldsymbol{x}_T \mid \boldsymbol{x}_0\right)=\prod_{t=1}^T q\left(\boldsymbol{x}_t \mid \boldsymbol{x}_{t-1}\right), 
\vspace{-0.1cm}
\end{equation}
where $t$ denotes as diffusion step, $q\left(\boldsymbol{x}_t \mid \boldsymbol{x}_{t-1}\right)=\mathcal{N}\left(\boldsymbol{x}_t; \sqrt{1-\beta_t} \boldsymbol{x}_{t-1}, \beta_t \boldsymbol{I}\right)$, and $\beta_t$ are fixed or learned variance schedule. 
An important property of the forward noising process is that any step $\boldsymbol{x}_t$ may be sampled directly from $\boldsymbol{x}_0$ through the following equation:
\vspace{-0.2cm}
% Noisy data $\boldsymbol{x}_t$ can be sampled from $\boldsymbol{x}_0$ directly through the following equation:
\begin{equation}
\begin{split}
% & 
& \boldsymbol{x}_t=\sqrt{\bar{\alpha}_t} \boldsymbol{x}_0+\sqrt{1-\bar{\alpha}_t} \boldsymbol{\epsilon}, 
% \\ & \text { where } \boldsymbol{\epsilon} \text { is a standard Gaussian noise. }
\end{split}
\label{eq2}
\end{equation}
\vspace{-0.2cm}
where $\boldsymbol{\epsilon} \sim \mathcal{N}(0, \boldsymbol{I})$, $\alpha_t=1-\beta_t$ and $\bar{\alpha}_t=\prod_{i=1}^t \alpha_i$. Proved by Ho~\etal~\cite{ho2020denoising}, there is a closed form expression for $q\left(\boldsymbol{x}_t \mid \boldsymbol{x}_0\right)$. We can obtain $q\left(\boldsymbol{x}_t \mid \boldsymbol{x}_0\right)=$ $\mathcal{N}\left(\boldsymbol{x}_t ; \sqrt{\bar{\alpha}_t} \boldsymbol{x}_0,\left(1-\bar{\alpha}_t\right) \boldsymbol{I}\right)$.
Herein, $\bar{\alpha}_t$ goes to 0 with large $T$, and $q\left(\boldsymbol{x}_T \mid \boldsymbol{x}_0\right)$ is close to the latent distribution $p_{\text {latent }}$. 

% where $\boldsymbol{\epsilon} \sim \mathcal{N}(0, \mathbf{I})$.
% where $\boldsymbol{\epsilon}$ is a standard Gaussian noise, which could naturally provide a novel way to add guidance. 
% \lyu{rewrite these two paragraphs. change the words you use. This is too close to my PDR paper.
% }

\noindent\textit{\underline{The Reverse Process}} is a Markov chain that iteratively denoises a sampled Gaussian noise to a clean image. Starting from noise $x_T \sim \mathcal{N}(0, \boldsymbol{I})$, the reverse process from latent $\boldsymbol{x}_T$ to clean data $\boldsymbol{x}_0$ is defined as:
\vspace{-0.1cm}
\begin{equation}
\begin{split}
& p_{\boldsymbol{\theta}}\left(\boldsymbol{x}_0, \cdots, \boldsymbol{x}_{T-1} \mid \boldsymbol{x}_T\right)=\prod_{t=1}^T p_{\boldsymbol{\theta}}\left(\boldsymbol{x}_{t-1} \mid \boldsymbol{x}_t\right),\quad \\ &
p_{\boldsymbol{\theta}}\left(\boldsymbol{x}_{t-1} \mid \boldsymbol{x}_t\right)=\mathcal{N}\left(\boldsymbol{x}_{t-1} ; \boldsymbol{\mu}_{\boldsymbol{\theta}}\left(\boldsymbol{x}_t, t\right), \Sigma_{\theta} \boldsymbol{I}\right)
% \\ & \text { where } p_{\boldsymbol{\theta}}\left(\boldsymbol{x}_{t-1} \mid \boldsymbol{x}_t\right)=\mathcal{N}\left(\boldsymbol{x}_{t-1} ; \boldsymbol{\mu}_{\boldsymbol{\theta}}\left(\boldsymbol{x}_t, t\right), \sigma_t^2 \boldsymbol{I}\right).
\end{split}
\label{eq:reverse}
\end{equation}
\vspace{-0.1cm}
% where $p_{\boldsymbol{\theta}}\left(\boldsymbol{x}_{t-1} \mid \boldsymbol{x}_t\right)=\mathcal{N}\left(\boldsymbol{x}_{t-1} ; \boldsymbol{\mu}_{\boldsymbol{\theta}}\left(\boldsymbol{x}_t, t\right), \sigma_t^2 \boldsymbol{I}\right)$.
According to Ho~\etal~\cite{ho2020denoising}, the mean $\boldsymbol{\mu}_{\boldsymbol{\theta}}\left(\boldsymbol{x}_t, t\right)$ is the target we want to estimate by a neural network $\boldsymbol{\theta}$. The variance $\Sigma_{\theta}$ can be either time-dependent constants~\cite{ho2020denoising} or learnable parameters~\cite{nichol2021improved}.
$\epsilon_{\theta}$ is a function approximator intended to predict $\epsilon$ from $\boldsymbol{x}_t$ as follow:
\vspace{-0.2cm}
\begin{equation}
    \boldsymbol{\mu}_\theta\left(\boldsymbol{x}_t, t\right)=\frac{1}{\sqrt{\alpha_t}}\left(\boldsymbol{x}_t-\frac{\beta_t}{\sqrt{1-\bar{\alpha}_t}} \boldsymbol{\epsilon}_\theta\left(\boldsymbol{x}_t, t\right)\right)
\end{equation}
\vspace{-0.2cm}
% \lyu{add explanation for $\epsilon_{\theta}$}

In practice, $\boldsymbol{\Tilde{x}}_0$ is usually predicted from $\boldsymbol{x}_t$, then  $\boldsymbol{x}_{t-1}$ is sampled using both $\boldsymbol{\Tilde{x}}_0$ and $\boldsymbol{x}_t$ computed as:
\vspace{-0.2cm}
\begin{equation}
\!\!\!\!  \boldsymbol{\tilde{x}}_{0} =  \frac{\boldsymbol{x}_{t}}{\sqrt{\bar{\alpha}_{t}}}-\frac{\sqrt{1-\bar{\alpha}_{t}} \epsilon_{\theta}\left(\boldsymbol{x}_{t}, t\right)}{\sqrt{\bar{\alpha}_{t}}}
\end{equation}
\vspace{-0.2cm}
\begin{equation}
\begin{split}
\!\!\!\!  q\left(\boldsymbol{x}_{t-1} \mid \boldsymbol{x}_t, \boldsymbol{\Tilde{x}}_0\right) & =\mathcal{N}\left(\boldsymbol{x}_{t-1} ; \tilde{\boldsymbol{\mu}}_t\left(\boldsymbol{x}_t, \boldsymbol{\Tilde{x}}_0\right), \tilde{\beta}_t \mathbf{I}\right), \\
\!\!\!\!  \text { where } \quad \tilde{\boldsymbol{\mu}}_t\left(\boldsymbol{x}_t, \boldsymbol{\Tilde{x}}_0\right) & =\frac{\sqrt{\bar{\alpha}_{t-1}} \beta_t}{1-\bar{\alpha}_t} \boldsymbol{\Tilde{x}}_0+\frac{\sqrt{\alpha_t}\left(1-\bar{\alpha}_{t-1}\right)}{1-\bar{\alpha}_t} \boldsymbol{x}_t \quad \\
\!\!\!\!  & \text { and } \quad \tilde{\beta}_t=\frac{1-\bar{\alpha}_{t-1}}{1-\bar{\alpha}_t} \beta_t
\end{split}
\label{eq4}
\end{equation}
\vspace{-0.5cm}

\begin{table}[t]\footnotesize
\centering
\tabcolsep=0.05cm
\caption{\textbf{Comparison of different generative priors and regularization priors for image restoration and enhancement.}}
\vspace{-8pt}
\scalebox{0.9}
{%
\begin{tabular}{lccccc}
\toprule[1.5pt]
Methods           & DGP~\cite{pan2021exploiting} & SNIPS~\cite{kawar2021snips}                                                             & RED~\cite{romano2017little}                                                                                & DDRM~\cite{kawar2022denoising} & GDP (Ours)  \\ \midrule[1pt]
Prior      & GAN & \begin{tabular}[c]{@{}c@{}}MMSE \\ Gaussian \\ denoiser\end{tabular} & \begin{tabular}[c]{@{}c@{}}Laplacian-based \\ regularization \\ function\end{tabular} & DDPM & DDPM \\ \midrule[1pt]
Linear     & \CheckmarkBold   & \CheckmarkBold                                                                 & \CheckmarkBold                                                                                  & \CheckmarkBold    & \CheckmarkBold    \\
Non-linear & \XSolidBrush   & \XSolidBrush                                                                 & \XSolidBrush                                                                                  & \XSolidBrush    & \CheckmarkBold    \\
Blind      & \XSolidBrush   & \XSolidBrush                                                                 & \XSolidBrush                                                                                  & \XSolidBrush    & \CheckmarkBold    \\ \bottomrule[1.5pt]
\end{tabular}%
}
\vspace{-1.8em}
\label{methods-comparison}
\end{table}
% !TEX root = ../PaperForReview.tex

\section{Generative Diffusion Prior}
\label{sec: Generative Diffusion Prior}

% \input{cvpr2023-author_kit-v1_1-1/latex/figures/xt_x0.tex}

% In this work, we mainly focus on unified image restoration, including linear inverse, non-linear, or blind problems. On the one hand, these problem solvers based on posterior sampling often face a dilemma: supervised ones are efficient but can only address specific problems, whereas unsupervised approaches apply to general problems but are inefficient. To tackle the dilemma, we introduce \GDP, an unsupervised solver for general problems capable of handling such tasks in the measurements. \GDP is efficient and exhibits competitive performance compared to popular unsupervised solvers. On the other hand, there is merely a robust network to tackle these problems comprehensively, but our \GDP can be theory applied to any linear inverse, non-linear, and blind problems.

In this study, we aim to exploit a well-trained DDPM as an effective prior for unified image restoration and enhancement, in particular, to handle degraded images of a wide range of varieties. 
% Despite the notable success of GAN in various image restoration and manipulation tasks, it has been under-explored for DDPM in unified image restoration. 
% As a commonly used practice, learning a mapping from $\mathbf{y}$ to $\mathbf{x}$ needs task-specific training for different $\phi$ s. Alternatively,
% \lyu{explain that $y$ is the degraded image that we want to restore.}
In detail, assume degraded image $\boldsymbol{y}$ is captured via $\boldsymbol{y}=\mathcal{D}(\boldsymbol{x})$, where $\boldsymbol{x}$ is the original natural image, and $\mathcal{D}$ is a degradation model. We employ statistics of $\boldsymbol{x}$ stored in some prior and search in the space of $\boldsymbol{x}$ for an optimal $\boldsymbol{x}$ that best matches $\boldsymbol{y}$, regarding $\boldsymbol{y}$ as corrupted observations of $\boldsymbol{x}$. Due to the limited GAN inversion performance and the restricted applications of previous works~\cite{kawar2021snips,pan2021exploiting,romano2017little,kawar2022denoising} in Table~\ref{methods-comparison}, in this paper, we focus on studying a more generic image prior,~\textit{i.e.}, the diffusion models trained on large-scale natural images for image synthesis. Inspired by the~\cite{chen2020wavegrad, saharia2022image, rombach2022high, choi2022perception, batzolis2021conditional}, the reverse denoising process of the DDPM can be conditioned on the degraded image $\boldsymbol{y}$. Specifically, the reverse denoising distribution $p_{\boldsymbol{\theta}}(\boldsymbol{x}_{t-1}|\boldsymbol{x}_{t})$ in Eq. \ref{eq:reverse} is adopted to a conditional distribution $p_{\boldsymbol{\theta}}\left(\boldsymbol{x}_{t-1} |\boldsymbol{x}_t, \boldsymbol{y}\right)$.  \cite{sohl2015deep, dhariwal2021diffusion} prove that
\vspace{-0.1cm}
\begin{equation}\small
\begin{split}
    \!\!\!\! \log p_{\boldsymbol{\theta}}\left(\boldsymbol{x}_{t-1}|\boldsymbol{x}_t, \boldsymbol{y}\right) &=\log \left(p_{\boldsymbol{\theta}}\left(\boldsymbol{x}_{t-1}|\boldsymbol{x}_t\right) p\left(\boldsymbol{y}| \boldsymbol{x}_t\right)\right)+K_1 
    \\ \!\!\!\!  & \approx \log p(\boldsymbol{r})+K_2, 
    % \\ & \text{where }  \boldsymbol{z} \sim \mathcal{N}\left(\boldsymbol{z} ; \boldsymbol{\mu}_{\boldsymbol{\theta}}\left(\boldsymbol{x}_t, t\right)+\sigma_t^2 \boldsymbol{g}, \sigma_t^2 \boldsymbol{I}\right), \\ & \boldsymbol{g}=\nabla_{\boldsymbol{x}_t} \log p\left(\boldsymbol{y} \mid \boldsymbol{x}_t\right), C_1 \text{ and } C_2 \text{ are constants.}
\end{split}
\end{equation}
% \vspace{-0.1cm}
where $\boldsymbol{r} \sim \mathcal{N}\left(\boldsymbol{r} ; \boldsymbol{\mu}_{\boldsymbol{\theta}}\left(\boldsymbol{x}_t, t\right)+\Sigma \boldsymbol{g}, \Sigma \right)$ and $\boldsymbol{g}=\nabla_{\boldsymbol{x}_t} \log p\left(\boldsymbol{y} \mid \boldsymbol{x}_t\right)$, where $\Sigma = \Sigma_{\theta}\left(\boldsymbol{x}_{t}\right)$ for conciseness. $K_1$ and $K_2$ are constants,
% In the above equation, 
$p_{\boldsymbol{\theta}}\left(\boldsymbol{x}_{t-1} \mid \boldsymbol{x}_t\right)$ is defined by Eq. \ref{eq:reverse}. 
% \lyu{do not add ``the'' before variables. fix other places as well.} 
$p\left(\boldsymbol{y} \mid \boldsymbol{x}_t\right)$ can be regarded as the probability that $\boldsymbol{x}_t$ will be denoised to a high-quality image consistent to $\boldsymbol{y}$. 
We propose a heuristic approximation of it:
% \begin{equation}
% \begin{split}
%     p\left(\boldsymbol{y} \mid \boldsymbol{x}_t\right) = \frac{1}{Z} & \exp \left(-s \left(\mathcal{L}\left(\boldsymbol{x}_t, \boldsymbol{y}\right) + \lambda \mathcal{Q}\left(\boldsymbol{x}_t\right) \right)\right)
%     \\ =\frac{1}{Z} & \exp( -s (\mathcal{L}\left(\sqrt{\bar{\alpha}_t} \boldsymbol{x}_0+\sqrt{1-\bar{\alpha}_t} \boldsymbol{\epsilon}, \boldsymbol{y}\right))
%     \\ + \,& \lambda \mathcal{Q}\left(\sqrt{\bar{\alpha}_t} \boldsymbol{x}_0+\sqrt{1-\bar{\alpha}_t} \boldsymbol{\epsilon}\right)), 
%     % \\ & \text { where } \mathcal{L} \text { is distance metric and } \\ & \mathcal{Q} \text { is the optional quality enhancement metric, }
% \end{split}
% \end{equation}
\vspace{-0.1cm}
\begin{equation}
\begin{split}
    p\left(\boldsymbol{y} \mid \boldsymbol{x}_t\right) = \frac{1}{Z} & \exp \left(-\left[s\mathcal{L}\left(\mathcal{D}(\boldsymbol{x}_t), \boldsymbol{y}\right) + \lambda \mathcal{Q}(\boldsymbol{x}_t) \right]\right)
\end{split}
\end{equation}
\vspace{-0.1cm}
where $\mathcal{L}$ is some image distance metric, $Z$ is a normalization factor, and $s$ is a scaling factor controlling the magnitude of guidance. 
Intuitively, this definition encourages $\boldsymbol{x}_t$ to be consistent with the corrupted image $\boldsymbol{y}$ to obtain a high probability of $p\left(\boldsymbol{y} \mid \boldsymbol{x}_t\right)$. 
% \lyu{$\mathcal{Q}\left(\mathcal{D}(\boldsymbol{x}_t)\right)$ or $\mathcal{Q}(\boldsymbol{x}_t)$}.
% \lyu{do not add ``the'' before variables}
$\mathcal{Q}$ is the optional quality enhancement loss to enhance the flexibility of GDP, which can be used to control some properties (such as brightness) or enhance the quality of the denoised image. 
$\lambda$ is the scale factor for adjusting the quality of images.
% \junzhe{attention}
% In this formulation, we requires $\mathcal{L}\left(\boldsymbol{x}_t, \boldsymbol{y}\right)$ and $\mathcal{Q}\left(\boldsymbol{x}_t, \boldsymbol{y}\right)$ to be small to increase $p\left(\boldsymbol{y} \mid \boldsymbol{x}_t\right)$, encouraging the reconstructed image to be consistent to the guidance image $\boldsymbol{y}$ in the reverse process. 
The gradients of both sides are computed as:
\vspace{-0.1cm}
\begin{equation}\small
    \begin{aligned}
    \!\!\!\! &\log p\left(\boldsymbol{y} \mid \boldsymbol{x}_t\right)=-\log Z-s \mathcal{L}\left(\mathcal{D}(\boldsymbol{x}_t), \boldsymbol{y}\right) - \lambda \mathcal{Q}\left(\boldsymbol{x}_t\right) \\
    \!\!\!\! &\nabla_{\boldsymbol{x}_t} \log p\left(\boldsymbol{y} \mid \boldsymbol{x}_t\right)=-s \nabla_{\boldsymbol{x}_t} \mathcal{L}\left(\mathcal{D}(\boldsymbol{x}_t), \boldsymbol{y}\right) - \lambda \nabla_{\boldsymbol{x}_t} \mathcal{Q}\left(\boldsymbol{x}_t\right).
    \end{aligned}
\end{equation}
\vspace{-0.1cm}
where the distance metric $\mathcal{L}$ and the optional quality loss $\mathcal{Q}$ can be found in Sec. \ref{sec:loss}.

% !TEX root = ../PaperForReview.tex

\begin{algorithm}[t]\footnotesize
	\caption{\textbf{GDP-}$\boldsymbol{x}_t$ with fixed degradation model: Conditioner guided diffusion sampling on $\boldsymbol{x}_{t}$, given a diffusion model $\left(\mu_{\theta}\left(\boldsymbol{x}_{t}\right), \Sigma_{\theta}\left(\boldsymbol{x}_{t}\right)\right)$, corrupted image conditioner $\boldsymbol{y}$. 
 % \lyu{why alg 1 has known degradation and alg2 has unknown.}
 }
	\KwIn{Corrupted image $\boldsymbol{y}$, gradient scale $s$, degradation model $\mathcal{D}$, distance measure $\mathcal{L}$, optional quality enhancement loss $\mathcal{Q}$, quality enhancement scale $\lambda$.}
	\KwOut{Output image $\boldsymbol{x}_{0}$ conditioned on $\boldsymbol{y}$}
        Sample $\boldsymbol{x}_{T}$ from $\mathcal{N}(0, \mathbf{I})$
        
	\For{$t$ from $T$ to 1}{
	    $\mu, \Sigma = \mu_{\theta}\left(\boldsymbol{x}_{t}\right), \Sigma_{\theta}\left(\boldsymbol{x}_{t}\right)$
        
        $\mathcal{L}^{total}_{ \boldsymbol{{x}}_{t}} = \mathcal{L}(\boldsymbol{y}, {\mathcal{D}}\left(\boldsymbol{{x}}_{t}\right)) + \mathcal{Q}\left(\boldsymbol{{x}}_{t}\right)$ 
  
	   % $\boldsymbol{{x}}_{t} \leftarrow \boldsymbol{x}_{t} -s\nabla_{\boldsymbol{{x}}_{t}} \mathcal{L}\left(\boldsymbol{y}, {\mathcal{D}}\left(\boldsymbol{x_{t}}\right)\right)$ 

	   Sample $\boldsymbol{x}_{t-1}$ by $\mathcal{N}\left(\mu+s\nabla_{\boldsymbol{{x}}_{t}} \mathcal{L}^{total}_{\boldsymbol{{x}}_{t}}, \Sigma\right)$
	}
	\Return $\boldsymbol{x}_{0}$
\label{algo5}
\vspace{-0.1em}
\end{algorithm}

% When $p\left(\boldsymbol{y} \mid \boldsymbol{x}_t\right)$ is defined, and the corresponding gradients are calculated, 
In this way, the conditional transition $p_{\boldsymbol{\theta}}\left(\boldsymbol{x}_{t-1} \mid \boldsymbol{x}_t, \boldsymbol{y}\right)$ can be approximately obtained through the unconditional transition $p_{\boldsymbol{\theta}}\left(\boldsymbol{x}_{t-1} \mid \boldsymbol{x}_t\right)$ by shifting the mean of the unconditional distribution  by $- (s \Sigma \nabla_{\boldsymbol{x}_t} \mathcal{L}\left(\mathcal{D}(\boldsymbol{x}_t), \boldsymbol{y}\right) + \lambda \Sigma \nabla_{\boldsymbol{x}_t} \mathcal{Q}(\boldsymbol{x}_t))$
% The $\Sigma$ is the variance of $\boldsymbol{x}_t$.
However, we find that the way of adding guidance~\cite{avrahami2022blended} and the variance $\Sigma$ negatively influence the reconstructed images.

\subsection{Single Image Guidance}

% \begin{figure}[htbp]
%   \centering
%   \setlength{\abovecaptionskip}{0.cm}
%   \includegraphics[width=\linewidth]{cvpr2023-author_kit-v1_1-1/latex/GDM-t.pdf}
%   \caption{Overview of the \GDP-$x_t$.}
% \vspace{-0.5cm}
% \label{fig:GDM-t}
% \end{figure}

% !TEX root = ../PaperForReview.tex

\begin{algorithm}[t]\footnotesize
	\caption{\textbf{GDP-$\boldsymbol{x}_0$}: Conditioner guided diffusion sampling on $\boldsymbol{\tilde{x}}_{0}$, given a diffusion model $\left(\mu_{\theta}\left(\boldsymbol{x}_{t}\right), \Sigma_{\theta}\left(\boldsymbol{x}_{t}\right)\right)$, corrupted image conditioner $\boldsymbol{y}$.}
	\KwIn{Corrupted image $\boldsymbol{y}$, gradient scale $s$, degradation model $\mathcal{D}_\phi$ with randomly initiated parameters $\phi$, learning rate $l$ for optimizable degradation model, distance measure $\mathcal{L}$, optional quality enhancement loss $\mathcal{Q}$, quality enhancement scale $\lambda$.}
	\KwOut{Output image $\boldsymbol{x}_{0}$ conditioned on $\boldsymbol{y}$}
        Sample $\boldsymbol{x}_{T}$ from $\mathcal{N}(0, \mathbf{I})$
        
	\For{$t$ from $T$ to 1}{
	    $\mu, \Sigma = \mu_{\theta}\left(\boldsymbol{x}_{t}\right), \Sigma_{\theta}\left(\boldsymbol{x}_{t}\right)$
	    
	    $\boldsymbol{\tilde{x}}_{0} =  \frac{\boldsymbol{x}_{t}}{\sqrt{\bar{\alpha}_{t}}}-\frac{\sqrt{1-\bar{\alpha}_{t}} \epsilon_{\theta}\left(\boldsymbol{x}_{t}, t\right)}{\sqrt{\bar{\alpha}_{t}}}$

	    % $\boldsymbol{\widehat{x}}_{0} = {\mathcal{D}_\theta}\left(\boldsymbol{\tilde{x}}_{0}\right)$

        $\mathcal{L}^{total}_{\phi, \boldsymbol{\tilde{x}}_{0}} = \mathcal{L}(\boldsymbol{y}, {\mathcal{D}_\phi}\left(\boldsymbol{\tilde{x}}_{0}\right)) + \mathcal{Q}\left(\boldsymbol{\tilde{x}}_{0}\right)$ 

        $\phi \leftarrow \phi - l \nabla_{\phi} \mathcal{L}^{total}_{\phi, \boldsymbol{\tilde{x}}_{0}}$
            
	    % $\boldsymbol{\tilde{x}}_{0} \leftarrow \boldsymbol{\tilde{x}}_{0} -s\nabla_{\boldsymbol{\tilde{x}}_{0}} \mathcal{L}\left(\boldsymbol{y}, \left({\mathcal{D}_\phi}\left(\boldsymbol{\tilde{x}}_{0}\right)\right)\right)$\\

	   Sample $\boldsymbol{x}_{t-1}$ by $\mathcal{N}\left(\mu+s\nabla_{\boldsymbol{\tilde{x}}_{0}} \mathcal{L}^{total}_{\phi, \boldsymbol{\tilde{x}}_{0}}, \Sigma\right)$
  
        % Sample $\boldsymbol{x}_{t-1}$ by $q\left(\boldsymbol{x}_{t-1} \mid \boldsymbol{x}_{t}, \boldsymbol{\tilde{x}}_{0}\right) =\mathcal{N}\left(\boldsymbol{x}_{t-1}; \tilde{\boldsymbol{\mu}}_t\left(\boldsymbol{x}_{t}, \boldsymbol{\tilde{x}}_{0}\right), \tilde{\beta}_t \mathbf{I}\right)$, 
        
        % where $\quad \tilde{\boldsymbol{\mu}}_t\left(\boldsymbol{x}_{t}, \boldsymbol{\tilde{x}}_{0}\right)=\frac{\sqrt{\bar{\alpha}_{t-1}} \beta_t}{1-\bar{\alpha}_t} \boldsymbol{\tilde{x}}_{0}+\frac{\sqrt{\alpha_t}\left(1-\bar{\alpha}_{t-1}\right)}{1-\bar{\alpha}_t} \boldsymbol{x}_{t} \quad$ and $\quad \tilde{\beta}_t=\frac{1-\bar{\alpha}_{t-1}}{1-\bar{\alpha}_t} \beta_t$
	}
	\Return $\boldsymbol{x}_{0}$
\label{algo2}
\vspace{-0.1em}
\end{algorithm}

% \lyu{explain when to use single image guidance. What tasks use single image guidance.}
The super-resolution, impainting, colorization, deblurring, and enlighting tasks use single-image guidance.

\noindent\textbf{The Influence of Variance $\boldsymbol{\Sigma}$ on the Guidance.} In previous conditional diffusion models~\cite{dhariwal2021diffusion, wang2022guided}, the variance $\Sigma$ is applied for the mean shift in the sampling process, which is theoretically proved in the Appendix. In our work, we find that the variance $\Sigma$ might exert a negative influence on the quality of the generated images in our experiments. Therefore, we remove the variance during the guided denoising process to improve our performance. With the absence of $\Sigma$ and the fixed guidance scale $s$, the guided denoising process can be controlled by the variable scale $\hat{s}$.

\noindent\textbf{Guidance on $\boldsymbol{x}_t$.} Further, as vividly shown in Fig. \ref{fig:comparison}b, Algo. \ref{algo5} and Algo. 3 in Appendix, this class of guided diffusion models is the commonly used one~\cite{wu2022guided,dhariwal2021diffusion,liu2021more}, where the guidance is conditioned on $\boldsymbol{x}_t$ but with the absence of $\Sigma$, named GDP-$x_t$. However, this variant that applies the guidance on $\boldsymbol{x}_t$ may still yield less satisfactory quality images. 
% The intuition is that the generated images by commonly used conditional diffusion models and GDP-$x_t$ are still blurred in the background.
The intuition is $x_t$ is a noisy image with a specific noise magnitude, but $y$ is in general a corrupted image with no noise or noises of different magnitude. We lack reliable ways to define the distance between $x_t$ and $y$.
A naive MSE loss or perceptual loss will make $x_t$ deviate from its original noise magnitude and result in low-quality image generation.

\noindent\textbf{Guidance on $\boldsymbol{\tilde{x}}_0$.} 
% \lyu{distinguish $x_0$ and $\tilde{x}_0$.}
To tackle the problem as mentioned above, we systematically study the conditional signal applied on $\boldsymbol{\tilde{x}}_0$. Detailly, in the sampling process, the pre-trained DDPM usually first predicts a clean image $\boldsymbol{\tilde{x}}_0$ from the noisy image $\boldsymbol{x}_{t}$ by estimating the noise in $\boldsymbol{x}_{t}$, which can be directly inferred when given $\boldsymbol{x}_t$ by the Eq. \ref{eq4} in every timestep $t$. Then the predicted $\boldsymbol{\tilde{x}}_0$ together with $\boldsymbol{x}_{t}$ are utilized to sample the next step latent $\boldsymbol{x}_{t-1}$. We can add guidance on this intermediate variable $\boldsymbol{\tilde{x}}_0$ to control the generation process of the DDPM. The detailed sampling process can be found in Fig. \ref{fig:comparison}c and Algo. \ref{algo2}, where there is only one corrupted image.

% \begin{figure}[htbp]
%   \centering
%   \setlength{\abovecaptionskip}{0.cm}
%   \includegraphics[width=\linewidth]{cvpr2023-author_kit-v1_1-1/latex/GDM-0.pdf}
%   \caption{Overview of the \GDP-$x_0$.}
% \vspace{-0.5cm}
% \label{fig:GDM-0}
% \end{figure}

% \input{cvpr2023-author_kit-v1_1-1/latex/algos/algo2.tex}

\noindent\textbf{Known Degradation.} Several tasks~\cite{haris2018deep, kupyn2019deblurgan,larsson2016learning,yeh2017semantic} can be categorized into the class that the degradation function is known. In detail, the degradation model for image deblurring and super-resolution can be formulated as $\boldsymbol{y}=(\boldsymbol{x} \otimes \mathbf{k}) \downarrow_{\mathbf{s}}$. It assumes the low-resolution (LR) image is obtained by first convolving the high-resolution (HR) image with a Gaussian kernel (or point spread function) $\mathbf{k}$ to get a blurry image $\boldsymbol{x} \otimes \mathbf{k}$, followed by a down-sampling operation $\downarrow_{\mathrm{s}}$ with scale factor $\mathbf{s}$. The goal of image inpainting is to recover the missing pixels of an image. The corresponding degradation transform is to multiply the original image with a binary mask $\mathbf{m}$: $\psi(\boldsymbol{x})=\boldsymbol{x} \odot \mathbf{m}$, where $\odot$ is Hadamard's product. Further, image colorization aims at restoring a gray-scale image ${\boldsymbol{y}} \in$ $\mathbb{R}^{H \times W}$ to a colorful image with $\mathrm{RGB}$ channels $\boldsymbol{x} \in \mathbb{R}^{3 \times H \times W}$. To obtain ${\boldsymbol{y}}$ from the colorful image $\boldsymbol{x}$, the degradation transform $\psi$ is a graying transform that only preserves the brightness of $\boldsymbol{x}$. 
% \lyu{Average the color channel?}

% Note that, different from the above tasks, image colorization needs auxiliary loss such as color constancy loss (Sec. \ref{sec:loss}) to yield a more realistic picture.

% !TEX root = ../PaperForReview.tex

\begin{figure}[t]
  \centering
  \includegraphics[width=\linewidth]{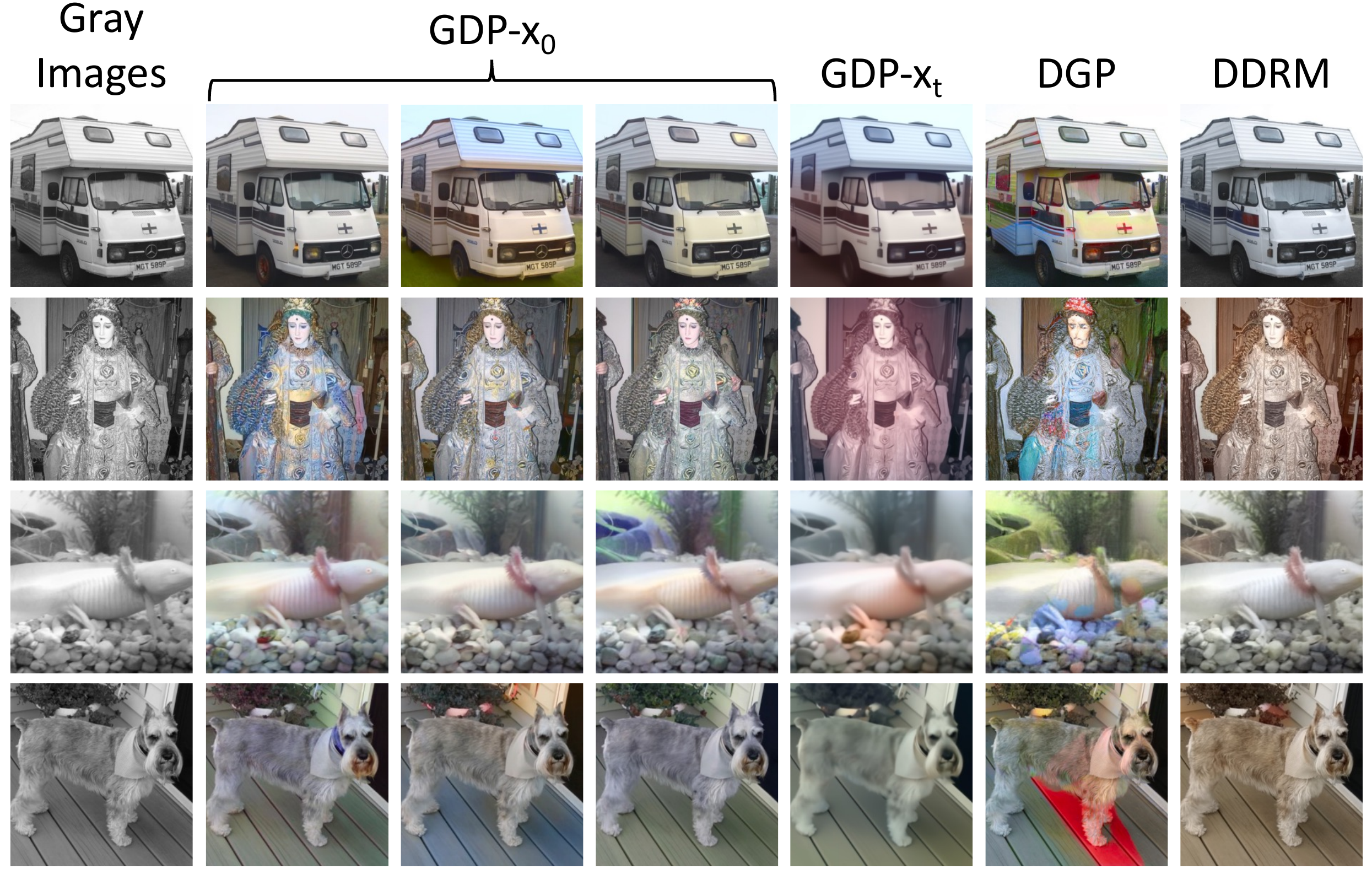}
  \vspace{-22pt}
  \caption{\textbf{Qualitative comparison of colorization results on ImageNet validation images}. GDP-$x_0$ generates various samples on the same input.}
\vspace{-2.1em}
\label{figure-color}
\end{figure}

\noindent\textbf{Unknown Degradation.} In the real world, many images undergo complicated degradations~\cite{zhang2021designing}, where the degradation models or the parameters of degradation models are unknown~\cite{wang2021real,liang2021flow}. 
In this case, the original images and the parameters of degradation models should be estimated simultaneously. For instance, in our work, the low-light image enhancement and the HDR recovery can be regarded as tasks with unknown degradation models. Here, we devise a simple but effective degradation model to simulate the complicated degradation, which can be formulated as follows:
\vspace{-0.1cm}
\begin{equation}
    \boldsymbol{y}=f \boldsymbol{x}+\boldsymbol{\mathcal{M}}, 
    % \text{where} f \text{ is a scalar and } 
    % \boldsymbol{\mathcal{M}} \text{ is a matrix}
\label{eq:light}
\end{equation}
\vspace{-0.1cm}
% \lyu{if $f$ is a scalar then you need to emphasize this and do not use bold letter for $f$.}
where the light factor $f$ is a scalar and the light mask $\boldsymbol{\mathcal{M}}$ is a vector of the same dimension as $\boldsymbol{x}$.
$f$ and $\boldsymbol{\mathcal{M}}$ are unknown parameters of the degradation model.
The reason that we can use this simple degradation model is that the transform between any pair of corrupted images and the corresponding high-quality image can be captured by $f$ and $\boldsymbol{\mathcal{M}}$ as long as they have the same size.
If they do not have the same size, we can first resize $\boldsymbol{x}$ to the same size as $\boldsymbol{y}$ and then apply this transform. 
It is worth noting that this degradation model is non-linear in general, since $f$ and $\boldsymbol{\mathcal{M}}$ depend on $\boldsymbol{x}$ and $\boldsymbol{y}$.
We need to estimate $f$ and $\boldsymbol{\mathcal{M}}$ for every individual corrupted image.
We achieve this by randomly initializing them and synchronously optimizing them in the reverse process of DDPMs as shown in Algo. ~\ref{algo2}.

% Specifically, $\boldsymbol{f}$ and $\boldsymbol{\mathcal{M}}$ are randomly initialized and synchronously optimized in the reverse process of DDPMs.

% Apart from the loss for image restoration, the enlighten task owns additional loss for controlling the degree of the lightness, including the exposure control loss and illumination smoothness loss (Sec. \ref{sec:loss}).

% \begin{figure}[htbp]
%   \centering
%   \setlength{\abovecaptionskip}{0.cm}
%   \includegraphics[width=\linewidth]{cvpr2023-author_kit-v1_1-1/latex/HDR-GDM-0.pdf}
%   \caption{Overview of the HDR-\GDP-$x_0$.}
% \vspace{-0.5cm}
%  \label{fig:HDR-GDM-0}
% \end{figure}

% \input{cvpr2023-author_kit-v1_1-1/latex/algos/algo3.tex}

\subsection{Extended version}

% !TEX root = ../PaperForReview.tex

% \usepackage{multirow}
% \usepackage{graphicx}
\begin{table*}[htbp]
\centering
\tabcolsep=0.08cm
\caption{\textbf{Quantitative comparison of  linear image restoration tasks on ImageNet 1k~\cite{pan2021exploiting}.} \GDP outperforms other methods in terms of FID and Consistency across all tasks.}
\vspace{-8pt}
\resizebox{\textwidth}{!}{%
\begin{tabular}{l|cccc|cccc|cccc|cccc}
\toprule[1.5pt]
\multirow{2}{*}{\large Method} & \multicolumn{4}{c|}{4$\times$ Super-resolution}           & \multicolumn{4}{c|}{Deblur}          & \multicolumn{4}{c|}{25$\%$ Inpainting} & \multicolumn{4}{c}{Colorization}    \\ \cmidrule{2-17}
                      & PSNR $\uparrow$  & SSIM $\uparrow$  & Consistency $\downarrow$ & FID $\downarrow$   & PSNR $\uparrow$  & SSIM $\uparrow$  & Consistency$\downarrow$ & FID $\downarrow$   & PSNR $\uparrow$  & SSIM $\uparrow$  & Consistency$\downarrow$ & FID $\downarrow$   & PSNR $\uparrow$  & SSIM $\uparrow$ & Consistency $\downarrow$ & FID $\downarrow$  \\ \midrule[1pt]
DGP~\cite{pan2021exploiting}                   & 21.65 & 0.56 & 158.74      & 152.85 & 26.00 & 0.54 & 475.10       & 136.53 & 27.59 & 0.82 & 414.60       & 60.65  & 18.42 & 0.71 & 305.59      & 94.59 \\
SNIPS~\cite{kawar2021snips}                 & 22.38 & 0.66 & 21.38       & 154.43 & 24.73 & 0.69 & 60.11       & 17.11  & 17.55 & 0.74 & 587.90       & 103.50 & -     & -     & -           & -     \\
RED~\cite{romano2017little}                  & 24.18 & 0.71 & 27.57       & 98.30  & 21.30 & 0.58 & 63.20       & 69.55  & -     & -     & -           & -      & -     & -     & -           & -     \\
DDRM~\cite{kawar2022denoising}                 & \textbf{26.53} & \textbf{0.78} & 19.39       & 40.75  & \textbf{35.64} & \textbf{0.98} & 50.24       & 4.78   & 34.28 & 0.95 & \textbf{4.08}        & 24.09  & \textbf{22.12} & 0.91 & 37.33       & 47.05 \\
GDP-$x_t$                & 24.27 & 0.67 & 80.32       & 64.67  & 25.86 & 0.75 & 54.08       & 5.00   & 31.06 & 0.93 & 8.80        & 20.24  & 21.30 & 0.86 & 75.24       & 66.43 \\
GDP-$x_0$               & 24.42 & 0.68 & \textbf{6.49}        & \textbf{38.24}  & 25.98 & 0.75 & \textbf{41.27}       & \textbf{2.44}   & \textbf{34.40} & \textbf{0.96} & 5.29        & \textbf{16.58}  & 21.41 & \textbf{0.92} & \textbf{36.92}       & \textbf{37.60} \\ \bottomrule[1.5pt]
\end{tabular}%
}
\vspace{-1em}
\label{table1}
\end{table*}

\begin{table*}[htbp]
\centering
\caption{\textbf{Quantitative comparison of image enlighten task on LOL~\cite{wei2018deep}, VE-LOL-L~\cite{liu2021benchmarking}, and LoLi-phone~\cite{li2021low} benchmarks.} Bold font indicates the best performance in zero-shot learning, and the underlined font denotes the best results in all models.}
\vspace{-8pt}
\resizebox{1\textwidth}{!}{%
\begin{tabular}{l|l|ccccc|ccccc|cc}
\toprule[1.5pt]
\multirow{2}{*}{Learning} & \multirow{2}{*}{Methods} & \multicolumn{5}{c|}{LOL~\cite{wei2018deep}}                   & \multicolumn{5}{c|}{VE-LOL-L~\cite{liu2021benchmarking}}                & \multicolumn{2}{c}{LoLi-Phone~\cite{li2021low}} \\ \cmidrule{3-14} 
                          &                          & PSNR $\uparrow$  & SSIM$\uparrow$  & FID$\downarrow$    & LOE$\downarrow$    & PI$\downarrow$   & PSNR$\uparrow$   & SSIM$\uparrow$  & FID $\downarrow$    & LOE$\downarrow$    & PI$\downarrow$   & LOE$\downarrow$            & PI$\downarrow$          \\ \midrule[1pt]
\multirow{8}{*}{Supervised learning}       & LLNet  \cite{lore2017llnet}                   & \underline{17.91} & 0.76 & 169.20 & 384.21 & \underline{4.10} & 17.38 & 0.73  & 124.98 & 291.59  & \underline{5.54}  & 343.34         & \underline{5.36}        \\
                          & LightenNet~\cite{li2018lightennet}               & 10.29 & 0.45 & 90.91  & 273.21 & 7.09  & 13.26  & 0.57 & 82.26  & 199.45 & 7.29 & 500.22         & 6.63        \\
                          & Retinex-Net~\cite{wei2018deep}              & 17.24 & 0.55  & 129.99  & 513.28 & 8.63 & 16.41 & 0.64 & 135.20 & 421.41 & 8.62 & 542.29         & 8.23         \\
                          & MBLLEN~\cite{lv2018mbllen}                   & 17.90 & 0.77 & 122.69 & 175.10 & 8.39  & 15.95 & 0.70 & 105.74 & 114.91 & 7.45 & 137.34         & 6.46        \\
                          & KinD~\cite{zhang1019kindling}                   & 17.57 & 0.82 & \underline{74.52}  & 377.59 & 7.41 & 18.07  & 0.78  & 80.12   & 253.79 & 7.51 & 265.47         & 6.84        \\
                          & KinD++~\cite{zhang2021beyond}                  & 17.60 & 0.80 & 100.15 & 712.12 & 7.96 & 16.80 & 0.74 & 101.23 & 421.97 & 7.98 & 382.51         & 7.71        \\
                          & TBFEN~\cite{lu2020tbefn}                    & 17.25 & \underline{0.83} & 90.59   & 367.66 & 8.29 & \underline{18.91} & \underline{0.81}  & 91.30  & 276.65 & 8.02 & 214.30         & 7.34        \\
                          & DSLR~\cite{lim2020dslr}                    & 14.98 & 0.67 & 183.92 & 272.68 & 7.09 & 15.70 & 0.68 & 124.80 & 271.63 & 7.27 & 281.25         & 6.99       \\ \midrule[1pt]
Unsupervised learning                        & EnlightenGAN~\cite{jiang2021enlightengan}             & 17.44 & 0.74 & 82.60  & 379.23 & 8.78 & 17.45 & 0.75 & 86.51  & 311.85 & 8.27 & 373.41         & 7.26        \\ \midrule[1pt]
Self-supervised learning                       & DRBN~\cite{yang2020from}                    & 15.15 & 0.52 & 94.96  & 692.99 & 5.53 & 18.47 & 0.78 & 88.10  & 268.70   & 6.15 & 285.06         & 5.31        \\ \midrule[1pt]
\multirow{6}{*}{Zero-shot learning}      & ExCNet~\cite{zhang2019zero}                   & \textbf{16.04} & 0.62 & 111.18 & 220.38 & 8.70  & 16.20 & 0.66 & 115.24 & 225.15 & 8.62 & 359.96         & 7.95        \\
                          & Zero-DCE~\cite{guo2020zero}                 & 14.91 & \textbf{0.70} & 81.11  & 245.54  & 8.84 & \textbf{17.84} & \textbf{0.73} & 85.72  & 194.10 & 8.12 & 214.30         & 7.34        \\
                          & Zero-DCE++~\cite{li2021learning}               & 14.86   & 0.62 & 86.22   & 302.06  & 7.08 & 16.12 & 0.45 & 86.96  & 313.50 & 7.92 & 308.15         & 7.18        \\
                          & RRDNet~\cite{zhu2020zero}                   & 11.37 & 0.53  & 89.09  & 127.22 & 8.17   & 13.99 & 0.58 & 83.41  & 94.23  & 7.36 & 92.73          & 7.20          \\
                          & GDP-$x_t$                    & 7.32  & 0.57 & 238.92  & 364.15 & 8.26 & 9.45  & 0.50 & 152.68  & 194.49 & 7.12 & 508.73          & 8.06        \\
                          & GDP-$x_0$                     & 13.93 & 0.63 & \textbf{75.16} & \underline{\textbf{110.39}} & \textbf{6.47} & 13.04 & 0.55 & \underline{\textbf{78.74}} & \underline{\textbf{79.08}}  & \textbf{6.47} & \underline{\textbf{75.29}}          & \textbf{6.35}        \\ \bottomrule[1.5pt]
\end{tabular}%
}
\vspace{-1.5em}
\label{table-enlighten}
\end{table*}

\noindent\textbf{Multi-images Guidance.} Under specific circumstances, there are several images could be utilized to guide the generation of a single image~\cite{niu2021hdr, zheng2021ultra}, which is merely studied and much more challenging than single-image guidance. To this end, we propose the HDR-\GDP for the HDR image recovery with multiple images as guidance, consisting of three input LDR images, \textit{i.e.} short, medium, and long exposures. Similar to low-light enhancement, the degradation models are also treated as Eq. \ref{eq:light}, where the parameters remain unknown that determine the HDR recovery is the blind problem. However, as shown in Fig. \ref{fig:comparison}c and Algo. 5 in Appendix, in the reverse process, there are three corrupted images ($n$ = 3) to guide the generation so that \textbf{three pairs of blind parameters} for three LDR images are randomly initiated and optimized. 
% \lyu{ref appendix alg. emphasize we have 3 set of parameters.}

% But there are three independent degradation models for three LDR images, respectively.

% !TEX root = ../PaperForReview.tex

\begin{figure}[t]
   \centering
   \includegraphics[width=\linewidth]{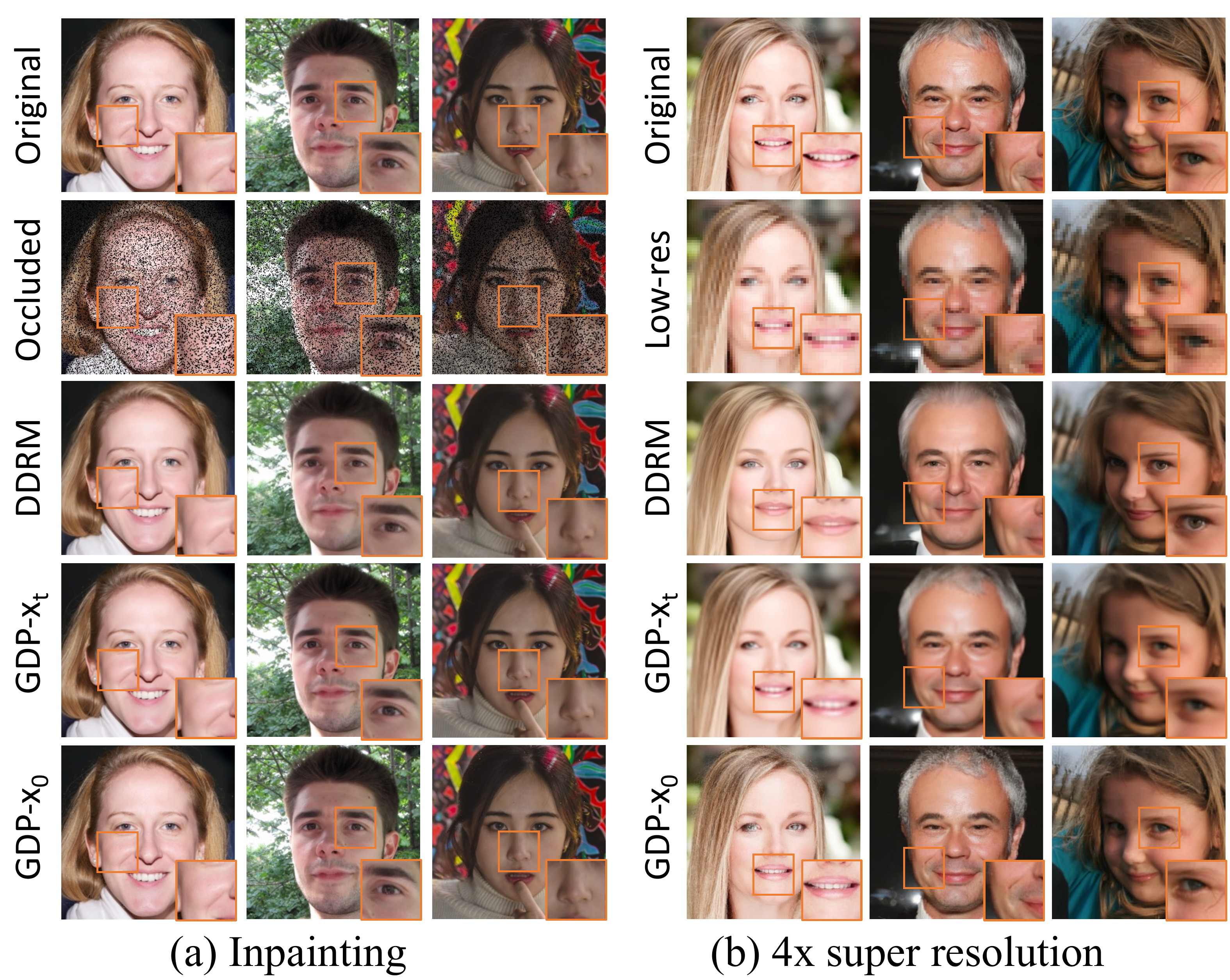}
   \vspace{-19pt}
   \caption{\textbf{Qualitative results of (a) 25 $\%$ inpainting and (b) $4\times$ super-resolution on CelebA~\cite{karras2017progressive}.}}
\vspace{-1.8em}
\label{figure-face}
\end{figure}

\noindent\textbf{Restore Any-size Image.} 
% \lyu{more specific about the ddpm, 256.}
Furthermore, the pre-trained diffusion models provide by~\cite{dhariwal2021diffusion} with the size of 256 are only able to generate the fixed size of images, while the sizes of images from various image restoration are diverse. 
Herein, we employ the patch-based method as \cite{liu2021benchmarking} to tackle this problem. 
% But differently, shown in Fig.~\ref{SM:hierarchial} and Algo.~\ref{algo:patch} in the Supplementary Material, we resize the corrupted images $\boldsymbol{y} \in \mathbb{R}^{3 \times H \times W} $ to $\boldsymbol{\overline{y}} \in \mathbb{R}^{3 \times 256 \times \overline{W} \text{ or } 3 \times \overline{H} \times 256}$, then apply the patch-based methods on the reshaped images. After, the light masks $\boldsymbol{\mathcal{\overline{M}}}$ are interpolated to the original image size to obtain $\boldsymbol{\mathcal{M}}$, representing the global light shift. 
% Following that, the light factor $f$ and the light mask $\boldsymbol{\mathcal{M}}$ will be fixed and utilized to generate the image patches, which will be recombined as the output images. 
By the merits of this patch-based strategy (Fig. 13 and Algo. 6 in the Appendix), GDP can be extended to recover the images of arbitrary resolution to promote the versatility of the GDP.
% \lyu{unclear. better give a illustration in the main text.}

\section{Loss Function}
\label{sec:loss}
% \begin{figure}[t]
%   \centering
%   \includegraphics[width=\linewidth]{cvpr2023-author_kit-v1_1-1/Manuscript/figure-blur-bedroom-few.pdf}
%   \caption{Deblurring results on bedroom images.}
% \vspace{-0.5cm}
% \end{figure}

In GDP, the loss function can be divided into two main parts: Reconstruction loss and quality enhancement loss, where the former aims to recover the information contained in the conditional signal while the latter is integrated to promote the quality of the final outputs. 
% \lyu{you need to first introduce the individual losses and then explain what tasks use what losses. And the give weight of each loss.}

% \subsection{Reconstruction Loss}
\noindent\textbf{Reconstruction Loss.} The reconstruction loss can be MSE, structural similarity index measure (SSIM), perceptual loss, or other reconstructive loss. Here, we primarily choose MSE loss as our reconstruction loss. 

% Here, we briefly discuss the denoising loss function. Given a training output image $\mathbf{x}^*$, we generate a noisy version $\mathbf{\tilde{x}}^*$, and train a neural network $f_\theta$ to denoise $\mathbf{\tilde{x}}^*$ given $\boldsymbol{x}$ and a noise level indicator $\gamma$, for which the L2 norm (p = 2) loss is:

% \begin{equation}
% \mathbb{E}_{(\mathbf{x}, \mathbf{x}^*)} \mathbb{E}_{\boldsymbol{\epsilon} \sim \mathcal{N}(0, I)} \mathbb{E}_\gamma\left\|f_\theta(\mathbf{x}, \underbrace{\sqrt{\gamma} \mathbf{x}^*+\sqrt{1-\gamma} \boldsymbol{\epsilon}}_{\widetilde{\mathbf{x}^*}}, \gamma)-\boldsymbol{\epsilon}\right\|_p^p,
% \end{equation}

% !TEX root = ../PaperForReview.tex

\begin{figure}[t]
  \centering
   \includegraphics[width=\linewidth]{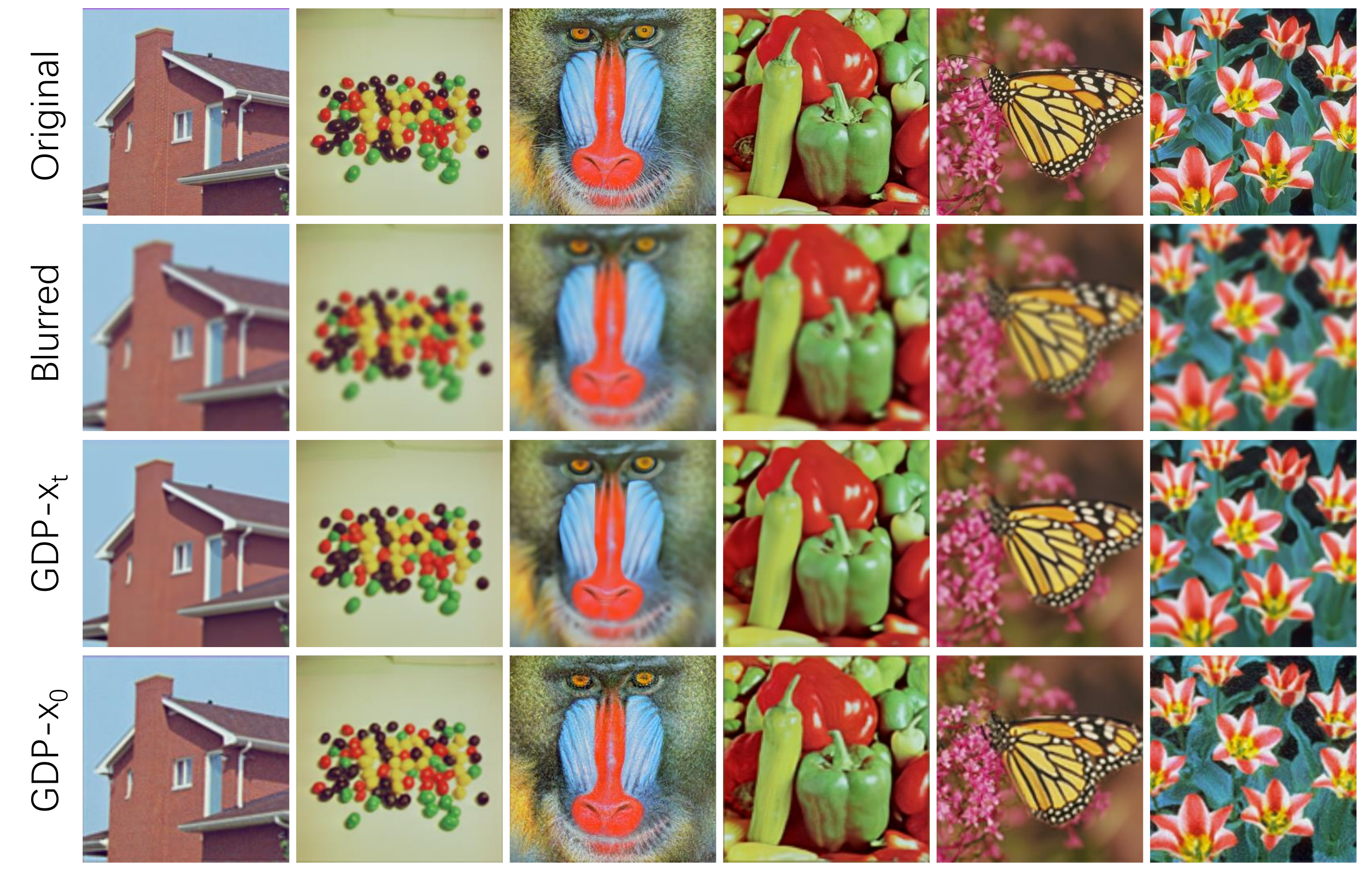}
   \vspace{-20pt}
   \caption{\textbf{Results of image deblurring task on 256 $\times$ 256 USC-SIPI images~\cite{weber2006usc} using an ImageNet model.
   % \junzhe{Results of <task> on <dataset>.}
   }}
\vspace{-1.8em}
\label{figure-ood}
\end{figure}

% !TEX root = ../PaperForReview.tex

\begin{figure}[t]
    \centering
    \includegraphics[width=\linewidth]{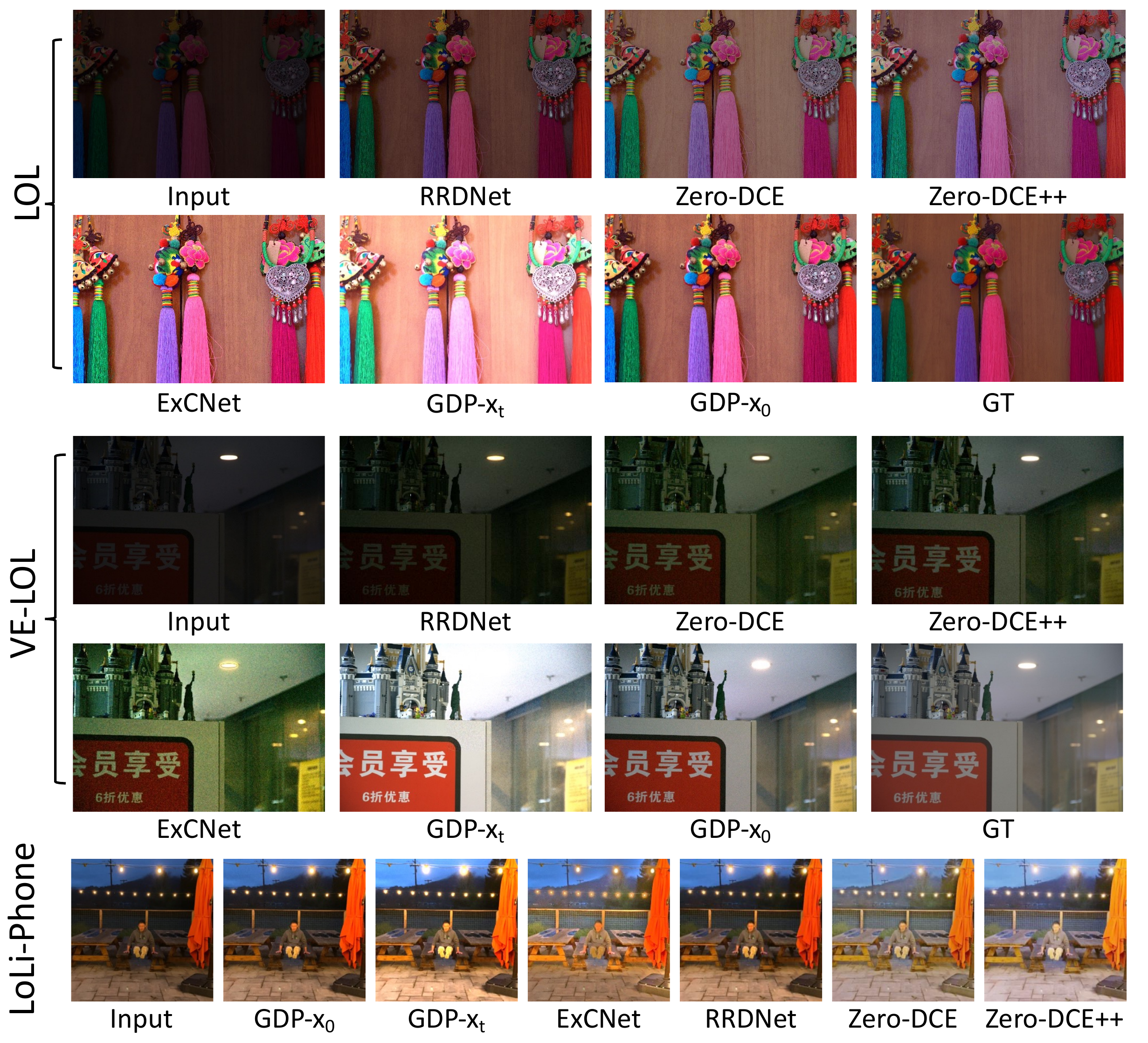}
    \vspace{-20pt}
    \caption{\textbf{Qualitative results of low-light enhancement on the LOL~\cite{wei2018deep}, VE-LOL~\cite{liu2021benchmarking}, and LoLi-Phone~\cite{li2021low} datasets.}}
\vspace{-1.8em}
\label{figure-light}
\end{figure} 

\noindent\textbf{Quality Enhancement Loss.}
%不同亮度的调节（正文里）。介绍loss之前general表达。
% \noindent\textbf{Exposure Control Loss.}
\textit{1) \underline{Exposure Control Loss}:} To enhance the versatility of \GDP, an exposure control loss $L_{exp}$~\cite{guo2020zero} is employed to control the exposure level for low-light image enhancement, which is written as:
% The exposure control loss tends to regulate the average intensity value of a local region close to the well-exposedness level $E$. Following the previous works~\cite{mertens2007exposure, mertens2009exposure}, $E$ is set as the gray level in the RGB color space. As expected, the $E$ can be adjusted to control the lightness in our experiments. Thes loss $L_{\text{exp}}$ is written as:
\vspace{-0.1cm}
\begin{equation}\small
    L_{\text{exp}}=\frac{1}{U} \sum_{k=1}^U\left|R_k-E\right|,
\end{equation}
\vspace{-0.1cm}
where $U$ stands for the number of non-overlapping local regions of size $8 \times 8$, and $R$ represents the average intensity value of a local region in the reconstructed image.
Following the previous works~\cite{mertens2007exposure, mertens2009exposure}, $E$ is set as the gray level in the RGB color space. As expected, $E$ can be adjusted to control the brightness in our experiments. 

% \noindent\textbf{Color Constancy Loss.}
\textit{2) \underline{Color Constancy Loss}:} Following the Gray-World color constancy hypothesis~\cite{buchsbaum1980spatial}, a color constancy loss $L_{\text{col}}$ is exploited to correct the potential color deviations in the restored image and bridge the relations among the three adjusted channels in the colorization task, formulated as:
\vspace{-0.1cm}
\begin{equation}\footnotesize
    L_{\text{col}}=\sum_{\forall(m, n) \in \varepsilon}\left(Y^m-Y^n\right)^2\text{,} \varepsilon=\{(R, G),(R, B),(G, B)\}
\end{equation}
\vspace{-0.1cm}
where $Y^m$ is the average intensity value of $m$ channel in the recovered image, $(m, n)$ represents a pair of channels.

% \noindent\textbf{Illumination Smoothness Loss.} 
\textit{3) \underline{Illumination Smoothness Loss}:} To maintain the monotonicity relations between neighboring pixels in the optimized light mask $\mathcal{M}$, an illumination smoothness loss~\cite{guo2020zero} is utilized for each light variance $\mathcal{M}$. The illumination smoothness loss $L_{tv_{\mathcal{M}}}$ is defined as:
\vspace{-0.2cm}
\begin{equation}\footnotesize
    L_{t v_{\mathcal{M}}}=\frac{1}{N} \sum_{n=1}^N \sum_{c \in \xi}\left(\left|\nabla_h \mathcal{M}_n^c\right|+\nabla_v \mathcal{M}_n^c \mid\right)^2\text{,}\xi=\{R, G, B\},
\end{equation}
\vspace{-0.1cm}
where $N$ is iteration times, $\nabla_h$ and $\nabla_v$ are the horizontal and vertical gradient operations, respectively.

Specifically, the image colorization task uses color constancy loss to obtain more natural colors. 
The low-light enhancement requires color constancy loss for the same reason.
In addition, the low-light enhancement task uses illumination smoothness loss to make the estimated light mask $\boldsymbol{\mathcal{M}}$ smoother. 
Exposure control loss enables us to manually control the brightness of the restored image. The weights of the losses can be found in Appendix.
% \lyu{explain weights of the losses}

% !TEX root = ../PaperForReview.tex

\section{Experiments}

% We apply \GDP to a suite of challenging image restoration tasks: (1) \textbf{HDR image recovery} aims to obtains a HDR images with the aids of three LDR images. (2) \textbf{Enlighting} enables the dark images turned into normal images. (3) \textbf{Colorization} transforms an input grayscale image to a plausible color image. (4) \textbf{Inpainting} fills in user-specified masked regions of an image with realistic content. (5) \textbf{Super-resolution} extends a low resolution image into a higher one. (6) \textbf{Deblurring} corrects the blurred images, restoring plausible image detail.

% We apply \GDP to a suite of challenging image restoration tasks: (1) \textbf{Super-resolution}; (2) \textbf{Deblurring}; (3) \textbf{Inpainting}; (4) \textbf{Colorization}; (5) \textbf{Low-light enhancement}; (6) \textbf{HDR image recovery}. We do so without task-specific hyperparameter tuning and architecture customization. Inputs and outputs of the first four tasks are represented as $256 \times 256$ RGB images, while the last two tasks are various ($1900 \times 1060$ for HDR image recovery and $600 \times 400$ for image enlightening, respectively). Each task presents its own unique challenges, together with Implementation details can be found in the Supplementary Material.

In this section, we systematically compare \GDP, which uses \textbf{a single unconditional DDPM pre-trained on ImageNet} provide by~\cite{dhariwal2021diffusion}, with other methods of various image restoration and enhancement tasks, and ablate the effectiveness of the proposed design. We furthermore list details on implementation, datasets, evaluation, and more qualitative results for all tasks in Appendix. 
% \lyu{you need to emphasize that we use a single DDPM model trained on ImageNet.}

% !TEX root = ../PaperForReview.tex

\begin{figure}[t]
  \centering
   \includegraphics[width=\linewidth]{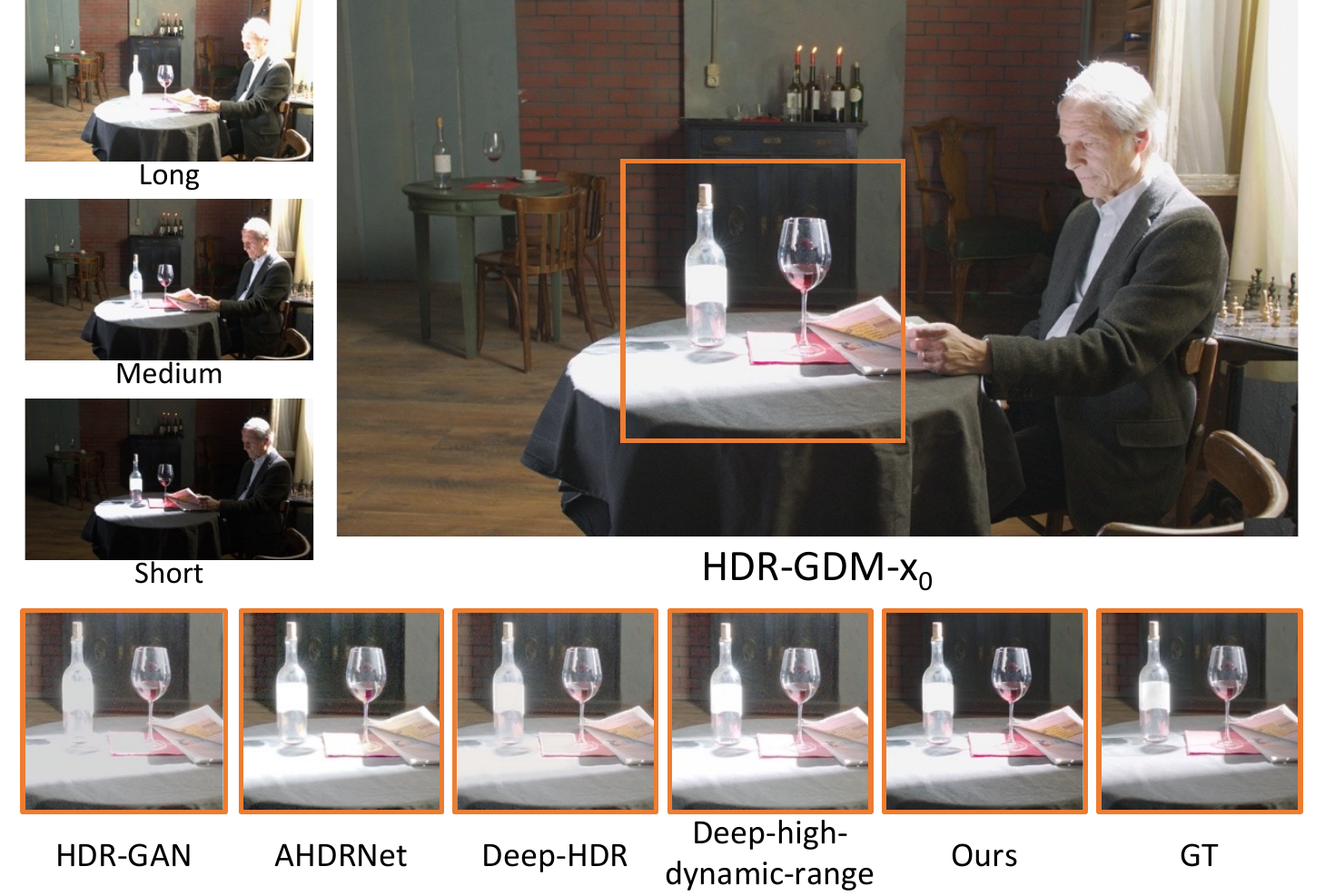}
   \vspace{-20pt}
   \caption{\textbf{Example from the NTIRE dataset ~\cite{perez2021ntire}.} We compare a set of patches cropped from the tone-mapped HDR images generated by state-of-the-art methods.}
\vspace{-1.8em}
\label{figure-hdr}
\end{figure}

\subsection{Linear and Multi-linear Degradation Tasks}
\label{sec:linear}
% \lyu{compare with dgp}
% \lyu{test hybrid degradation}
Aiming at quantifying the performance of GDP, we focus on the ImageNet dataset for its diversity. 
% The \GDP is performed under various linear tasks on the ImageNet dataset and the corresponding results is listed in Table \ref{table1}.
For each experiment, we report the average peak signal-to-noise ratio (PSNR), SSIM, and Consistency to measure faithfulness to the original image and the FID to measure the resulting image quality. \GDP is compared with other unsupervised methods that can operate on ImageNet, including RED~\cite{romano2017little}, DGP~\cite{pan2021exploiting}, SNIPS~\cite{kawar2021snips}, and DDRM~\cite{kawar2022denoising}. We evaluate all methods on the tasks of $4 \times$ super-resolution, deblurring, impainting, and colorization on one validation set from each of the $1000$ ImageNet classes, following~\cite{pan2021exploiting}. Table \ref{table1} shows that GDP-$x_0$ outperforms other methods in Consistency and FID. The only exception is that DDRM achieves better PSNR and SSIM than GDP, but it requires higher Consistency and FID~\cite{dhariwal2021diffusion,ho2022cascaded,saharia2022image,chen2018fsrnet,dahl2017pixel,dosovitskiy2016generating}. \GDP produces high-quality reconstructions across all the tested datasets and problems, which can be seen in Appendix. As a posterior sampling algorithm, \GDP can produce multiple outputs for the same input, as demonstrated in the colorization task in Fig. \ref{figure-color}. Moreover, the unconditional ImageNet DDPMs can be used to solve inverse problems on out-of-distribution images with general content. In Figs. \ref{figure-face} and \ref{figure-ood}, and more illustrations in Appendix, we show GDP successfully restores $256 \times 256$ images from USC-SIPI~\cite{weber2006usc}, LSUN~\cite{yu2015lsun}, and CelebA~\cite{karras2017progressive}, which do not necessarily belong to any ImageNet class. \GDP can also restore the images under multi-degradation (Fig. \ref{teaser} and Appendix).

% !TEX root = ../PaperForReview.tex

% \usepackage{graphicx}
\begin{table}[t]\scriptsize
\centering
\caption{\textbf{Quantitative comparison on the NTIRE dataset~\cite{perez2021ntire}.}}
\vspace{-8pt}
{%
\begin{tabular}{l|cccc}
\toprule[1.5pt]
Methods                                                            & PSNR$\uparrow$   & SSIM$\uparrow$   & LPIPS $\downarrow$ & FID$\downarrow$ \\ \midrule[1pt]
AHDRNet~\cite{yan2019attention}                                                           & 18.72 & 0.58 & 0.39 & 81.98 \\
HDR-GAN~\cite{niu2021hdr}                                                            & 21.67 & 0.74 & 0.26 & 52.71 \\
Deep-HDR~\cite{wu2018deep}                                                          & 21.66 & 0.76 & 0.26 & 57.52 \\
\begin{tabular}[l]{@{}c@{}} Deep-high-dyna\\mic-range~\cite{kalantari2017deep}\end{tabular} & 21.33 & 0.71 & 0.26 & 51.92 \\
GDP-$x_t$                                                            & 19.36     & 0.65     & 0.30 & 63.89      \\ \midrule[1pt]
GDP-$x_0$                                                             & \textbf{24.88} & \textbf{0.86} & \textbf{0.13} & \textbf{50.05} \\ \bottomrule[1.5pt]
\end{tabular}%
}
\vspace{-1.8em}
\label{table-hdr}
\end{table}

\subsection{Exposure Correction Tasks}
\label{sec:exposure}

Encouraged by the excellent performance on the linear inverse problem, we further evaluate our \GDP on the low-light image enhancement, which is categorized into non-linear and blind issues. Following the previous works~\cite{li2021low}, the three datasets LOL~\cite{wei2018deep}, VE-LOL-L~\cite{liu2021benchmarking}, and the most challenging LoLi-phone~\cite{li2021low} are leveraged to test the capability of \GDP on low-light enhancement. As shown in Table \ref{table-enlighten}, our GDP-$x_0$ fulfills the best FID, lightness order error (LOE)~\cite{wang2013naturalness}, and perceptual index (PI)~\cite{mittal2012making} across all the zero-shot methods under three datasets. The lower LOE demonstrates better preservation for the naturalness of lightness, while the lower PI indicates better perceptual quality. In Fig. \ref{figure-light} and Appendix, our GDP-$x_0$ yields the most reasonable and satisfactory results across all methods. For more control, by the merits of Exposure Control Loss, the brightness of the generated images can be adjusted by the well-exposedness level $E$ (Fig. \ref{teaser} and Appendix)

\subsection{HDR Image Recovery}
\label{sec:hdr}

To evaluate our model on the HDR recovery~\cite{li2021low}, we compare HDR-GDP-$x_0$ with the state-of-the-art HDR methods on the test images in the HDR dataset from the NTIRE2021 Multi-Frame HDR Challenge~\cite{perez2021ntire}, from which we randomly select 100 different scenes as the validation. Each scene consists of three LDR images with various exposures and corresponding HDR ground truth. The state-of-the-art methods used for comparison include AHDRNet~\cite{yan2019attention}, HDR-GAN~\cite{niu2021hdr}, DeepHDR~\cite{wu2018deep} and deep-high-dynamic-range~\cite{kalantari2017deep}. The quantitative results are provided in Table \ref{table-hdr}, where HDR-GDP-$x_0$ performs best in PSNR, SSIM, LPIPS, and FID. As shown in Fig. \ref{figure-hdr} and Appendix, HDR-GDP-$x_0$ achieves a better quality of reconstructed images, where the low-light parts can be enhanced, and the over-exposure regions are adjusted. Moreover, HDR-GDP-$x_0$ recovers the HDR images with more clear details.

% \lyu{Fix the parameter of Unknown Degradation}

% \lyu{Naively handle large size images}

% !TEX root = ../PaperForReview.tex

% \section{Ablation Study}
\subsection{Ablation Study}

\noindent\textbf{The Effectiveness of the Variance $\boldsymbol{\Sigma}$ and the Guidance Protocol.} The ablation studies on the variance $\Sigma$ and two ways of guidance are performed to unveil their effectiveness. As shown in Table.~\ref{ablation1}, the performance of GDP-$x_t$ and GDP-$x_0$ is superior to GDP-$x_t$ with $\Sigma$ and GDP-$x_0$ with $\Sigma$, respectively, verifying the absence of variance $\Sigma$ can yield better quality of images. Moreover, the results of GDP-$x_0$ and GDP-$x_0$ with $\Sigma$ are better than GDP-$x_t$ and GDP-$x_t$ with $\Sigma$, respectively, demonstrating the superiority of the guidance on $\boldsymbol{x}_0$ protocol.

% !TEX root = ../PaperForReview.tex

% \usepackage{multirow}
% \usepackage{graphicx}
\begin{table}[t]\scriptsize
\centering
\caption{\textbf{The ablation study on the variance $\Sigma$ and the way of the guidance.}}
\vspace{-8pt}
\tabcolsep=0.07cm
{%
\begin{tabular}{l|cccc|cccc}
\toprule[1pt]
\multirow{2}{*}{Task}             & \multicolumn{4}{c|}{4$\times$ Super resolution}           & \multicolumn{4}{c}{Deblur}          \\ \cline{2-9} 
                                  & PSNR   & SSIM  & Consistency & FID   & PSNR  & SSIM  & Consistency & FID   \\ \midrule[0.8pt]
\multicolumn{1}{l|}{\begin{tabular}[c]{@{}c@{}}  \GDP-$x_t$ \\ with $\Sigma$\end{tabular} } & 22.86      & 0.60     & 88.37           & 68.04     & 22.06     & 0.57     & 69.46           & 80.39     \\ \begin{tabular}[c]{@{}c@{}}\GDP-$x_0$ \\ with $\Sigma$\end{tabular}                       & 22.09      & 0.58     & 93.19           & 41.22     & 23.49     & 0.65     & 68.67           & 50.29     \\
\begin{tabular}[c]{@{}c@{}} \GDP-$x_t$\end{tabular}                            & 24.27  & 0.67 & 80.32       & 64.67 & 25.86 & 0.73 & 54.08       & 5.00  \\ \midrule[0.8pt]
\GDP-$x_0$                            & \textbf{24.42}  & \textbf{0.68} & \textbf{6.49}        & \textbf{38.24} & \textbf{25.98} & \textbf{0.75} & \textbf{41.27}       & \textbf{2.44}  \\ \midrule[0.8pt] \midrule[0.8pt]
\multirow{2}{*}{Task}             & \multicolumn{4}{c|}{25$\%$ Inpainting} & \multicolumn{4}{c}{Colorization}    \\ \cline{2-9} 
                                  & PSNR   & SSIM  & Consistency & FID   & PSNR  & SSIM  & Consistency & FID   \\ \midrule[0.8pt] 
\multicolumn{1}{l|}{\begin{tabular}[c]{@{}c@{}}\GDP-$x_t$ \\ with $\Sigma$\end{tabular} } & 25.28      & 0.70     & 171.44           & 73.32     & 17.67     & 0.70     & 246.26           & 145.20    \\
\begin{tabular}[c]{@{}c@{}}\GDP-$x_0$ \\ with $\Sigma$\end{tabular}                      & 24.58      & 0.75     & 65.59           & 22.77     & 21.28     & 0.91     & 66.57           & 38.39     \\
\begin{tabular}[c]{@{}c@{}}\GDP-$x_t$\end{tabular}                          & 31.06  & 0.93 & 8.80        & 20.24 & 21.30 & 0.86 & 75.24       & 66.43 \\
\GDP-$x_0$                            & \textbf{34.40}  & \textbf{0.96} & \textbf{5.29}        & \textbf{16.58} & \textbf{21.41} & \textbf{0.92} & \textbf{36.92}       & \textbf{37.60} \\ \bottomrule[1pt]
\end{tabular}%
}
\vspace{-1.5em}
\label{ablation1}
\end{table}

% !TEX root = ../PaperForReview.tex

% \usepackage{multirow}
% \usepackage{graphicx}
\begin{table}[t]\scriptsize
\centering
\caption{\textbf{The ablation study on the optimizable degradation and patch-based tactic.}}
\vspace{-8pt}
\tabcolsep=0.09cm
{%
\begin{tabular}{l|ccccc|cccc}
\toprule[1.5pt]
\multirow{2}{*}{Methods} & \multicolumn{5}{c|}{LOL}                   & \multicolumn{4}{c}{NTIRE}      \\ \cline{2-10}
                         & PSNR   & SSIM  & FID     & LOE     & PI    & PSNR  & SSIM  & LPIPS  & FID   \\ \midrule[1pt]
% Model A                  & 13.80      & 0.61     & 119.21       & 184.65       & 8.01     & 19.02     & 0.75     & 0.39      & 131.49     \\
Model A                  & 11.05      & 0.49     & 156.51       & 707.57       & 8.61     & 24.12     & 0.67     & 0.32      & 86.69     \\
Model B                  & 9.01      & 0.37     & 355.99       & 969.89       & 9.04     & 9.83     & 0.04     & 1.02      & 253.11     \\
GDP-$x_t$                   & 7.32  & 0.57 & 238.92 & 364.15 & 8.26 & 19.36 & 0.65 & 0.30 & 63.89 \\ \midrule[1pt]
GDP-$x_0$                   & \textbf{13.93} & \textbf{0.63} & \textbf{75.16}  & \textbf{110.39} & \textbf{6.47} & \textbf{24.88} & \textbf{0.86} & \textbf{0.13} & \textbf{50.05} \\ \bottomrule[1.5pt]
\end{tabular}%
}
\vspace{-2.4em}
\label{ablation2}
\end{table}

\noindent\textbf{The Effectiveness of the Trainable Degradation and the Patch-based Tactic.} Moreover, to validate the influence of trainable parameters of the degradation model and our patch-based methods, further experiments are carried out on the LOL~\cite{wei2018deep} and NTIRE~\cite{perez2021ntire} datasets. 
% Model A recovers the images in $256 \times N$ or $ N \times 256 $ sizes and interpolates them by the nearest neighbor to the original size. 
Model A is devised to naively restore the images from patches and patches where the parameters are not related. ModelB is designed with fixed parameters for all patches in the images. 
As shown in Table \ref{ablation2}, our GDP-$x_0$ ranks first across all models and obtains the best visualization results (Fig.~\ref{figure-hdr} and Appendix), revealing the strength of our proposed hierarchical guidance and patch-based method.

\section{Conclusion}

In this paper, we propose the Generative Diffusion Prior for unified image restoration that can be employed to tackle the linear inverse, non-linear and blind problems. Our \GDP is able to restore any-size images via hierarchical guidance and patch-based methods. We systematically studied the way of guidance to exploit the strength of the DDPM. The \GDP is comprehensively utilized on various tasks such as super-resolution, deblurring, inpainting, colorization, low-light enhancement, and HDR recovery, demonstrating the capabilities of \GDP on unified image restoration.

\noindent\textbf{Acknowledgement.} This project is funded in part by Shanghai AI Laboratory

%%%%%%%%% REFERENCES
{\small
\bibliographystyle{ieee_fullname}
\bibliography{arxiv}
}

\appendix

\twocolumn[{%
\renewcommand\twocolumn[1][]{#1}%
\maketitle
\begin{center}
    \centering
    % \vspace{-8pt}
    % \setlength{\abovecaptionskip}{0.15cm}
    \captionsetup{type=figure}
    \includegraphics[width=\linewidth]{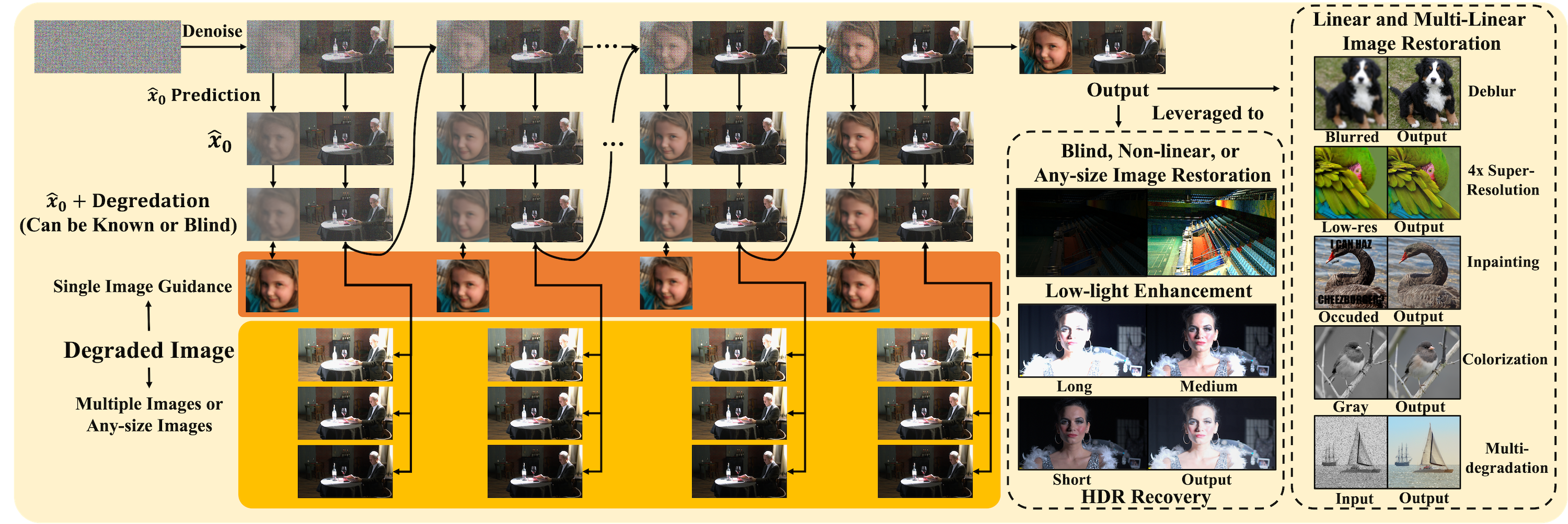}
    \captionof{figure}{\textbf{Illustration of our GDP method for unified image recovery}, including linear inverse problems (Deblurring, $4\times$ super-resolution, inpainting, and colorization), multi-degradation ($i. e.$ Colorization + inpainting), 
    non-linear and blind problems (Low-light enhancement and HDR recovery).
    Note that GDP can restore images of arbitrary sizes, and can accept multiple low-quality images as guidance as in the case of HDR recovery.
    GDP fulfills all the tasks using a single unconditional DDPM pre-trained on ImageNet. 
    % or multi-image guidance and any-size image generation (HDR recovery). 
    % The unsupervised DDPM models can be leveraged as a good solution to the GDP objective towards any image restoration problem.
    }
\label{fig:overview}
\end{center}%
}]

\begin{figure}[htbp]
  \centering
   \includegraphics[width=\linewidth]{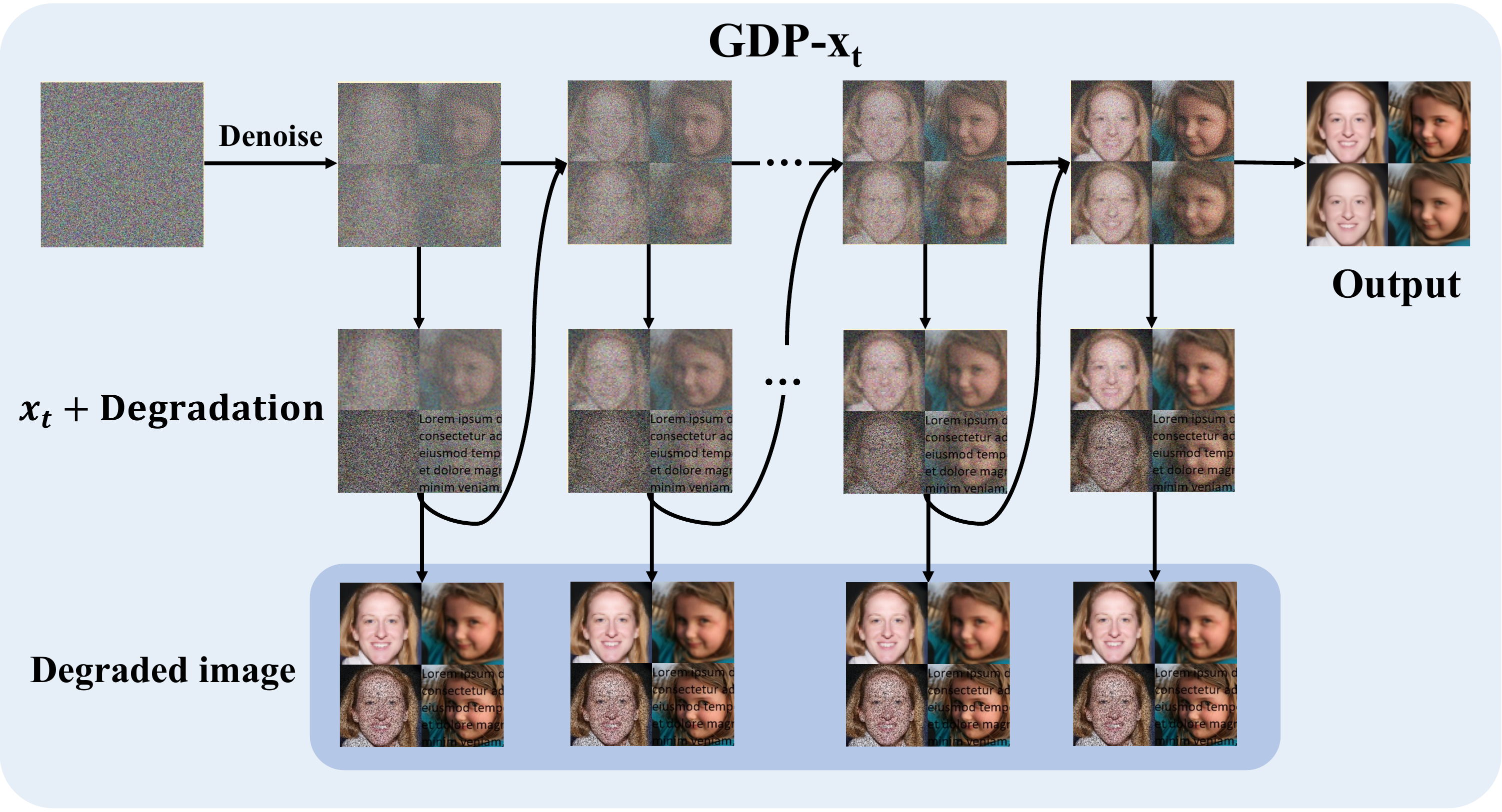}
   \caption{\textbf{Overview of the GDP-$\boldsymbol{x}_t$.} The guidance will be added on the noisy image $\boldsymbol{x}_t$ in every time step.}
% \vspace{-0.5cm}
\label{fig:GDM-t}
\end{figure}

\begin{figure}[htbp]
  \centering
   \includegraphics[width=\linewidth]{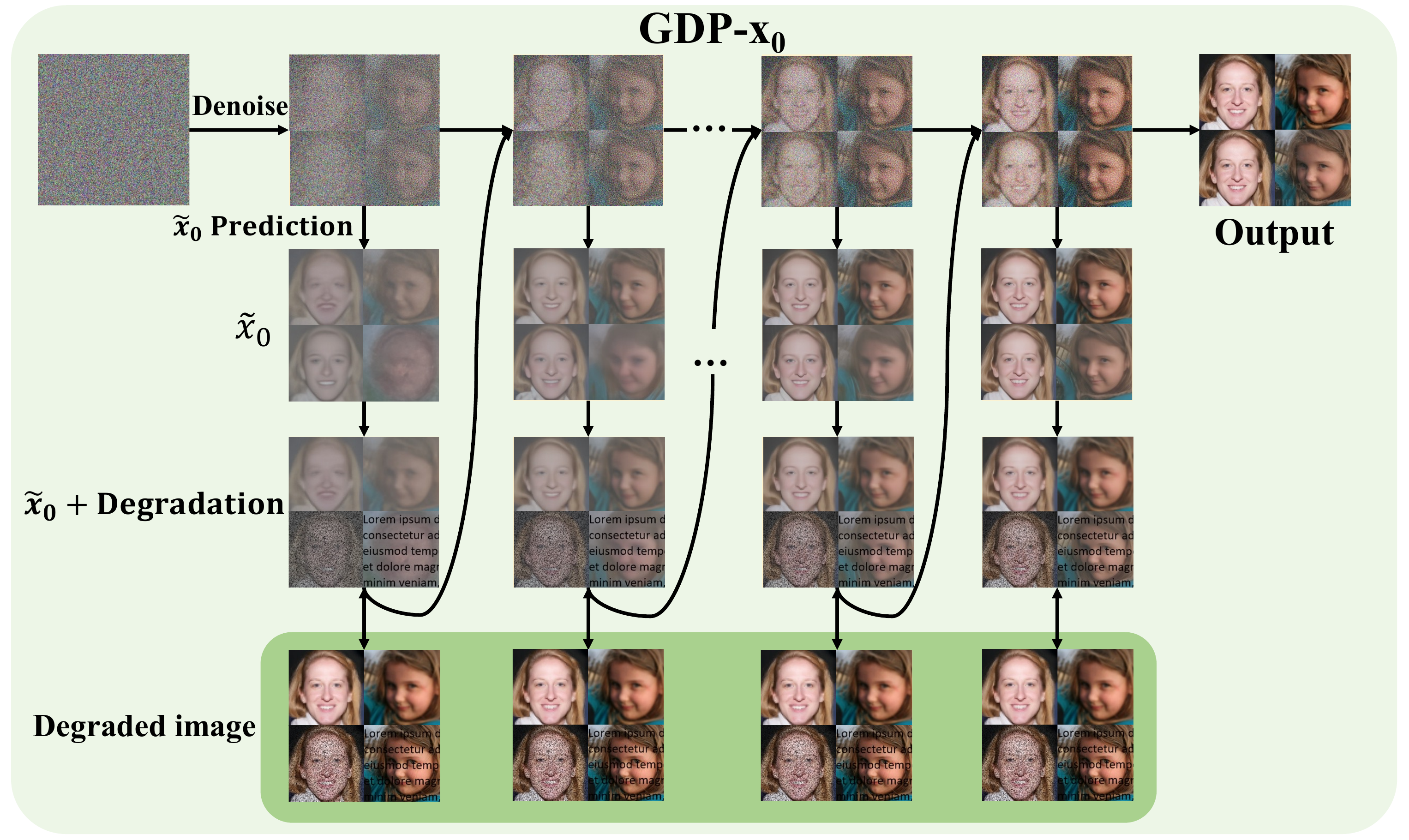}
   \caption{\textbf{Overview of the GDP-$\boldsymbol{x}_0$.} The guidance will be applied to a clean image $\tilde{\boldsymbol{x}}_0$ predicted from the noisy image $\boldsymbol{x}_t$.}
% \vspace{-0.5cm}
\label{fig:GDM-0}
\end{figure}

\begin{figure}[htbp]
  \centering
   \includegraphics[width=\linewidth]{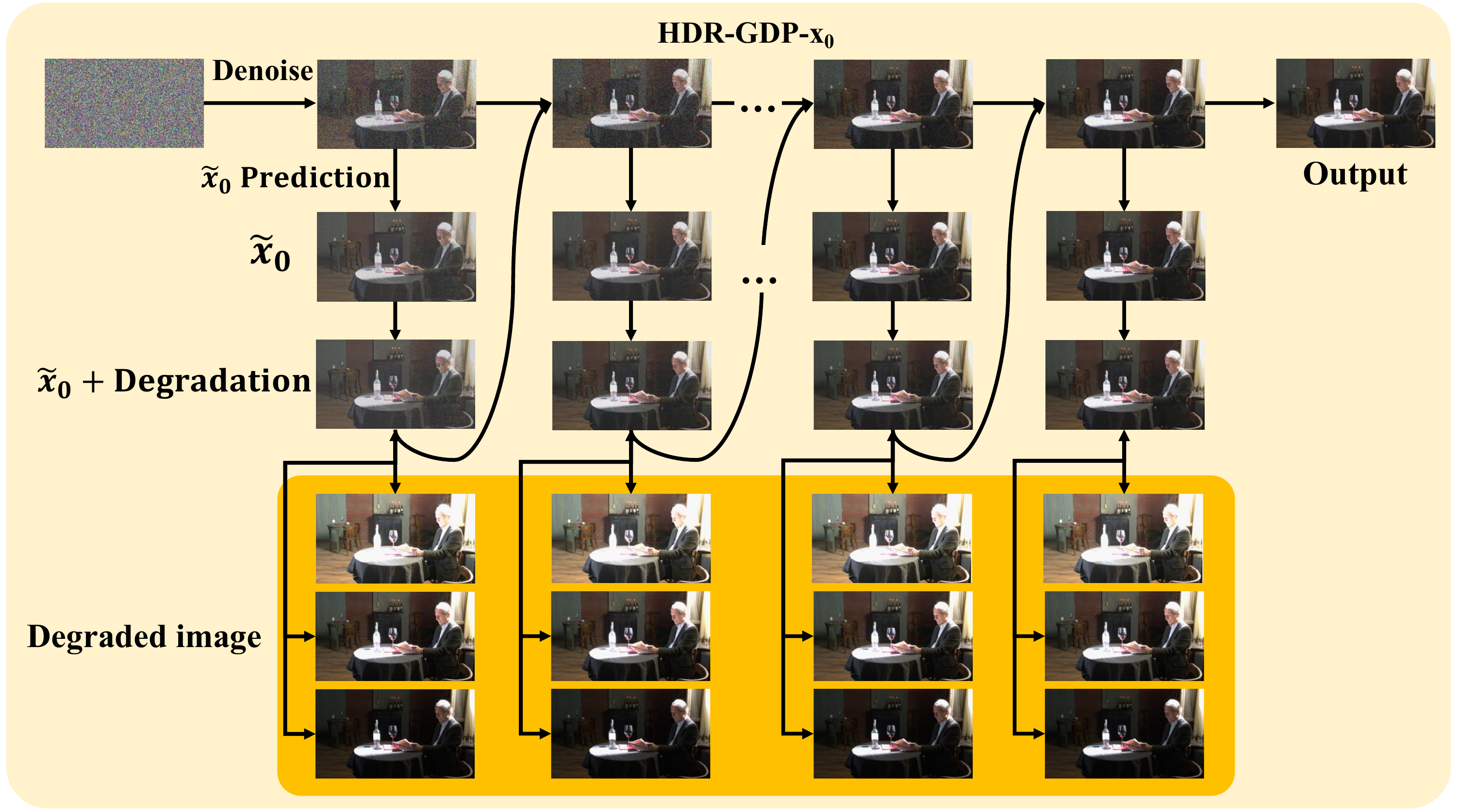}
   \caption{\textbf{Overview of the HDR-GDP-$\boldsymbol{x}_0$.} The guidance will also be applied to a clean image $\tilde{\boldsymbol{x}}_0$. Unlike the GDP-$x_0$, three degraded images are utilized to guide the reverse process, and three sets of degradation models are optimized along the reverse process.}
% \vspace{-0.5cm}
\label{fig:HDR-GDM-0}
\end{figure}

% !TEX root = ../PaperForReview.tex

\begin{algorithm}[t]
	\caption{\textbf{GDP-}$\boldsymbol{x}_t$: Conditioner guided diffusion sampling on $\boldsymbol{x}_{t}$, given a diffusion model $\left(\mu_{\theta}\left(\boldsymbol{x}_{t}\right), \Sigma_{\theta}\left(\boldsymbol{x}_{t}\right)\right)$, corrupted image conditioner $\boldsymbol{y}$. 
 }
	\KwIn{Corrupted image $\boldsymbol{y}$, gradient scale $s$, degradation model $\mathcal{D}_\phi$ with randomly initiated parameters $\phi$, learning rate $l$ for optimizable degradation model, distance measure $\mathcal{L}$.}
	\KwOut{Output image $\boldsymbol{x}_{0}$ conditioned on $\boldsymbol{y}$}
        Sample $\boldsymbol{x}_{T}$ from $\mathcal{N}(0, \mathbf{I})$
        
	\For{$t$ from $T$ to 1}{

	    $\mu, \Sigma = \mu_{\theta}\left(\boldsymbol{x}_{t}\right), \Sigma_{\theta}\left(\boldsymbol{x}_{t}\right)$
     
        $\mathcal{L}^{total}_{\phi, \boldsymbol{x}_{t}} = \mathcal{L}(\boldsymbol{y}, {\mathcal{D}_\phi}\left(\boldsymbol{x}_{t}\right)) + \mathcal{Q}\left(\boldsymbol{x}_{t}\right)$ 
        
        $\phi \leftarrow \phi - l \nabla_{\phi} \mathcal{L}^{total}_{\phi, \boldsymbol{x}_{t}}$

	   % $\boldsymbol{{x}}_{t} \leftarrow \boldsymbol{x}_{t} -s\nabla_{\boldsymbol{{x}}_{t}} \mathcal{L}^{total}_{\phi, \boldsymbol{x}_{t}}$ \\
  
	   Sample $\boldsymbol{x}_{t-1}$ by $\mathcal{N}\left(\mu+s\nabla_{\boldsymbol{{x}}_{t}} \mathcal{L}^{total}_{\phi, \boldsymbol{x}_{t}}, \Sigma\right)$

	}
	\Return $\boldsymbol{x}_{0}$
\label{algo1}
\end{algorithm}

% !TEX root = ../PaperForReview.tex

\begin{algorithm}[t]
	\caption{\textbf{GDP-$\boldsymbol{x}_0$}: Conditioner guided diffusion sampling on $\boldsymbol{\tilde{x}}_{0}$, given a diffusion model $\left(\mu_{\theta}\left(\boldsymbol{x}_{t}\right), \Sigma_{\theta}\left(\boldsymbol{x}_{t}\right)\right)$, corrupted image conditioner $\boldsymbol{y}$.}
	\KwIn{Corrupted image $\boldsymbol{y}$, gradient scale $s$, degradation model $\mathcal{D}$, distance measure $\mathcal{L}$.}
	\KwOut{Output image $\boldsymbol{x}_{0}$ conditioned on $\boldsymbol{y}$}
        Sample $\boldsymbol{x}_{T}$ from $\mathcal{N}(0, \mathbf{I})$
        
	\For{$t$ from $T$ to 1}{
	    $\mu, \Sigma = \mu_{\theta}\left(\boldsymbol{x}_{t}\right), \Sigma_{\theta}\left(\boldsymbol{x}_{t}\right)$
     
	    $\boldsymbol{\tilde{x}}_{0} =  \frac{\boldsymbol{x}_{t}}{\sqrt{\bar{\alpha}_{t}}}-\frac{\sqrt{1-\bar{\alpha}_{t}} \epsilon_{\theta}\left(\boldsymbol{x}_{t}, t\right)}{\sqrt{\bar{\alpha}_{t}}}$
   
	    % $\boldsymbol{\tilde{x}}_{0} \leftarrow \boldsymbol{\tilde{x}}_{0} -s\nabla_{\boldsymbol{\tilde{x}}_{0}} \mathcal{L}\left(\boldsymbol{y}, \left({\mathcal{D}}\left(\boldsymbol{\tilde{x}}_{0}\right)\right)\right)$\\

        $\mathcal{L}^{total}_{ \boldsymbol{\tilde{x}}_{0}} = \mathcal{L}(\boldsymbol{y}, {\mathcal{D}}\left(\boldsymbol{\tilde{x}}_{0}\right)) + \mathcal{Q}\left(\boldsymbol{\tilde{x}}_{0}\right)$ 
  
        % $\boldsymbol{\tilde{x}}_{0} \leftarrow \boldsymbol{\tilde{x}}_{0} -s\nabla_{\boldsymbol{\tilde{x}}_{0}} \mathcal{L}^{total}_{ \boldsymbol{\tilde{x}}_{0}}$ 

        Sample $\boldsymbol{x}_{t-1}$ by $\mathcal{N}\left(\mu+s\nabla_{\boldsymbol{\tilde{x}}_{0}} \mathcal{L}^{total}_{ \boldsymbol{\tilde{x}}_{0}}, \Sigma\right)$
    
        % Sample $\boldsymbol{x}_{t-1}$ by $q\left(\boldsymbol{x}_{t-1} \mid \boldsymbol{x}_{t}, \boldsymbol{\tilde{x}}_{0}\right) =\mathcal{N}\left(\boldsymbol{x}_{t-1}; \tilde{\boldsymbol{\mu}}_t\left(\boldsymbol{x}_{t}, \boldsymbol{\tilde{x}}_{0}\right), \tilde{\beta}_t \mathbf{I}\right)$, 
        
        % where $\quad \tilde{\boldsymbol{\mu}}_t\left(\boldsymbol{x}_{t}, \boldsymbol{\tilde{x}}_{0}\right)=\frac{\sqrt{\bar{\alpha}_{t-1}} \beta_t}{1-\bar{\alpha}_t} \boldsymbol{\tilde{x}}_{0}+\frac{\sqrt{\alpha_t}\left(1-\bar{\alpha}_{t-1}\right)}{1-\bar{\alpha}_t} \boldsymbol{x}_{t} \quad$ and $\quad \tilde{\beta}_t=\frac{1-\bar{\alpha}_{t-1}}{1-\bar{\alpha}_t} \beta_t$
	}
	\Return $\boldsymbol{x}_{0}$
\label{algo6}
\end{algorithm}

% !TEX root = ../PaperForReview.tex

\begin{algorithm}[t]
	\caption{\textbf{GDP-}$\boldsymbol{x}_0$: Conditioner guided diffusion sampling on $\boldsymbol{x}_{0}$, given a diffusion model $\left(\mu_{\theta}\left(\boldsymbol{x}_{t}\right), \Sigma_{\theta}\left(\boldsymbol{x}_{t}\right)\right)$, corrupted images conditioner $\{\boldsymbol{y}^i \mid i=1,2,\ldots, n\}$.}
	\KwIn{Corrupted image $\{\boldsymbol{y}^i \mid i=1,2,\ldots, n\}$ ($n$ = 3 for HDR recovery (LDR-long image $\boldsymbol{y}^1$, LDR-medium image $\boldsymbol{y}^2$, LDR-short image $\boldsymbol{y}^3$) and $n$ = 1 for other tasks), gradient scale $s$, degradation models $\{\mathcal{D}_{\phi^i} | i=1,2,\dots,n\}$ with randomly initiated parameters $\{{\phi^i} | i=1,2,\dots,n\}$, learning rate $l$ for optimizable degradation model, distance measure $\mathcal{L}$.}
	\KwOut{Output image $\boldsymbol{x}_{0}$ conditioned on $\{\boldsymbol{y}^i \mid i=1,2,\ldots,n\}$}
        Sample $\boldsymbol{x}_{T}$ from $\mathcal{N}(0, \mathbf{I})$
        
	\For{$t$ from $T$ to 1}{
	    $\mu, \Sigma = \mu_{\theta}\left(\boldsymbol{x}_{t}\right), \Sigma_{\theta}\left(\boldsymbol{x}_{t}\right)$
	    
	    $\boldsymbol{\tilde{x}}_{0}  =  \frac{\boldsymbol{x}_{t}}{\sqrt{\bar{\alpha}_{t}}}-\frac{\sqrt{1-\bar{\alpha}_{t}} \epsilon_{\theta}\left(\boldsymbol{x}_{t}, t\right)}{\sqrt{\bar{\alpha}_{t}}}$

        $\mathcal{L}^{total}_{\phi, \boldsymbol{\tilde{x}}_{0}} = 0$
	    
     \For{$j$ from 1 to n}{

        $\mathcal{L}_{{\phi^{j}}, \boldsymbol{\tilde{x}}_{0}} = \mathcal{L}(\boldsymbol{y}^j, {\mathcal{D}_{\phi^j}}\left(\boldsymbol{\tilde{x}}_{0}\right)) + \mathcal{Q}\left(\boldsymbol{\tilde{x}}_{0}\right)$

        ${\phi^j} = {\phi^j} - l \nabla_{{\phi^j}} \mathcal{L}_{{\phi^{j}}, \boldsymbol{\tilde{x}}_{0}}$
                
        $\mathcal{L}^{total}_{\phi, \boldsymbol{\tilde{x}}_{0}} = \mathcal{L}^{total}_{\phi, \boldsymbol{\tilde{x}}_{0}} + \mathcal{L}_{\phi^{j}, \boldsymbol{\tilde{x}}_{0}}$ 
	    }

        % $\boldsymbol{\tilde{x}}_{0} = \boldsymbol{\tilde{x}}_{0} - s\nabla_{\boldsymbol{\tilde{x}}_{0}} \mathcal{L}^{total}_{\phi, \boldsymbol{\tilde{x}}_{0}}$

        % Sample $\boldsymbol{x}_{t-1}$ by $q\left(\boldsymbol{x}_{t-1} \mid \boldsymbol{x}_{t}, \boldsymbol{\tilde{x}}_{0}\right)=\mathcal{N}\left(\boldsymbol{x}_{t-1}; \tilde{\boldsymbol{\mu}}_t\left(\boldsymbol{x}_{t}, \boldsymbol{\tilde{x}}_{0}\right), \tilde{\beta}_t \mathbf{I}\right)$, 
        
        % where $\quad \tilde{\boldsymbol{\mu}}_t\left(\boldsymbol{x}_{t}, \boldsymbol{\tilde{x}}_{0}\right)=\frac{\sqrt{\bar{\alpha}_{t-1}} \beta_t}{1-\bar{\alpha}_t} \boldsymbol{\tilde{x}}_{0}+\frac{\sqrt{\alpha_t}\left(1-\bar{\alpha}_{t-1}\right)}{1-\bar{\alpha}_t} \boldsymbol{x}_{t} \quad$ and $\quad \tilde{\beta}_t=\frac{1-\bar{\alpha}_{t-1}}{1-\bar{\alpha}_t} \beta_t$
        
        Sample $\boldsymbol{x}_{t-1}$ by $\mathcal{N}\left(\mu+s\nabla_{\boldsymbol{\tilde{x}}_{0}} \mathcal{L}^{total}_{\phi, \boldsymbol{\tilde{x}}_{0}}, \Sigma\right)$
	}
	\Return $\boldsymbol{x}_{0}$
\label{algo4}
\end{algorithm}

% \input{cvpr2023-author_kit-v1_1-1/latex/algos/algo3.tex}

% !TEX root = ../PaperForReview.tex

\begin{algorithm}[t]\footnotesize
    \caption{Restore Any-size Image}
    \KwIn{Conditioner guided diffusion sampling on $\boldsymbol{\tilde{x}}_{0}$, given a diffusion model $\left(\mu_{\theta}\left(\boldsymbol{x}_{t}\right), \Sigma_{\theta}\left(\boldsymbol{x}_{t}\right)\right)$, corrupted image conditioner $\boldsymbol{y}$, degradation model $\mathcal{D}_{\phi}: \boldsymbol{y}=f \boldsymbol{x}+\boldsymbol{\mathcal{M}}$  with randomly initiated parameters $\phi$, learning rate $l$ for optimizable degradation model. Dictionary of $K$ overlapping patch locations, and a binary patch mask $\mathbf{P}^k$.}
    \KwOut{Output image $\boldsymbol{x}_{0}$ conditioned on $\boldsymbol{y}$}

    Sample $\boldsymbol{x}_{T}$ from $\mathcal{N}(0, \mathbf{I})$
    
    \For{$t$ from $T$ to 1}{

    $\mu, \Sigma = \mu_{\theta}\left(\boldsymbol{x}_{t}\right), \Sigma_{\theta}\left(\boldsymbol{x}_{t}\right)$

    $\text {Mean vector 
    }\boldsymbol{\Omega}_t=\mathbf{0} \text { and variance vector } \boldsymbol{\psi}_t=\mathbf{0} \text { and weight vector } \mathbf{G}=\mathbf{0} \text { and } f=\mathbf{0} \text { and } \boldsymbol{\mathcal{M}}=\mathbf{0} $
    
    \For{$k=1,\ldots,K$}{
    $\boldsymbol{x}_t^{k}=\operatorname{Crop}\left(\mathbf{P}^k \circ \boldsymbol{x}_t\right) $
    
    $ \boldsymbol{y}^{k}=\operatorname{Crop}\left(\mathbf{P}^k \circ \boldsymbol{y}\right) $

    $ \boldsymbol{\mathcal{M}}^{k}=\operatorname{Crop}\left(\mathbf{P}^k \circ \boldsymbol{\mathcal{M}}\right) $

    $\boldsymbol{\tilde{x}}_{0}^{k} =  \frac{\boldsymbol{x}_{t}^{k}}{\sqrt{\bar{\alpha}_{t}}}-\frac{\sqrt{1-\bar{\alpha}_{t}} \epsilon_{\theta}\left(\boldsymbol{x}_{t}^{k}, t\right)}{\sqrt{\bar{\alpha}_{t}}}$

    $\mathcal{L}^{total}_{\phi, \boldsymbol{\tilde{x}}_{0}^{k}} = \mathcal{L}(\boldsymbol{y}^{k}, {\mathcal{D}_\phi}\left(\boldsymbol{\tilde{x}}_{0}^{k}\right)) + \mathcal{Q}\left(\boldsymbol{\tilde{x}}_{0}^{k}\right)$ 

     $f^{k} \leftarrow f^{k} - l \nabla_{f^{k}} \mathcal{L}^{total}_{f^{k}, \boldsymbol{\tilde{x}}_{0}^{k}}$

     $\boldsymbol{\mathcal{M}}^{k} \leftarrow \boldsymbol{\mathcal{M}}^{k} - l \nabla_{\boldsymbol{\mathcal{M}}^{k}} \mathcal{L}^{total}_{\boldsymbol{\mathcal{M}}^{k}, \boldsymbol{\tilde{x}}_{0}^{k}}$

    $\mu^{k} = \mu+s\nabla_{\boldsymbol{\tilde{x}}_{0}^{k}} \mathcal{L}^{total}_{\phi, \boldsymbol{\tilde{x}}_{0}^{k}}$

    $f = f + f^{k}$
    
    $\boldsymbol{\Omega}_t=\mathbf{\Omega}_t+\mathbf{P}_k \cdot \mu^{k} $

    $\boldsymbol{\psi}_t=\mathbf{\psi}_t+\mathbf{P}^k \cdot \sigma^k $

    $\boldsymbol{\mathcal{M}} =\boldsymbol{\mathcal{M}} +\mathbf{P}^k \cdot \boldsymbol{\mathcal{M}}^{k} $
    
    $ \mathbf{G}=\mathbf{G}+\mathbf{P}^k$
    
    }
    
    $\mathbf{\Omega}_t=\mathbf{\Omega}_t \oslash \mathbf{G} \quad \quad / / \oslash: \text { element-wise division } $

    $\mathbf{\psi}_t=\mathbf{\psi}_t \oslash \mathbf{G}$

    $\boldsymbol{\mathcal{M}}=\boldsymbol{\mathcal{M}} \oslash \mathbf{G}$

    $f = f/K$
    
    Sample $\boldsymbol{x}_{t-1}$ by $\mathcal{N}\left(\mathbf{\Omega}_t, \mathbf{\psi}_t \right)$

    }  
    \Return Restored any-size image $\boldsymbol{x}_0$
\label{algo:patch}
\end{algorithm}

%%%%%%%%% BODY TEXT - ENTER YOUR RESPONSE BELOW

\section{Limitations and Future works}

\noindent\textbf{Limitations.}
The main limitation of our work is its inference time. Since we might add several guidance steps in every time step $t$, the sampling time is extended. This limits the applicability of our method to real-time applications and weak end-user devices such as mobile devices. To address this issue, further research into accelerated diffusion sampling techniques is required.

In addition, the choice of the guidance scale is also obtained through experiments, which means that for samples with different distributions, it is necessary to manually select the optimal guidance scale. However, we found that for the same distribution of data, an approximate degradation model may lead to close guidance scales. This phenomenon may be proved mathematically in future work.

\noindent\textbf{Future works.}
In future work, in addition to further optimizing the time step and variance schedules, it would be interesting to investigate the following:

(i) The Guided Diffusion Prior can also theoretically be applied to 3D data restoration. For instance, point cloud completion and upsampling can be regarded as linear inverse problems in 3D vision. Shapeinversion~\cite{zhang2021unsupervised} tackles the point cloud completion by GAN inversion, where the GDP can hopefully be integrated. 

(ii) Moreover, since LiDAR is affected by various kinds of weather in the real world and also produces various non-linear degradations, GDP should also be explored for the recovery of these point clouds.

(iii) Self-supervised training techniques inspired by our GDP and techniques used in supervised techniques~\cite{saharia2022palette} that further improve the performance of unsupervised image restoration models.

\section{Implementation Details}

We apply GDP to a suite of challenging image restoration tasks: (1) \textbf{Colorization} transforms an input gray-scale image to a plausible color image. (2) \textbf{Inpainting} fills in user-specified masked regions of an image with realistic content. (3) \textbf{Super-resolution} extends a low-resolution image into a higher one. (4) \textbf{Deblurring} corrects the blurred images, restoring plausible image detail. (5) \textbf{Enlighting} enables the dark images turned into normal images. (6) \textbf{HDR image recovery} aims to obtain HDR images with the aid of three LDR images.  Inputs and outputs of the first four tasks are represented as $256 \times 256$ RGB images, while the last two tasks are various ($1900 \times 1060$ for HDR image recovery and $600 \times 400$ for image enlightening, respectively). We do so without task-specific hyperparameter tuning and architecture customization.

Colorization requires the representation of objects, segmentation, and layouts with long-range image dependencies.
Inpainting is challenging due to large masks, image diversity, and cluttered scenes.
Super-resolution and deblurring are also not trivial because the degradation might damage the content of the images.
While the other tasks are linear in nature, low-light enhancement and HDR recovery are non-linear inverse problems; they require a good model of natural image statistics to detect and correct over-exposed and under-exposed areas.
Although previous works have studied these problems extensively, it is rare that a model with no task-specific engineering achieves strong performance in all tasks, beating strong task-specific GAN and regression baselines. Our GDP is devised to achieve this goal.

% GDP uses two ways to guide the conditional diffusion models.
\begin{figure*}[t]
  \centering
   \includegraphics[width=\linewidth]{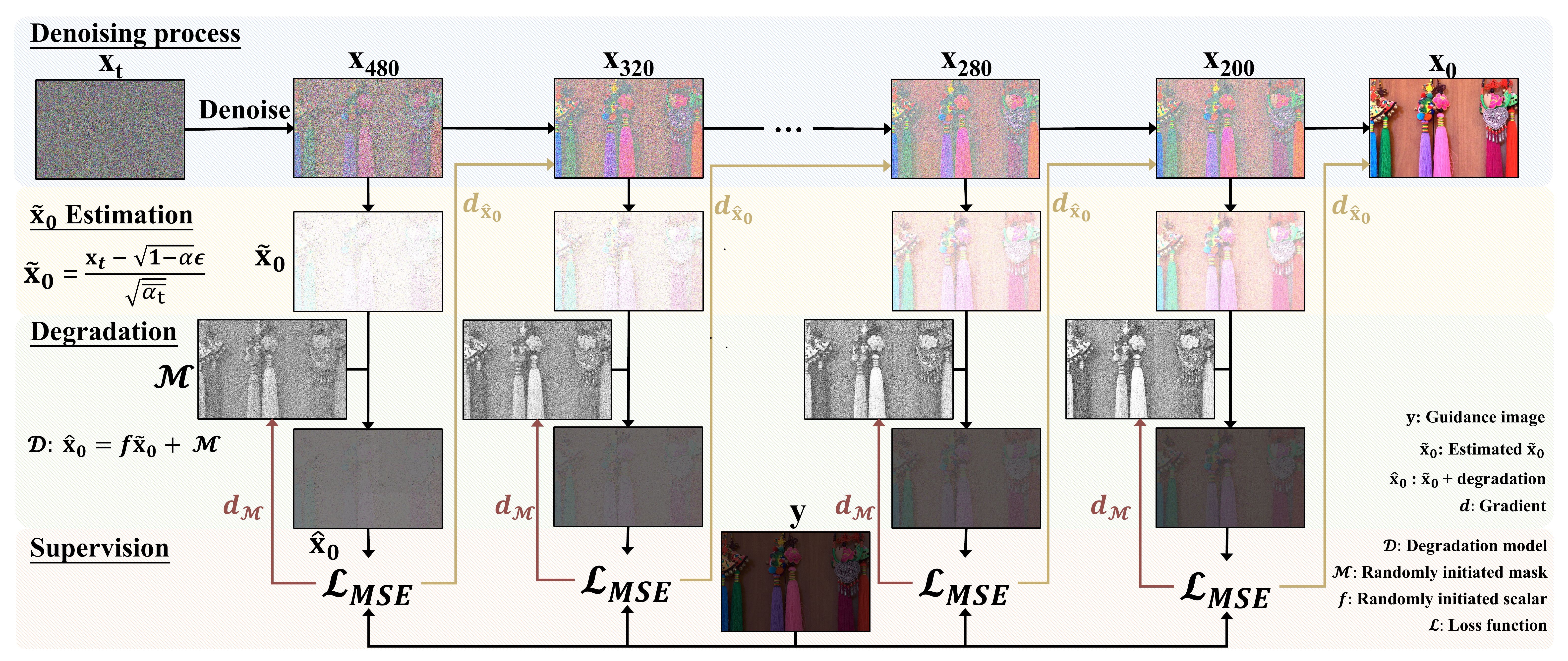}
   \caption{Illustration of the patch-based method for any-size image restoration.}
\label{SM:hierarchial}
\end{figure*}

\begin{figure*}[t]
  \centering
   \includegraphics[width=\linewidth]{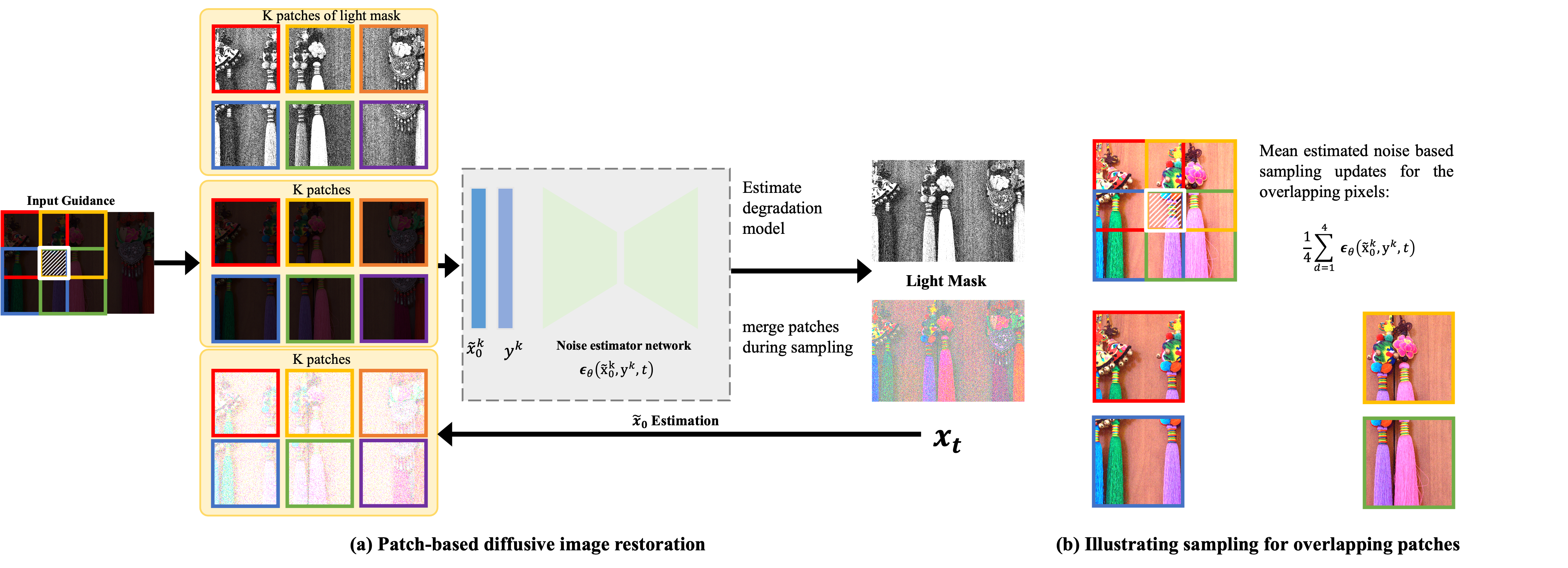}
   \caption{(a) Illustration of the patch-based image restoration pipeline detailed in Algorithm ~\ref{algo:patch}. (b) Illustrating mean estimated noise-guided sampling updates for overlapping pixels across patches. We demonstrate a simplified example where $r = p/2$, $r$ is the stride and $p$ is the patch size of images. And there are only four overlapping patches sharing the grid cell marked with the white border and gratings.
   The pixels in this region would be updated at each denoising step $t$ using the mean estimated noise over the four overlapping patches.
   }
   % \caption{\textbf{Overview of the hierarchical guidance patch-based methods.} Given a low-light image or LDR image, we firstly resize it to $\boldsymbol{\overline{y}} \in \mathbb{R}^{3 \times 256 \times \overline{W} \text{ or } 3 \times \overline{H} \times 256}$, then apply the patch-based methods on the reshaped images. After, the light masks $\boldsymbol{\mathcal{\overline{M}}}$ are interpolated to the original image size to obtain the $\boldsymbol{\mathcal{M}}$, representing the global light shift. Following that, the light factor $\boldsymbol{f}$ and the light mask $\boldsymbol{\mathcal{M}}$ will be fixed and utilized to generate the image patches, which will be recombined as the output images.}
\label{SM:hierarchial1}
\end{figure*}

% \subsection{Implementation}
\subsection{Dataset briefs}

\textbf{ImageNet, LSUN, CelebA, and USC-SIPI Datasets.} To quantitatively evaluate GDP on linear image restoration tasks, we test on 1k images from the ImageNet validation set following~\cite{pan2021exploiting}. The CelebA-HQ~\cite{lee2020maskgan} dataset is a high-quality subset of the Large-Scale CelebFaces Attributes (CelebA) dataset~\cite{liu2015deep}. LSUN dataset~\cite{yu2015lsun} contains around one million labeled images for each of 10 scene categories and 20 object categories. And the USC-SIPI dataset~\cite{weber2006usc} is a collection of various digitized images. We utilize the images from CelebA, LSUN, and USC-SIPI provided by~\cite{kawar2022denoising}.

\textbf{LOL Dataset.} The LOL dataset~\cite{wei2018deep} is composed of 500 low-light and normal-light image pairs and divided into 485 training pairs and 15 testing pairs. The low-light images contain noise produced during the photo capture process. Most of the images are indoor scenes. All the images have a resolution of 400 $\times$ 600.

\textbf{VE-LOL-L Dataset.} For underexposure correction experiments, we use the paired data of the VE-LOL-L dataset~\cite{liu2021benchmarking}, in which each captured well-exposed image has its underexposed version with different underexposure levels. Note that the VE-LOL-L dataset, consisting of VE-LOL-Cap and VE-LOL-Syn, is also carried out. Due to the different distribution of the two sub-set, we solve them under different guidance scales.

\textbf{LoLi-Phone Dataset.} LoLi-Phone~\cite{li2021low} is a large-scale low-light image and video dataset for low-light image enhancement. The images and videos are taken by different mobile phone cameras under diverse illumination conditions.

\textbf{NTIRE Dataset~\cite{lugmayr2021ntire}.} In the NTIRE dataset, there are 1494 LDRs/HDR for training, 60 images for validation, and 201 images for testing. The 1494 frames consist of 26 long shots. Each scene contains three LDR images, their corresponding exposure and alignment information, and HDR ground truth. The size of an image is $1060 \times 1900$. Since the ground truth of the validation and test sets are not available, we only do experiments on the training set. We select 100 images as the test set.

\subsection{Experimental Setup}

In each inverse problem, the pixel values are in the range [0,1], and the resulting degradation measures are as follows: (i) For super-resolution, a block averaging filter is utilized to downscale the image on each axis 4 times; (ii) In terms of deblurring, the image is blurred by a 9$\times 9$ unified kernel. (iii) For colorization, the gray-scale image is the average of the red, green, and blue channels of the original image; (iv) For inpainting, we cover parts of the original image with text overlays or randomly delete 25$\%$ pixels.

In the non-linear and blind problem, the images from the low-light dataset and NTIRE dataset are naturally over-exposed or under-exposed. Therefore, no additional operations are required for the images.

\section{Evaluation Metrics}

Apart from the commonly used PSNR and SSIM, other metrics are also utilized for evaluation: 
(i) \textbf{FID}~\cite{heusel2017gans} is an objective metric used to assess the quality of synthesized images. 
(ii) \textbf{Consistency}~\cite{saharia2022image} measures MSE between the degraded inputs and the outputs undergoing the same degradation. 
(iii) \textbf{Learned perceptual image patch similarity (LPIPS)}~\cite{zhang2018unreasonable} is also adopted, a deep feature-based perceptual distance metric to further assess the image quality. 
(iv) The non-reference \textbf{perceptual index (PI)}~\cite{mittal2012making} is also employed to evaluate perceptual quality. The PI metric is originally utilized to measure perceptual quality in image super-resolution. It has also been used to assess the performance of other image restoration tasks. A lower PI value indicates better perceptual quality.
(v) The \textbf{lightness order error (LOE)}~\cite{wang2013naturalness} is employed as our objective metric to measure the performance. The definition of LOE is as follows:
\begin{equation}
    L O E=\frac{1}{m} \sum_{x=1}^m \sum_{y=1}^m\left(U(\mathbf{T}(x), \mathbf{T}(y)) \oplus U\left(\mathbf{T}_r(x), \mathbf{T}_r(y)\right)\right)
\end{equation}
where $m$ is the pixel number. The function $U(p, q)$ returns 1 if $p>=q, 0$ otherwise. $\oplus$ stands for the exclusive-or operator. In addition, $\mathbf{T}(x)$ and $\mathbf{T}_r(x)$ are the maximum values among $\mathrm{R}, \mathrm{G}$ and $\mathrm{B}$ channels at location $x$ of the enhanced and reference images, respectively. The lower the LOE is, the better the enhancement preserves the naturalness of lightness.

\section{Further elaboration of the models}

\noindent\textbf{GDP-$\boldsymbol{x}_t$.} As shown in Fig.~\ref{fig:GDM-t}, the guidance is conditioned on $\boldsymbol{x}_t$ but with the absence of $\Sigma$. The noisy images are gradually denoised during the reverse process. And the $\boldsymbol{x}_t$ undergoing the degradation model is more similar to the corrupted image. The gradients $\nabla$ of the loss function are utilized to control the mean of the conditional distribution.

\noindent\textbf{GDP-$\boldsymbol{x}_0$.} To make a clear comparison, we also illustrate the GDP-$\boldsymbol{x}_0$ in Fig.~\ref{fig:GDM-0}, and Algorithm 2 in the main paper. Different from the GDP-$\boldsymbol{x}_t$, GDP-$\boldsymbol{x}_0$ will predict the intermediate variance $\boldsymbol{\tilde{x}}_0$ from the noisy image $\boldsymbol{x}_{t}$ by estimating the noise in $\boldsymbol{x}_{t}$, which can be directly inferred when given the $\boldsymbol{x}_t$ in every time steps $t$. Then the predicted $\boldsymbol{\tilde{x}}_0$ goes through the same degradation as input to obtain $\boldsymbol{\hat{x}}_0$. Note that the degradation might be unknown. Then the loss between the $\boldsymbol{\hat{x}}_0$ and the corrupted image $\boldsymbol{y}$, the gradients will be applied to optimize the unknown degradation models and sample the next step latent $\boldsymbol{x}_{t-1}$.

\noindent\textbf{HDR-GDP-$\boldsymbol{x}_0$.} As depicted in Fig.~\ref{fig:HDR-GDM-0}, and Algorithm~\ref{algo4}, there are three images to guide the reverse process. As a blind problem, we randomly initiate three sets of the parameters of the degradation models. At every time step, $\boldsymbol{\tilde{x}}_0$ will undergo the three degradation models $\mathcal{D}^i$, respectively. Unlike GDP-$\boldsymbol{x}_0$, the gradients of the three losses are used to optimize the corresponding degradation model and all leveraged to sample the next step latent $\boldsymbol{x}_{t-1}$.

\noindent\textbf{Hierarchical Guidance and Patch-based Methods.} As vividly illustrated in Fig.~\ref{SM:hierarchial} and ~\ref{SM:hierarchial1}, we resize the corrupted images $\boldsymbol{y} \in \mathbb{R}^{3 \times H \times W} $ to $\boldsymbol{\overline{y}} \in \mathbb{R}^{3 \times 256 \times \overline{W} \text{ or } 3 \times \overline{H} \times 256}$, then apply the patch-based methods~\cite{ozdenizci2022restoring} on the reshaped images. Following that, the light masks $\boldsymbol{\mathcal{\overline{M}}}$ are interpolated to the original image size to obtain the $\boldsymbol{\mathcal{M}}$, which can be regarded as the global light shift. After, the light factor $\boldsymbol{f}$ and the light mask $\boldsymbol{\mathcal{M}}$ will be fixed and utilized to generate the image patches of the original images, which will be finally recombined as the output images. In our experiments, low-light enhancement and HDR recovery problems can be tackled by this strategy.

% \usepackage{graphicx}
\begin{table*}[t]
\centering
\caption{\textbf{The guidance scales and the number of optimization per time step on the various tasks.} Note that these parameters may not be optimal due to the infinite number of possible combinations.}
{%
\begin{tabular}{l|c|cc}
\toprule[1.5pt]
               Tasks       & Dataset       & Guidance scale & \begin{tabular}[c]{@{}c@{}}The number of optimization\\  per time step\end{tabular} \\ \midrule[1pt]
4$\times$ Super-resolution   & ImageNet~\cite{pan2021exploiting}      & 2E+03              & 6                                                                                   \\
Deblurring            & ImageNet~\cite{pan2021exploiting}      & 6E+03             & 6                                                                                   \\
25$\%$ Inpainting       & ImageNet~\cite{pan2021exploiting}      & 4E+03              & 6                                                                                   \\
Colorization          & ImageNet~\cite{pan2021exploiting}      & 6E+03            & 6                                                                                   \\
Low-light enhancement & LOL dataset~\cite{wei2018deep}   & 1E+05            & 6                                                                                   \\
HDR recovery          & NTIRE dataset~\cite{lugmayr2021ntire} & 1E+05            & 1                                                                                   \\ \bottomrule[1.5pt]
\end{tabular}%
}
\end{table*}

\section{Further Ablation Study on the Guidance}

To gain insight into the way of guidance, apart from GDP-$\boldsymbol{x}_t$ and GDP-$\boldsymbol{x}_0$, two more variants GDP-$\boldsymbol{x}_t$-v1 and GDP-$\boldsymbol{x}_0$-v1 are devised for comparison.

The main difference among these four variants is the way of mean shift. The mean shift of four variants can be written as follow:
\begin{equation}
\begin{aligned}
    &\text{GDP-}\boldsymbol{x}_0: \tilde{\boldsymbol{\mu}}_t\left(\boldsymbol{x}_{t}, \boldsymbol{\tilde{x}}_{0}\right)=\\ & \frac{\sqrt{\bar{\alpha}_{t-1}} \beta_t}{1-\bar{\alpha}_t} \boldsymbol{\tilde{x}}_{0}+\frac{\sqrt{\alpha_t}\left(1-\bar{\alpha}_{t-1}\right)}{1-\bar{\alpha}_t} \boldsymbol{x}_{t} + s\nabla_{\boldsymbol{\tilde{x}}_{0}} \mathcal{L}^{total}_{ \boldsymbol{\tilde{x}}_{0}}
    \\ &  \text{GDP-}\boldsymbol{x}_t: \tilde{\boldsymbol{\mu}}_t\left(\boldsymbol{x}_{t}, \boldsymbol{\tilde{x}}_{0}\right)=\\  &\frac{\sqrt{\bar{\alpha}_{t-1}} \beta_t}{1-\bar{\alpha}_t} \boldsymbol{\tilde{x}}_{0}+\frac{\sqrt{\alpha_t}\left(1-\bar{\alpha}_{t-1}\right)}{1-\bar{\alpha}_t} \boldsymbol{x}_{t} + s\nabla_{\boldsymbol{x}_{t}} \mathcal{L}^{total}_{\boldsymbol{x}_{t}}
    \\ & \text{GDP-}\boldsymbol{x}_0\text{-v1}: \tilde{\boldsymbol{\mu}}_t\left(\boldsymbol{x}_{t}, \boldsymbol{\tilde{x}}_{0}\right)=\\  & \frac{\sqrt{\bar{\alpha}_{t-1}} \beta_t}{1-\bar{\alpha}_t} (\boldsymbol{\tilde{x}}_{0} + s\nabla_{\boldsymbol{\tilde{x}}_{0}} \mathcal{L}^{total}_{ \boldsymbol{\tilde{x}}_{0}})+\frac{\sqrt{\alpha_t}\left(1-\bar{\alpha}_{t-1}\right)}{1-\bar{\alpha}_t} \boldsymbol{x}_{t}
    \\ & \text{GDP-}\boldsymbol{x}_t\text{-v1}: \tilde{\boldsymbol{\mu}}_t\left(\boldsymbol{x}_{t}, \boldsymbol{\tilde{x}}_{0}\right)=\\  & \frac{\sqrt{\bar{\alpha}_{t-1}} \beta_t}{1-\bar{\alpha}_t}\boldsymbol{\tilde{x}}_{0}+\frac{\sqrt{\alpha_t}\left(1-\bar{\alpha}_{t-1}\right)}{1-\bar{\alpha}_t} (\boldsymbol{x}_{t} + s\nabla_{\boldsymbol{x}_{t}} \mathcal{L}^{total}_{\boldsymbol{x}_{t}}).
\end{aligned} 
\label{mean-comparison}
\end{equation}
where GDP-$\boldsymbol{x}_0$ directly add the mean shift $s\nabla_{\boldsymbol{x}_{0}} \mathcal{L}^{total}_{\boldsymbol{x}_{0}}$ into $\tilde{\boldsymbol{\mu}}_t\left(\boldsymbol{x}_{t}, \boldsymbol{\tilde{x}}_{0}\right)$ without the coefficient $\frac{\sqrt{\bar{\alpha}_{t-1}} \beta_t}{1-\bar{\alpha}_t}$, compared with GDP-$\boldsymbol{x}_0$-v1. 

It is experimentally found that GDP-$\boldsymbol{x}_0$ and GDP-$\boldsymbol{x}_t$ fulfills better performance on four linear tasks than GDP-$\boldsymbol{x}_0$-v1 and GDP-$\boldsymbol{x}_t$-v1 in Table.~\ref{tables:guidance}. 

% \usepackage{graphicx}
\begin{table*}[t]
\centering
\caption{The performance of ablation studies on the way of guidance. We compare four ways of guidance in terms of FID.}
{%
\begin{tabular}{l|cccc}
\toprule[1.5pt]
FID     & 4x super-resolution & Deblur & 25$\%$ Inpainting & Colorization \\ \midrule[1pt]
GDP-$\boldsymbol{x}_t$-v1 & 108.06                   & 88.52      & 113.47               & 102.37           \\
GDP-$\boldsymbol{x}_0$-v1 & 44.16               & 10.35      & 37.32           & 41.53        \\
GDP-$\boldsymbol{x}_t$ & 64.67               & 5.00   & 20.24           & 66.43        \\
GDP-$\boldsymbol{x}_0$ & 38.24               & 2.44   & 16.58           & 37.60        \\ \bottomrule[1.5pt]
\end{tabular}%
}
\label{tables:guidance}
\end{table*}

% !TEX root = ../PaperForReview.tex

\begin{algorithm}[t]
	\caption{\textbf{GDP-$\boldsymbol{x}_t$-v1} with fixed degradation model: Conditioner guided diffusion sampling on $\boldsymbol{x}_{t}$, given a diffusion model $\left(\mu_{\theta}\left(\boldsymbol{x}_{t}\right), \Sigma_{\theta}\left(\boldsymbol{x}_{t}\right)\right)$, corrupted image conditioner $\boldsymbol{y}$. 
 % \lyu{why alg 1 has known degradation and alg2 has unknown.}
 }
	\KwIn{Corrupted image $\boldsymbol{y}$, gradient scale $s$, degradation model $\mathcal{D}$, distance measure $\mathcal{L}$, optional quality enhancement loss $\mathcal{Q}$, quality enhancement scale $\lambda$.}
	\KwOut{Output image $\boldsymbol{x}_{0}$ conditioned on $\boldsymbol{y}$}
        Sample $\boldsymbol{x}_{T}$ from $\mathcal{N}(0, \mathbf{I})$
        
	\For{$t$ from $T$ to 1}{
	    $\mu, \Sigma = \mu_{\theta}\left(\boldsymbol{x}_{t}\right), \Sigma_{\theta}\left(\boldsymbol{x}_{t}\right)$

        $\boldsymbol{\tilde{x}}_{0} =  \frac{\boldsymbol{x}_{t}}{\sqrt{\bar{\alpha}_{t}}}-\frac{\sqrt{1-\bar{\alpha}_{t}} \epsilon_{\theta}\left(\boldsymbol{x}_{t}, t\right)}{\sqrt{\bar{\alpha}_{t}}}$
        
        $\mathcal{L}^{total}_{ \boldsymbol{{x}}_{t}} = \mathcal{L}(\boldsymbol{y}, {\mathcal{D}}\left(\boldsymbol{{x}}_{t}\right)) + \mathcal{Q}\left(\boldsymbol{{x}}_{t}\right)$ 
  
	   $\boldsymbol{{x}}_{t} \leftarrow \boldsymbol{x}_{t} -s\nabla_{\boldsymbol{{x}}_{t}} \mathcal{L}\left(\boldsymbol{y}, {\mathcal{D}}\left(\boldsymbol{x_{t}}\right)\right)$

        Sample $\boldsymbol{x}_{t-1}$ by $q\left(\boldsymbol{x}_{t-1} \mid \boldsymbol{x}_{t}, \boldsymbol{\tilde{x}}_{0}\right) =\mathcal{N}\left(\boldsymbol{x}_{t-1}; \tilde{\boldsymbol{\mu}}_t\left(\boldsymbol{x}_{t}, \boldsymbol{\tilde{x}}_{0}\right), \tilde{\beta}_t \mathbf{I}\right)$, 
        
        where $\quad \tilde{\boldsymbol{\mu}}_t\left(\boldsymbol{x}_{t}, \boldsymbol{\tilde{x}}_{0}\right)=\frac{\sqrt{\bar{\alpha}_{t-1}} \beta_t}{1-\bar{\alpha}_t} \boldsymbol{\tilde{x}}_{0}+\frac{\sqrt{\alpha_t}\left(1-\bar{\alpha}_{t-1}\right)}{1-\bar{\alpha}_t} \boldsymbol{x}_{t} \quad$ and $\quad \tilde{\beta}_t=\frac{1-\bar{\alpha}_{t-1}}{1-\bar{\alpha}_t} \beta_t$
	}
	\Return $\boldsymbol{x}_{0}$
\label{algo5_v1}
\vspace{-0.1em}
\end{algorithm}

% !TEX root = ../PaperForReview.tex

\begin{algorithm}[t]
	\caption{\textbf{GDP-$\boldsymbol{x}_0$-v1}: Conditioner guided diffusion sampling on $\boldsymbol{\tilde{x}}_{0}$, given a diffusion model $\left(\mu_{\theta}\left(\boldsymbol{x}_{t}\right), \Sigma_{\theta}\left(\boldsymbol{x}_{t}\right)\right)$, corrupted image conditioner $\boldsymbol{y}$.}
	\KwIn{Corrupted image $\boldsymbol{y}$, gradient scale $s$, degradation model $\mathcal{D}$, distance measure $\mathcal{L}$.}
	\KwOut{Output image $\boldsymbol{x}_{0}$ conditioned on $\boldsymbol{y}$}
        Sample $\boldsymbol{x}_{T}$ from $\mathcal{N}(0, \mathbf{I})$
        
	\For{$t$ from $T$ to 1}{
	    $\mu, \Sigma = \mu_{\theta}\left(\boldsymbol{x}_{t}\right), \Sigma_{\theta}\left(\boldsymbol{x}_{t}\right)$
     
	    $\boldsymbol{\tilde{x}}_{0} =  \frac{\boldsymbol{x}_{t}}{\sqrt{\bar{\alpha}_{t}}}-\frac{\sqrt{1-\bar{\alpha}_{t}} \epsilon_{\theta}\left(\boldsymbol{x}_{t}, t\right)}{\sqrt{\bar{\alpha}_{t}}}$
   
	    % $\boldsymbol{\tilde{x}}_{0} \leftarrow \boldsymbol{\tilde{x}}_{0} -s\nabla_{\boldsymbol{\tilde{x}}_{0}} \mathcal{L}\left(\boldsymbol{y}, \left({\mathcal{D}}\left(\boldsymbol{\tilde{x}}_{0}\right)\right)\right)$\\

        $\mathcal{L}^{total}_{ \boldsymbol{\tilde{x}}_{0}} = \mathcal{L}(\boldsymbol{y}, {\mathcal{D}}\left(\boldsymbol{\tilde{x}}_{0}\right)) + \mathcal{Q}\left(\boldsymbol{\tilde{x}}_{0}\right)$ 
  
        $\boldsymbol{\tilde{x}}_{0} \leftarrow \boldsymbol{\tilde{x}}_{0} -s\nabla_{\boldsymbol{\tilde{x}}_{0}} \mathcal{L}^{total}_{ \boldsymbol{\tilde{x}}_{0}}$
    
        Sample $\boldsymbol{x}_{t-1}$ by $q\left(\boldsymbol{x}_{t-1} \mid \boldsymbol{x}_{t}, \boldsymbol{\tilde{x}}_{0}\right) =\mathcal{N}\left(\boldsymbol{x}_{t-1}; \tilde{\boldsymbol{\mu}}_t\left(\boldsymbol{x}_{t}, \boldsymbol{\tilde{x}}_{0}\right), \tilde{\beta}_t \mathbf{I}\right)$, 
        
        where $\quad \tilde{\boldsymbol{\mu}}_t\left(\boldsymbol{x}_{t}, \boldsymbol{\tilde{x}}_{0}\right)=\frac{\sqrt{\bar{\alpha}_{t-1}} \beta_t}{1-\bar{\alpha}_t} \boldsymbol{\tilde{x}}_{0}+\frac{\sqrt{\alpha_t}\left(1-\bar{\alpha}_{t-1}\right)}{1-\bar{\alpha}_t} \boldsymbol{x}_{t} \quad$ and $\quad \tilde{\beta}_t=\frac{1-\bar{\alpha}_{t-1}}{1-\bar{\alpha}_t} \beta_t$
	}
	\Return $\boldsymbol{x}_{0}$
\label{algo6_v1}
\end{algorithm}

\section{The ELBO objective of GDP}
GDP is a Markov chain conditioned on $\boldsymbol{y}$, resulting in the following ELBO objective~\cite{song2020denoising}:
\begin{equation}\small
    \begin{aligned}
    & \mathbb{E}_{\boldsymbol{x}_0 \sim q\left(\boldsymbol{x}_0\right), \boldsymbol{y} \sim q\left(\boldsymbol{y} \mid \boldsymbol{x}_0\right)}\left[\log p_\theta\left(\boldsymbol{x}_0 \mid \boldsymbol{y}\right)\right]
    \geq \\ &-\mathbb{E}\left[\sum_{t=1}^{T-1} {\mathrm{KL}}\left(q^{t}\left(\boldsymbol{x}_t \mid \boldsymbol{x}_{t+1}, \boldsymbol{x}_0, \boldsymbol{y}\right) \| p_\theta^{t}\left(\boldsymbol{x}_t \mid \boldsymbol{x}_{t+1}, \boldsymbol{y}\right)\right)\right]
    \\ &+\mathbb{E}\left[\log p_\theta^{0}\left(\boldsymbol{x}_0 \mid \boldsymbol{x}_1, \boldsymbol{y}\right)\right] \\
    &-\mathbb{E}\left[{\mathrm{KL}}\left(q^{T}\left(\boldsymbol{x}_T \mid \boldsymbol{x}_0, \boldsymbol{y}\right) \| p_\theta^{T}\left(\boldsymbol{x}_T \mid \boldsymbol{y}\right)\right)\right]
    \end{aligned}
\end{equation}
where $q\left(\boldsymbol{x}_0\right)$ denotes the data distribution, $q\left(\boldsymbol{y} \mid \boldsymbol{x}_0\right)$ in the main paper, the expectation on the right-hand side is given by sampling $\boldsymbol{x}_0 \sim q\left(\boldsymbol{x}_0\right), \boldsymbol{y} \sim q\left(\boldsymbol{y} \mid \boldsymbol{x}_0\right), \boldsymbol{x}_T \sim q^{T}\left(\boldsymbol{x}_T \mid \boldsymbol{x}_0, \boldsymbol{y}\right)$, and $\boldsymbol{x}_t \sim q^{t}\left(\boldsymbol{x}_t \mid \boldsymbol{x}_{t+1}, \boldsymbol{x}_0, \boldsymbol{y}\right)$ for $t \in[1, T-1]$.

\section{Sampling with DDIM}
\begin{figure*}[htbp]
  \centering
   \includegraphics[width=23pc]{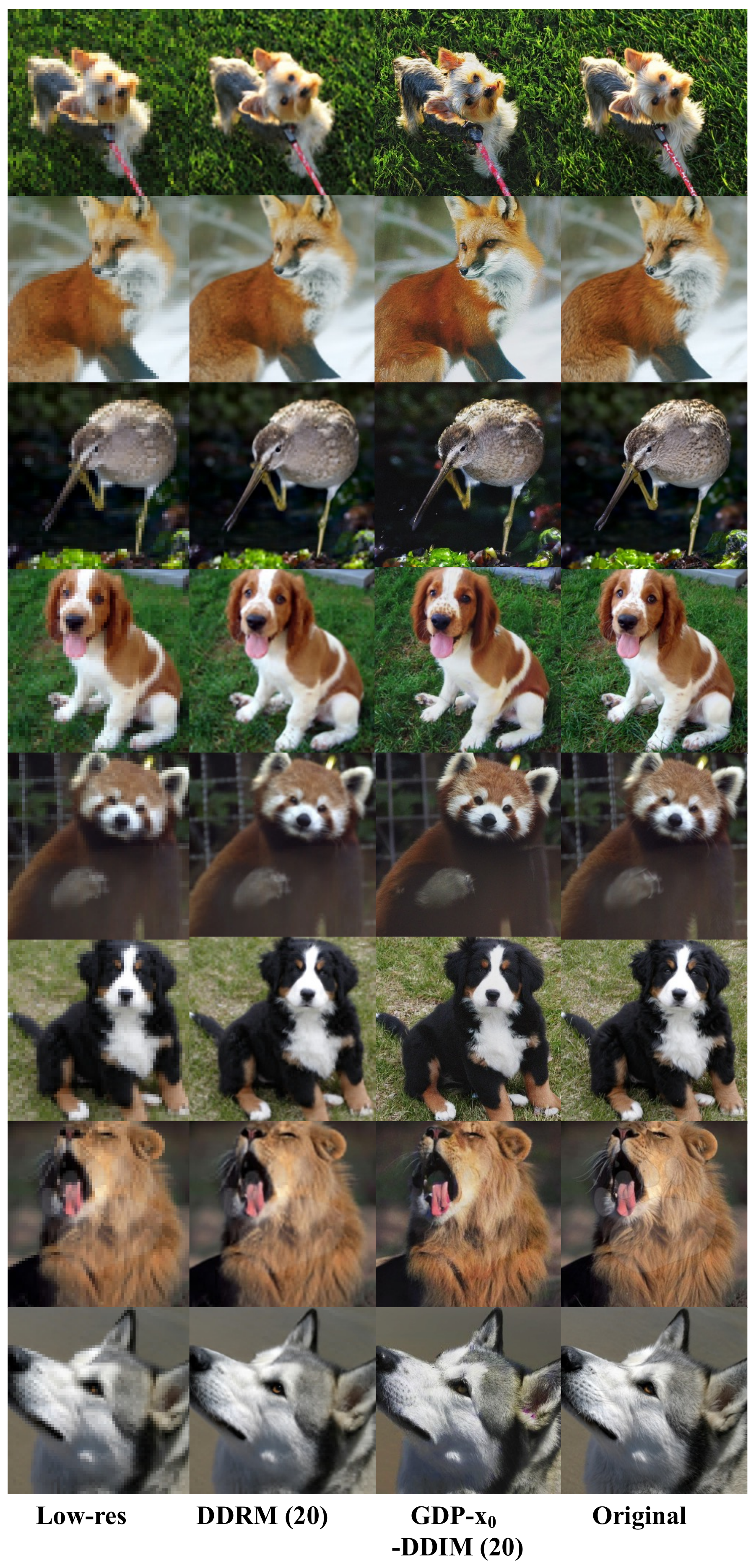}
   \caption{\textbf{More samples from the \underline{4 $\times$ super-resolution} task of GDP-$\boldsymbol{x}_0$-DDIM (20) compare with DDRM (20) on 256 $\times$ 256 \underline{ImageNet 1K.}} The generated images by the DDRM (20) are still blurred, while our proposed GDP-$x_0$ with 20 steps of DDIM sampling can restore more details.}
\label{DDIM-sr}
% \vspace{-0.5cm}
\end{figure*}

\begin{figure*}[htbp]
  \centering
   \includegraphics[width=23pc]{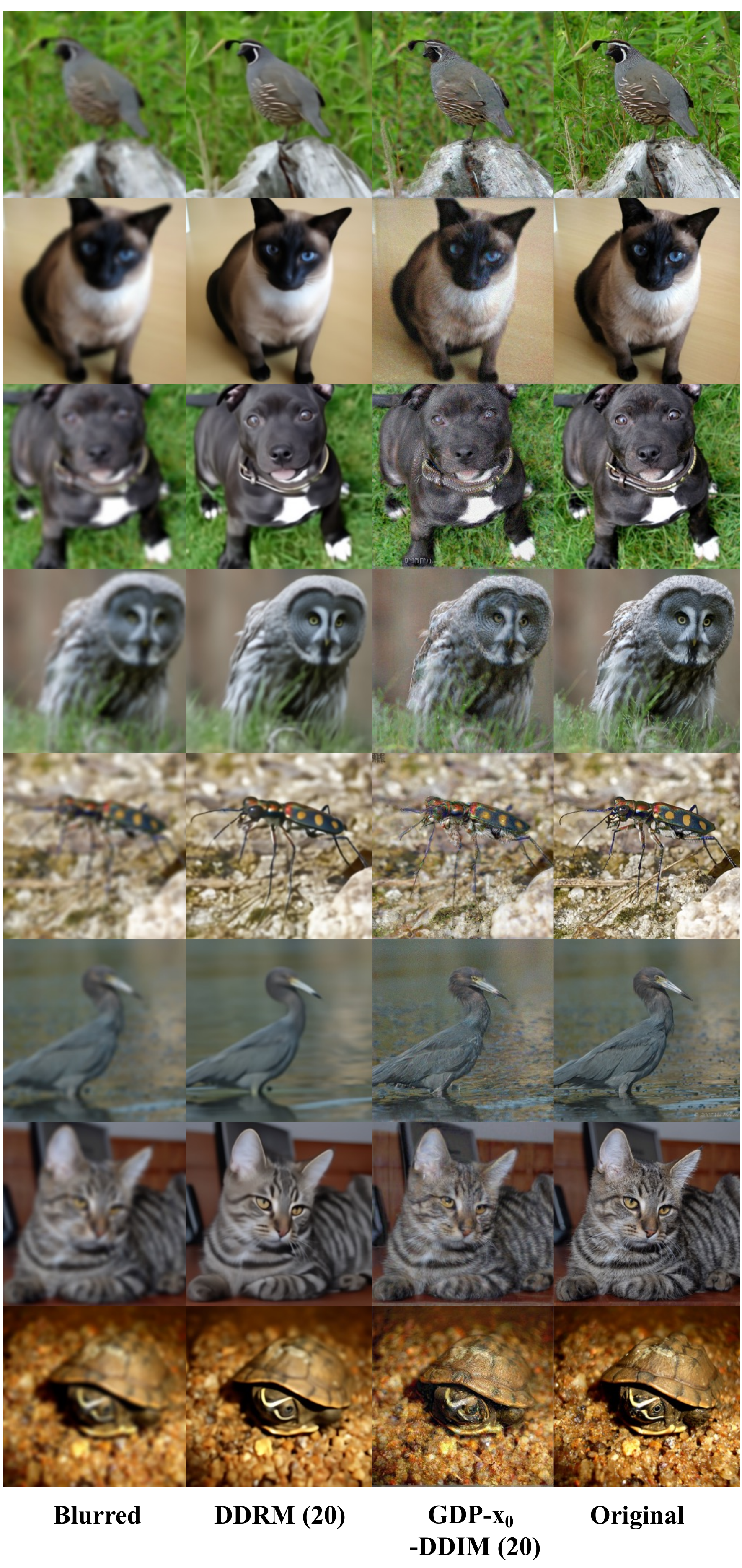}
   \caption{\textbf{More samples from the \underline{deblurring} task of GDP-$\boldsymbol{x}_0$-DDIM (20) compare with DDRM (20) on 256 $\times$ 256 \underline{ImageNet 1K.}} Our GDP-$x_0$-DDIM (20) can recover more details than DDRM (20) under the same DDIM steps. }
\label{DDIM-blur}
% \vspace{-0.5cm}
\end{figure*}

To accelerate the sampling strategy, GDP follows~\cite{nichol2021improved} to use DDIM, which skipping steps in the reverse process to speed up the DDPM generating process. We apply this method to the ImageNet dataset on the four tasks. We set the $T$=20 in the sampling process, while DDRM also utilizes the same time steps for a fair comparison. As shown in Table~\ref{DDIM}, our GDP-$x_0$-DDIM(20) outperforms DDRM(20) on consistency and FID across four tasks. Although DDRM(20) obtains better PSNR and SSIM than our GDP-$x_0$-DDIM(20), the qualitative results of DDRM(20) are still worse than our GDP-$x_0$-DDIM(20), which can be seen from Figs.~\ref{DDIM-sr} and~\ref{DDIM-blur}. 
Previous work~\cite{saharia2022image, chen2018fsrnet, dahl2017pixel, dosovitskiy2016generating} demonstrated that these conventional automated evaluation measures (PSNR and SSIM) do not correlate well with human perception when the input resolution is low, and the magnification is large. This is not surprising since these metrics tend to penalize any synthesized high-frequency detail that is not perfectly aligned with the target image.

% \usepackage{multirow}
% \usepackage{graphicx}
\begin{table*}[htbp]
\centering
\caption{\textbf{The performances of DDRM (20) and GDM-$\boldsymbol{x}_0$-DDIM(20) towards the four tasks on \underline{ImageNet 1k}.} The DDIM sample steps are all set to 20 to make a fair comparison.}
\resizebox{\textwidth}{!}{%
\begin{tabular}{l|cccc|cccc|cccc|cccc}
\toprule[1.5pt]
\multirow{2}{*}{Task} & \multicolumn{4}{c|}{4$\times$ super resolution}                     & \multicolumn{4}{c|}{Deblur}                    & \multicolumn{4}{c|}{25\% Impainting}         & \multicolumn{4}{c}{Colorization}       \\ \cline{2-17} 
                      & PSNR  & SSIM  & Consistency   & FID            & PSNR  & SSIM  & Consistency    & FID           & PSNR  & SSIM  & Consistency & FID            & PSNR  & SSIM  & Consistency    & FID   \\ \midrule[1pt]
DDRM(20)~\cite{kawar2022denoising}              & \textbf{26.53} & \textbf{0.784} & 19.39         & 40.75          & \textbf{35.64} & \textbf{0.978} & 50.24          & 4.78          & \textbf{34.28} & 
\textbf{0.958} & \textbf{4.08}        & 24.09          & \textbf{22.12} & \textbf{0.924} & 38.66 & 47.05 \\
GDP-$x_0$-DDIM(20)          & 23.77 & 0.623 & \textbf{9.24} & \textbf{39.46} & 24.87 & 0.683 & \textbf{44.39} & \textbf{3.66} & 30.82 & 0.892 & 7.10        & \textbf{19.70} & 21.13 & 0.840 &    \textbf{37.33}       & \textbf{41.38} \\ \bottomrule[1.5pt]
\end{tabular}%
}
\label{DDIM}
\end{table*}

% \usepackage{graphicx}
\begin{table*}[htbp]
\centering
\caption{The time comparison of GDP-$x_0$-DDIM(20) and GDP-$x_0$ on 4x super-resolution. These experiments are compared on Tesla A100.}
{%
\begin{tabular}{l|ccccc}
\toprule[1.5pt]
         & & Guidance scale & Total steps & Guidance times per steps & Generation time per image \\ \midrule[1pt]
GDP-$x_0$ & w.o. DDIM & 2e3            & 1000        & 6                        & 69.55                     \\
GDP-$x_0$-DDIM(20) & w. DDIM   & 22e5           & 20          & 20                       & 1.74                      \\ \bottomrule[1.5pt]
\end{tabular}%
}
\label{ddim-time}
\end{table*}

\begin{table*}[t]
\renewcommand\arraystretch{1}
\centering
% \vspace{-0.4cm}
\tabcolsep=0.05cm
\caption{The quantitative comparison of performance on CelebA.}
% \vspace{-0.3cm}
{%
\begin{tabular}{c|cccc|cccc|cccc}
\toprule[1.5pt]
\multirow{2}{*}{CelebA} & \multicolumn{4}{c|}{4x SR}      & \multicolumn{4}{c|}{Deblur}     & \multicolumn{4}{c}{25\% Inpainting}  \\ \cline{2-13} 
                      & PSNR & SSIM & Consistency & FID & PSNR & SSIM & Consistency & FID & PSNR   & SSIM  & Consistency  & FID  \\ \midrule[1pt]
DDRM                  & 29.50     &   0.863   &  6.82           &  87.71   & \textbf{36.51}    &  \textbf{0.98}    &   35.91          &  14.30   &  31.99      &  0.918     &      \textbf{0.47 }       & 69.46      \\
GDP-$x_t$                &  29.19    &  0.847    &     14.11        & 94.98    & 27.35     & 0.81     &    34.87         &   9.97  &  36.19      &  0.963     &  1.94           & 22.53     \\
GDP-$x_0$                &  \textbf{30.26}    &   \textbf{0.868}   &      \textbf{5.33}       & \textbf{46.64}    &   28.66   & 0.83     &    \textbf{32.66}         &  \textbf{4.50}   &  \textbf{37.70}      &   \textbf{0.972}    &    0.51         & \textbf{11.62}      \\ \bottomrule[1.5pt]
\end{tabular}%
}
% \vspace{-0.35cm}
\end{table*}

% \usepackage{multirow}
% \usepackage{graphicx}

\begin{table*}[t]
% \renewcommand\arraystretch{0.7}
% % \vspace{-0.9cm}
\centering
\caption{\footnotesize{The quantitative comparison of results on LSUN bedroom.}}
% \vspace{-0.4cm}
{%
\begin{tabular}{c|cc|cc|cc|cc}
\toprule[1.5pt]
\multirow{2}{*}{LSUN Bedroom} & \multicolumn{2}{c|}{4x SR} & \multicolumn{2}{c|}{Deblur} & \multicolumn{2}{c|}{25\% Inpainting} & \multicolumn{2}{c}{Colorization} \\ \cline{2-9} 
                      & Consistency      & FID     & Consistency      & FID      & Consistency           & FID          & Consistency         & FID        \\ \midrule[1pt]
DDRM                  &    20.33              &  40.12        &      43.78            &  10.16        &       \textbf{5.33}                &    22.49          &      35.16               &   45.22         \\
GDP-$x_t$                &       70.46             &   58.62      &      46.90            &  12.50       &  9.33                      &         20.63     &          66.88           &   57.13         \\
GDP-$x_0$                &     \textbf{7.66}             &  \textbf{36.94}       &     \textbf{42.28}             &    \textbf{9.51}      &        6.77               &     \textbf{18.34}         &  \textbf{33.51}                    &  \textbf{34.59}           \\ \bottomrule[1.5pt]
\end{tabular}%
}
% \vspace{-0.5cm}
\end{table*}

% \usepackage{graphicx}
\begin{table*}[t]
\centering
\caption{The weight of reconstruction loss and quality enhancement loss.}
{%
\begin{tabular}{lcccc}
\toprule[1.5pt]
                      & MSE loss & Exposure Control Loss & Color Constancy Loss & Illumination Smoothness Loss \\ \midrule[1pt]
Colorization          & 1        & 0                     & 500                  & 0                            \\
Low-light Enhancement & 1        & 1/100                 & 1/200                & 1                            \\
HDR recovery          & 1        & 1/100                 & 1/200                & 1 \\                         \bottomrule[1.5pt]  
\end{tabular}%
}
\end{table*}

\section{Image Guidance}

A conditioner $p(\boldsymbol{y} \mid {\boldsymbol{x}})$ is exploited to improve a diffusion generator. Specifically, we can utilize a conditioner $p_{\phi}\left(\boldsymbol{y} \mid \boldsymbol{x}_{t}, t\right)$ on input images, and then use gradients $\nabla_{\boldsymbol{x}_{t}} \log p_{\phi}\left(\boldsymbol{y} \mid \boldsymbol{x}_{t}, t\right)$ to guide the diffusion sampling process towards a given the degraded images $\boldsymbol{y}$.

In this section, we will describe how to use such conditioners to improve the quality of sampled images. The notation is chosen as $p_{\phi}\left(\boldsymbol{y} \mid \boldsymbol{x}_{t}, t\right)=p_{\phi}\left(\boldsymbol{y} \mid \boldsymbol{x}_{t}\right)$ and $\epsilon_{\theta}\left(\boldsymbol{x}_{t}, t\right)=\epsilon_{\theta}\left(\boldsymbol{x}_{t}\right)$ for brevity. Note that they refer to separate functions for each time step $t$.

\subsection{Conditional Reverse Process}

Assume a diffusion model with an unconditional reverse noising process $p_{\theta}\left(\boldsymbol{x}_{t} \mid \boldsymbol{x}_{t+1}\right)$. In image restoration and enhancement, the corrupted inputs can be regarded as conditions. Therefore, we regard $\boldsymbol{y}$ as the input images, and $\boldsymbol{x}_{t}$ as the generated images in time step $t$. 
% The $\boldsymbol{y}$ can be obtained by Equation 2 by the given $t$. 
Then, the conditioner is formulated as follows:
\begin{equation}
p_{\phi}\left(\boldsymbol{y} \mid \boldsymbol{x}_{t}\right) = \frac{1}{K}exp\left(-\mathcal{L}\left(\boldsymbol{y}, \mathcal{D}\left(\boldsymbol{x}_{t}\right)\right)\right),
\label{eq-g}
\end{equation}
where $\mathcal{D}$ represents the degradation function, $\mathcal{L}$ stands for Mean Square Error together with optional Quality Enhancement Loss, and $K$ is an arbitrary constant. 
In order to condition this on the input corrupted image $\boldsymbol{y}$, it is sufficient to sample each transition based on the following:
\begin{equation}
p_{\theta, \phi}\left(\boldsymbol{x}_{t} \mid \boldsymbol{x}_{t+1}, \boldsymbol{y}\right)=C p_{\theta}\left(\boldsymbol{x}_{t} \mid \boldsymbol{x}_{t+1}\right) p_{\phi}\left(\boldsymbol{y} \mid \boldsymbol{x}_{t}\right)
\label{eq-z}
\end{equation}
where $C$ denotes a normalizing constant. 
It is typically intractable to sample from this distribution exactly, but Sohl-Dickstein~\etal~\cite{sohl2015deep} show that it can be approximated as a perturbed Gaussian distribution.
Sampling accurately from this distribution is often tricky, but Sohl-Dickstein~\etal~\cite{sohl2015deep} prove that it could be approximated as a perturbed Gaussian distribution.
% Here, we review this derivation.
It is formulated that the diffusion model samples the previous time step $\boldsymbol{x}_{t}$ from time step $\boldsymbol{x}_{t+1}$ via a Gaussian distribution:

\begin{equation}
p_{\theta}\left(\boldsymbol{x}_{t} \mid \boldsymbol{x}_{t+1}\right)=\mathcal{N}(\mu, \Sigma)
\end{equation}
\begin{equation}
\log p_{\theta}\left(\boldsymbol{x}_{t} \mid \boldsymbol{x}_{t+1}\right)=-\frac{1}{2}\left(\boldsymbol{x}_{t}-\mu\right)_{T} \Sigma^{-1}\left(\boldsymbol{x}_{t}-\mu\right)+Z
\end{equation}
We can assume that $\log _{\phi} p\left(\boldsymbol{y} \mid \boldsymbol{x}_{t}\right)$ owns low curvature when compared with $\Sigma^{-1}$. 
This assumption is reasonable under the constraint that the infinite diffusion step, where $\|\Sigma\| \rightarrow 0$. 
Under the circumstances, $\log p_{\phi}\left(\boldsymbol{y} \mid \boldsymbol{x}_{t}\right)$ can be approximated via a Taylor expansion around $\boldsymbol{x}_{t}=\mu$ as:
\begin{equation}
\begin{aligned}
    \log p_{\phi}\left(\boldsymbol{y} \mid \boldsymbol{x}_{t}\right) &\left.\approx \log p_{\phi}\left(\boldsymbol{y} \mid \boldsymbol{x}_{t}\right)\right|_{\boldsymbol{x}_{t}=\mu} \\ &  + \left.\left(\boldsymbol{x}_{t}-\mu\right) \nabla_{\boldsymbol{x}_{t}} \log p_{\phi}\left(\boldsymbol{y} \mid \boldsymbol{x}_{t}\right)\right|_{\boldsymbol{x}_{t}=\mu} \\
    &=\left(\boldsymbol{x}_{t}-\mu\right) g+Z_{1}
\end{aligned}
\end{equation}
Here, $g=\nabla_{\boldsymbol{x}_{t}} \log p_{\phi}\left(\boldsymbol{y} \mid \boldsymbol{x}_{t}\right) \|_{\boldsymbol{x}_{t}=\mu}$, and $Z_{1}$ is a constant. We can replace the $g$ with Eq.~\ref{eq-g} as follows:
\begin{align}
\log p\left(\boldsymbol{y} \mid \boldsymbol{x}_{t}\right)=-\mathcal{L}\left(\boldsymbol{y}, \mathcal{D}\left(\boldsymbol{x}_{t}\right)\right)-\log K \\
g = \nabla_{\boldsymbol{x}_{t}} \log p\left(\boldsymbol{y} \mid \boldsymbol{x}_{t}\right)=-\nabla_{\boldsymbol{x}_{t}} \mathcal{L}\left(\boldsymbol{y}, \mathcal{D}\left(\boldsymbol{x}_{t}\right)\right)
\end{align}
This gives:
\begin{equation}\small
\begin{aligned}
    & \log \left(p_{\theta}\left(\boldsymbol{x}_{t} \mid \boldsymbol{x}_{t+1}\right) p_{\phi}\left(\boldsymbol{y}  \mid \boldsymbol{x}_{t}\right)\right) \\ & \approx-\frac{1}{2}\left(\boldsymbol{x}_{t}-\mu\right)^{T} \Sigma^{-1}\left(\boldsymbol{x}_{t}-\mu\right)+\left(\boldsymbol{x}_{t}-\mu\right) g+Z_{2} \\
    &=-\frac{1}{2}\left(\boldsymbol{x}_{t}-\mu-\Sigma g\right)^{T} \Sigma^{-1}\left(\boldsymbol{x}_{t}-\mu-\Sigma g\right)+\frac{1}{2} g^{T} \Sigma g+Z_{2} \\
    &=-\frac{1}{2}\left(\boldsymbol{x}_{t}-\mu-\Sigma g\right)^{T} \Sigma^{-1}\left(\boldsymbol{x}_{t}-\mu-\Sigma g\right)+Z_{3} \\
    &=\log p(z)+Z_{4}, z \sim \mathcal{N}(\mu+\Sigma g, \Sigma)
\label{SM:eq2}
\end{aligned}
\end{equation}
% We can safely ignore the constant term $C_{4}$ since it corresponds to the normalizing coefficient $Z$ in Eq.~\ref{eq-z}. 
where the constant term $C_{4}$ could be safely ignored because it is equivalent to the normalizing coefficient $Z$ in Eq.~\ref{eq-z}.
Thus, we find that the conditional transition operator can be approximated by a Gaussian similar to the unconditional transition operator, but with a mean shifted by $\Sigma g$.
% Algorithm 1 summarizes the corresponding sampling algorithm. 
Moreover, an optional scaling factor $s$ is included for gradients, which will be described in more detail in Sec.~\ref{SM:Scaling Conditioner Gradients}. However, it is experimentally found that this guidance way might not be effective enough, where our GDP-$x_0$ is systematically studied. 

\subsection{Conditional Diffusion Process}

Here, we figure out that conditional sampling can be fulfilled with a transition operator proportional to $p_{\theta}\left(\boldsymbol{x}_{t} \mid \boldsymbol{x}_{t+1}\right) p_{\phi}\left(\boldsymbol{y} \mid \boldsymbol{x}_{t}\right)$, where $p_{\theta}\left(\boldsymbol{x}_{t} \mid \boldsymbol{x}_{t+1}\right)$ approximates $q\left(\boldsymbol{x}_{t} \mid \boldsymbol{x}_{t+1}\right)$ and $p_{\phi}\left(\boldsymbol{y} \mid \boldsymbol{x}_{t}\right)$ approximates the distribution of the input for a noised sample $\boldsymbol{x}_{t}$.

A conditional Markovian noising process $\hat{q}$ is similar to $q$. 
And $\hat{q}\left(\boldsymbol{y} \mid \boldsymbol{x}_{0}\right)$ is assumed as a known and readily available degraded images distribution for each sample.
\begin{align}
\hat{q}\left(\boldsymbol{x}_{0}\right) &:=q\left(\boldsymbol{x}_{0}\right) \\
\hat{q}\left(\boldsymbol{y} \mid \boldsymbol{x}_{0}\right) &:=\text {Corrupted input image per sample} \\
\hat{q}\left(\boldsymbol{x}_{t+1} \mid \boldsymbol{x}_{t}, \boldsymbol{y}\right) &:=q\left(\boldsymbol{x}_{t+1} \mid \boldsymbol{x}_{t}\right) \\
\hat{q}\left(\boldsymbol{x}_{1: T} \mid \boldsymbol{x}_{0}, \boldsymbol{y}\right) &:=\prod_{t=1}^{T} \hat{q}\left(\boldsymbol{x}_{t} \mid \boldsymbol{x}_{t-1}, \boldsymbol{y}\right)
\end{align}

Assuming that the noising process $\hat{q}$ is conditioned on $\boldsymbol{y}$, we can reveal that $\hat{q}$ behaves exactly like $q$ when not conditioned on $\boldsymbol{y}$. 
According to this idea, we first derive the unconditional noising operator $\hat{q}\left(\boldsymbol{x}_{t+1} \mid \boldsymbol{x}_{t}\right)$ :
\begin{align}
\hat{q}\left(\boldsymbol{x}_{t+1} \mid \boldsymbol{x}_{t}\right) &=\int_{\boldsymbol{y}} \hat{q}\left(\boldsymbol{x}_{t+1}, \boldsymbol{y} \mid \boldsymbol{x}_{t}\right) d \boldsymbol{y} \\
&=\int_{\boldsymbol{y}} \hat{q}\left(\boldsymbol{x}_{t+1} \mid \boldsymbol{x}_{t}, \boldsymbol{y}\right) \hat{q}\left(\boldsymbol{y} \mid \boldsymbol{x}_{t}\right) d \boldsymbol{y} \\
&=\int_{\boldsymbol{y}} q\left(\boldsymbol{x}_{t+1} \mid \boldsymbol{x}_{t}\right) \hat{q}\left(\boldsymbol{y} \mid \boldsymbol{x}_{t}\right) d \boldsymbol{y} \\
&=q\left(\boldsymbol{x}_{t+1} \mid \boldsymbol{x}_{t}\right) \int_{\boldsymbol{y}} \hat{q}\left(\boldsymbol{y} \mid \boldsymbol{x}_{t}\right) d \boldsymbol{y} \\
&=q\left(\boldsymbol{x}_{t+1} \mid \boldsymbol{x}_{t}\right) \\
&=\hat{q}\left(\boldsymbol{x}_{t+1} \mid \boldsymbol{x}_{t}, \boldsymbol{y}\right)
\end{align}
Similarly, the joint distribution $\hat{q}\left(\boldsymbol{x}_{1: T} \mid \boldsymbol{x}_{0}\right)$ can be written as:
\begin{align}
\hat{q}\left(\boldsymbol{x}_{1: T} \mid \boldsymbol{x}_{0}\right) &=\int_{\boldsymbol{y}} \hat{q}\left(\boldsymbol{x}_{1: T}, \boldsymbol{y} \mid \boldsymbol{x}_{0}\right) d \boldsymbol{y} \\
&=\int_{\boldsymbol{y}} \hat{q}\left(\boldsymbol{y} \mid \boldsymbol{x}_{0}\right) \hat{q}\left(\boldsymbol{x}_{1: T} \mid \boldsymbol{x}_{0}, \boldsymbol{y}\right) d \boldsymbol{y} \\
&=\int_{\boldsymbol{y}} \hat{q}\left(\boldsymbol{y} \mid \boldsymbol{x}_{0}\right) \prod_{t=1}^{T} \hat{q}\left(\boldsymbol{x}_{t} \mid \boldsymbol{x}_{t-1}, \boldsymbol{y}\right) d \boldsymbol{y} \\
&=\int_{\boldsymbol{y}} \hat{q}\left(\boldsymbol{y} \mid \boldsymbol{x}_{0}\right) \prod_{t=1}^{T} q\left(\boldsymbol{x}_{t} \mid \boldsymbol{x}_{t-1}\right) d \boldsymbol{y} \\
&=\prod_{t=1}^{T} q\left(\boldsymbol{x}_{t} \mid \boldsymbol{x}_{t-1}\right) \int_{\boldsymbol{y}} \hat{q}\left(\boldsymbol{y} \mid \boldsymbol{x}_{0}\right) d \boldsymbol{y} \\
&=\prod_{t=1}^{T} q\left(\boldsymbol{x}_{t} \mid \boldsymbol{x}_{t-1}\right) \\
&=q\left(\boldsymbol{x}_{1: T} \mid \boldsymbol{x}_{0}\right)
\label{SM:eq1}
\end{align}
$\hat{q}\left(\boldsymbol{x}_{t}\right)$ can be derived by using Eq.~\ref{SM:eq1} as follows:
\begin{align}
\hat{q}\left(\boldsymbol{x}_{t}\right) &=\int_{\boldsymbol{x}_{0: t-1}} \hat{q}\left(\boldsymbol{x}_{0}, \ldots, \boldsymbol{x}_{t}\right) d \boldsymbol{x}_{0: t-1} \\
&=\int_{\boldsymbol{x}_{0: t-1}} \hat{q}\left(\boldsymbol{x}_{0}\right) \hat{q}\left(\boldsymbol{x}_{1}, \ldots, \boldsymbol{x}_{t} \mid \boldsymbol{x}_{0}\right) d \boldsymbol{x}_{0: t-1} \\
&=\int_{\boldsymbol{x}_{0: t-1}} q\left(\boldsymbol{x}_{0}\right) q\left(\boldsymbol{x}_{1}, \ldots, \boldsymbol{x}_{t} \mid \boldsymbol{x}_{0}\right) d \boldsymbol{x}_{0: t-1} \\
&=\int_{\boldsymbol{x}_{0: t-1}} q\left(\boldsymbol{x}_{0}, \ldots, \boldsymbol{x}_{t}\right) d \boldsymbol{x}_{0: t-1} \\
&=q\left(\boldsymbol{x}_{t}\right)
\end{align}

It is proved by Bayes rule that the unconditional reverse process $\hat{q}\left(\boldsymbol{x}_{t} \mid \boldsymbol{x}_{t+1}\right)=q\left(\boldsymbol{x}_{t} \mid \boldsymbol{x}_{t+1}\right)$ when using the identities $\hat{q}\left(\boldsymbol{x}_{t}\right)=q\left(\boldsymbol{x}_{t}\right)$ and $\hat{q}\left(\boldsymbol{x}_{t+1} \mid \boldsymbol{x}_{t}\right)=q\left(\boldsymbol{x}_{t+1} \mid \boldsymbol{x}_{t}\right)$.

Note that $\hat{q}$ is able to produce an input function $\hat{q}\left(\boldsymbol{y} \mid \boldsymbol{x}_{t}\right)$. It is shown that this distribution of the input does not depend on $\boldsymbol{x}_{t+1}$ (the noisy version of $\boldsymbol{x}_{t}$), we will discuss this fact later by exploiting:
\begin{align}
\hat{q}\left(\boldsymbol{y} \mid \boldsymbol{x}_{t}, \boldsymbol{x}_{t+1}\right) &=\hat{q}\left(\boldsymbol{x}_{t+1} \mid \boldsymbol{x}_{t}, \boldsymbol{y}\right) \frac{\hat{q}\left(\boldsymbol{y} \mid \boldsymbol{x}_{t}\right)}{\hat{q}\left(\boldsymbol{x}_{t+1} \mid \boldsymbol{x}_{t}\right)} \\
&=\hat{q}\left(\boldsymbol{x}_{t+1} \mid \boldsymbol{x}_{t}\right) \frac{\hat{q}\left(\boldsymbol{y} \mid \boldsymbol{x}_{t}\right)}{\hat{q}\left(\boldsymbol{x}_{t+1} \mid \boldsymbol{x}_{t}\right)} \\
&=\hat{q}\left(\boldsymbol{y} \mid \boldsymbol{x}_{t}\right)
\end{align}
In this way, the conditional reverse process can be derived as:
\begin{align}
\hat{q}\left(\boldsymbol{x}_{t} \mid \boldsymbol{x}_{t+1}, \boldsymbol{y}\right) &=\frac{\hat{q}\left(\boldsymbol{x}_{t}, \boldsymbol{x}_{t+1}, \boldsymbol{y}\right)}{\hat{q}\left(\boldsymbol{x}_{t+1}, \boldsymbol{y}\right)} \\
&=\frac{\hat{q}\left(\boldsymbol{x}_{t}, \boldsymbol{x}_{t+1}, \boldsymbol{y}\right)}{\hat{q}\left(\boldsymbol{y} \mid \boldsymbol{x}_{t+1}\right) \hat{q}\left(\boldsymbol{x}_{t+1}\right)} \\
&=\frac{\hat{q}\left(\boldsymbol{x}_{t} \mid \boldsymbol{x}_{t+1}\right) \hat{q}\left(\boldsymbol{y} \mid \boldsymbol{x}_{t}, \boldsymbol{x}_{t+1}\right) \hat{q}\left(\boldsymbol{x}_{t+1}\right)}{\hat{q}\left(\boldsymbol{y} \mid \boldsymbol{x}_{t+1}\right) \hat{q}\left(\boldsymbol{x}_{t+1}\right)} \\
&=\frac{\hat{q}\left(\boldsymbol{x}_{t} \mid \boldsymbol{x}_{t+1}\right) \hat{q}\left(\boldsymbol{y} \mid \boldsymbol{x}_{t}, \boldsymbol{x}_{t+1}\right)}{\hat{q}\left(\boldsymbol{y} \mid \boldsymbol{x}_{t+1}\right)} \\
&=\frac{\hat{q}\left(\boldsymbol{x}_{t} \mid \boldsymbol{x}_{t+1}\right) \hat{q}\left(\boldsymbol{y} \mid \boldsymbol{x}_{t}\right)}{\hat{q}\left(\boldsymbol{y} \mid \boldsymbol{x}_{t+1}\right)} \\
&=\frac{q\left(\boldsymbol{x}_{t} \mid \boldsymbol{x}_{t+1}\right) \hat{q}\left(\boldsymbol{y} \mid \boldsymbol{x}_{t}\right)}{\hat{q}\left(\boldsymbol{y} \mid \boldsymbol{x}_{t+1}\right)}
\end{align}
where the $\hat{q}\left(\boldsymbol{y} \mid \boldsymbol{x}_{t+1}\right)$ can be treated as a constant because it does not depend on $\boldsymbol{x}_{ t+1}$.
Therefore, we want to sample from the distribution $C q\left(\boldsymbol{x}_{t} \mid \boldsymbol{x}_{t+1}\right) \hat{q}\left(\boldsymbol{y} \mid \boldsymbol{x}_{t}\right)$ where $C$ denotes the normalization constant.
We already have a neural network approximation of $q\left(\boldsymbol{x}_{t} \mid \boldsymbol{x}_{t+1}\right)$ called $p_{\theta}\left (\boldsymbol{x}_{t} \mid \boldsymbol{x}_{t+1}\right)$, so the rest is $\hat{q}\left(\boldsymbol{y} \mid \boldsymbol{x}_{t}\right)$ that can be obtained by computing a conditioner $p_{\phi}\left(\boldsymbol{y} \mid \boldsymbol{x}_{t}\right)$ on noised images $\boldsymbol{x}_{t}$ derived by sampling from $q\left(\boldsymbol{x}_{t}\right)$.

\subsection{Scaling Conditioner Gradients}
\label{SM:Scaling Conditioner Gradients}
The conditioner is incorporated into the sampling process of the diffusion model using Eq.~\ref{SM:eq2}. 
To unveil the effect of scaling conditioner gradients, note that $s \cdot \nabla_{\boldsymbol{x}} \log p(\boldsymbol{y} \mid \boldsymbol{x})=\nabla_{\boldsymbol{x}} \log \frac{1}{K} p(\boldsymbol{y} \mid \boldsymbol{x})^{s}$, where $K$ is an arbitrary constant. 
Thus, the conditioning process is still theoretically based on the re-normalized distribution of the input proportional to $p(\boldsymbol{y} \mid \boldsymbol{x})^{s}$.
If $s>1$, this distribution becomes sharper than $p(\boldsymbol{y} \mid \boldsymbol{x})$ because larger values are exponentially magnified.
Therefore, using a larger gradient scale to focus more on the modes of the conditioner may be beneficial in producing higher fidelity (but less diverse) samples.
In this paper, due to the observation that $\Sigma$ might exert a negative influence on the quality of images. Therefore, with the absence of the $\Sigma$, the guidance scale can be a variable scale $\hat{s}$, where $s = \Sigma \hat{s}$.
Thanks to this variable scale $\hat{s}$, the quality of images can be promoted

\section{Additional Results on Linear inverse problems}

\begin{figure*}[htbp]
  \centering
   \includegraphics[width=\linewidth]{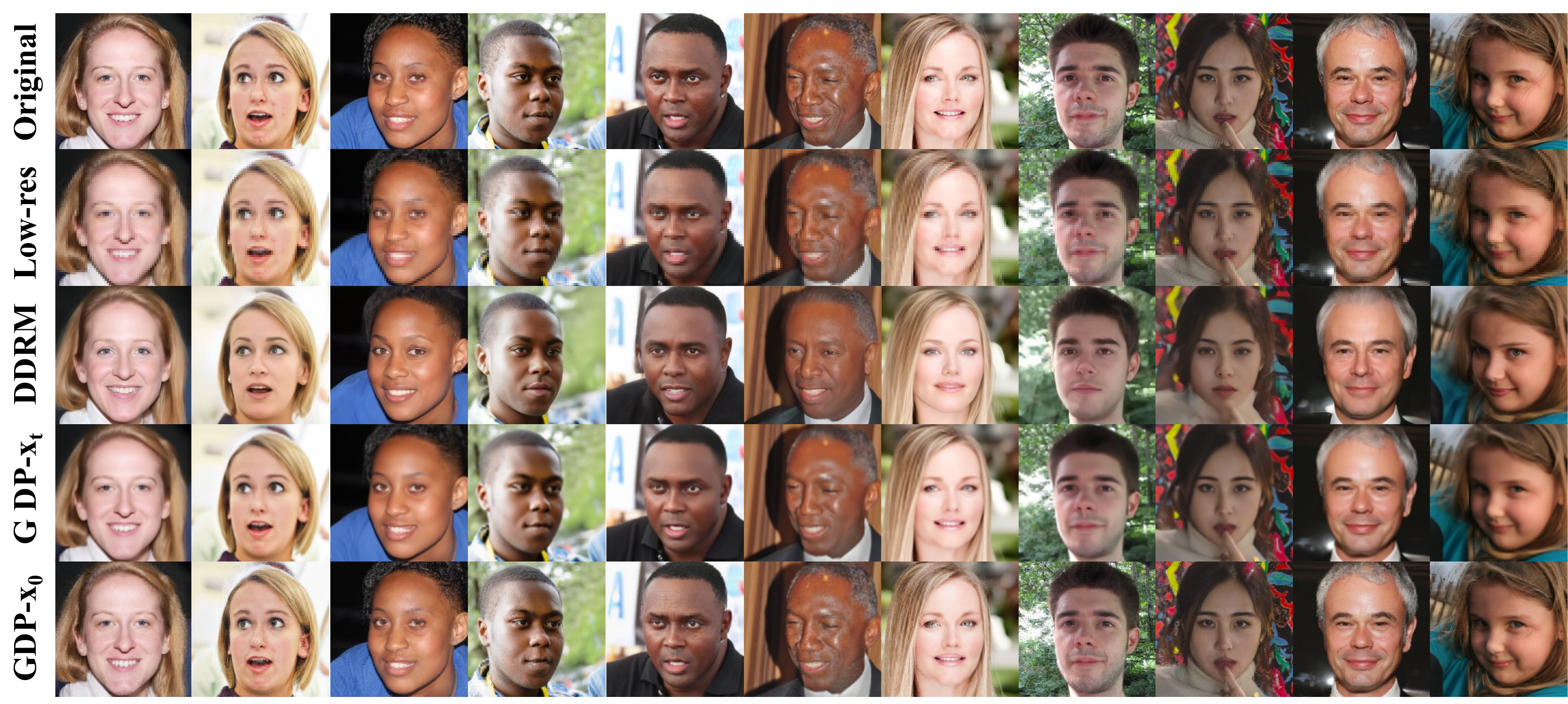}
   \caption{\textbf{\underline{4 $\times$ super-resolution} results of DDRM, GDP-$\boldsymbol{x}_t$, and GDP-$\boldsymbol{x}_0$ on \underline{CelebA} face images.} Compared with GDP-$x_t$ and DDRM, GDP-$x_0$ can restore more realistic faces, such as the wrinkles on the faces, systematically demonstrating the superiority of the guidance on $x_0$ protocol.}
\label{SM:figure-sr4-celeba}
% \vspace{-0.5cm}
\end{figure*}

\begin{figure*}[htbp]
  \centering
   \includegraphics[width=\linewidth]{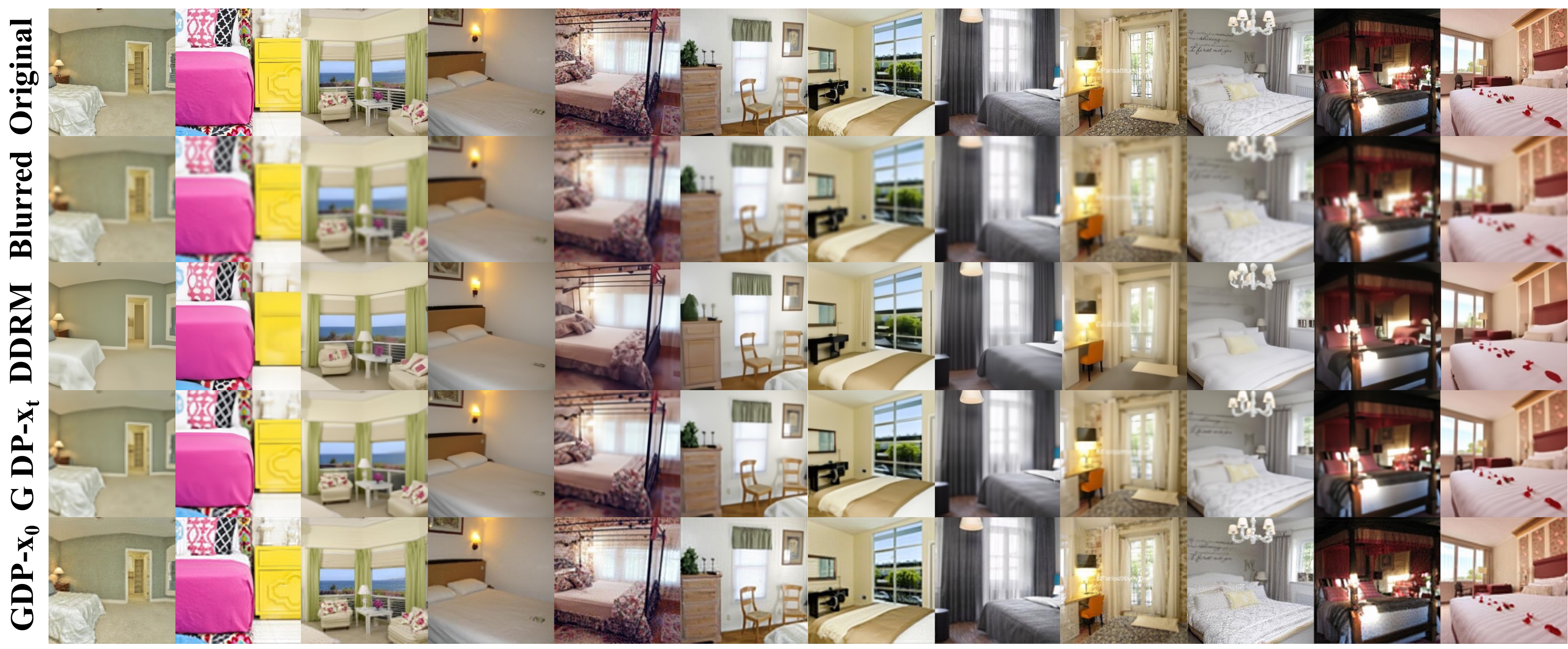}
   \caption{\textbf{\underline{Deblurring} results of DDRM, GDP-$\boldsymbol{x}_t$, and GDP-$\boldsymbol{x}_0$ on \underline{LSUN bedroom} images.}}
% \vspace{-0.5cm}
\label{SM:figure-blur-bedroom}
\end{figure*}

\begin{figure*}[htbp]
  \centering
   \includegraphics[width=\linewidth]{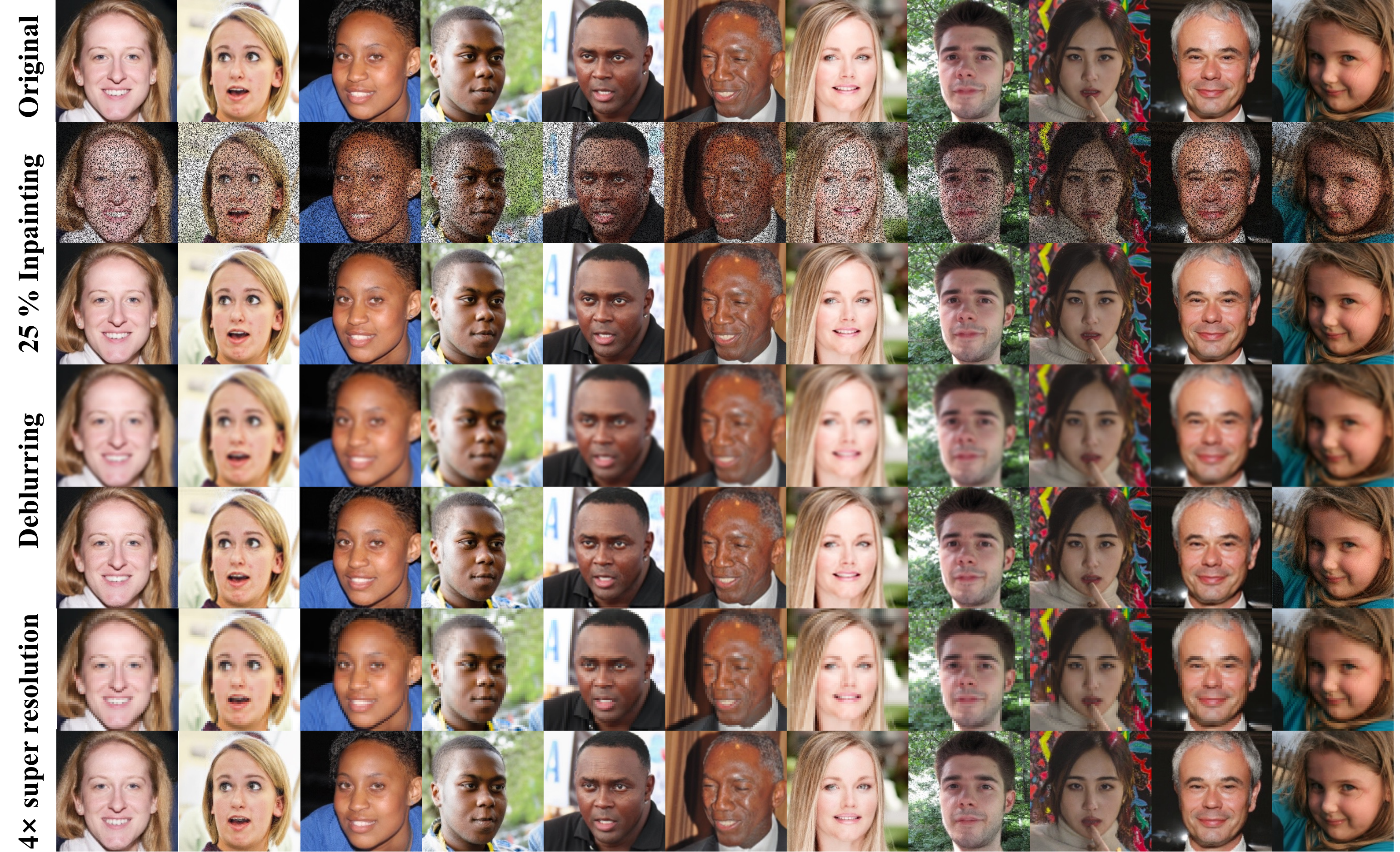}
   \caption{\textbf{Pairs of degraded and recovered 256 $\times$ 256 \underline{CelebA face} images with a GDP-$\boldsymbol{x}_0$.} Three tasks including $25 \%$ inpainting, deblurring and 4 $\times$ super-resolution are vividly depicted.}
% \vspace{-0.5cm}
\label{SM:figure-3task-face}
\end{figure*}

\begin{figure*}[htbp]
  \centering
   \includegraphics[width=\linewidth]{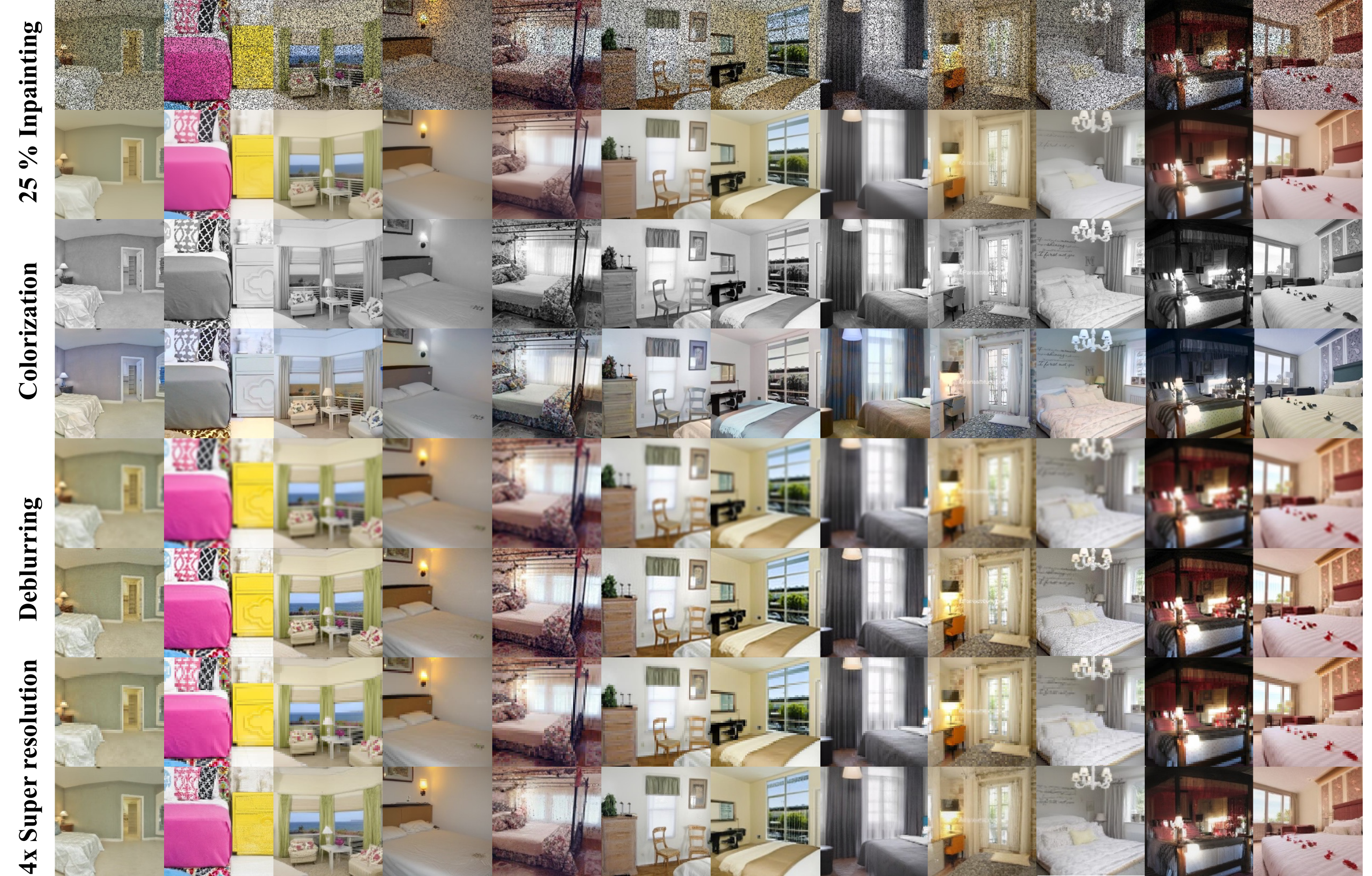}
   \caption{\textbf{Pairs of degraded and recovered 256 $\times$ 256 \underline{LSUN bedroom} images with a GDP-$\boldsymbol{x}_0$.} We show more samples under the $25 \%$ inpainting, colorization, deblurring and 4 $\times$ super-resolution.}
% \vspace{-0.5cm}
\label{SM:figure-4task-bedroom}
\end{figure*}

\begin{figure*}[htbp]
  \centering
   \includegraphics[width=35pc]{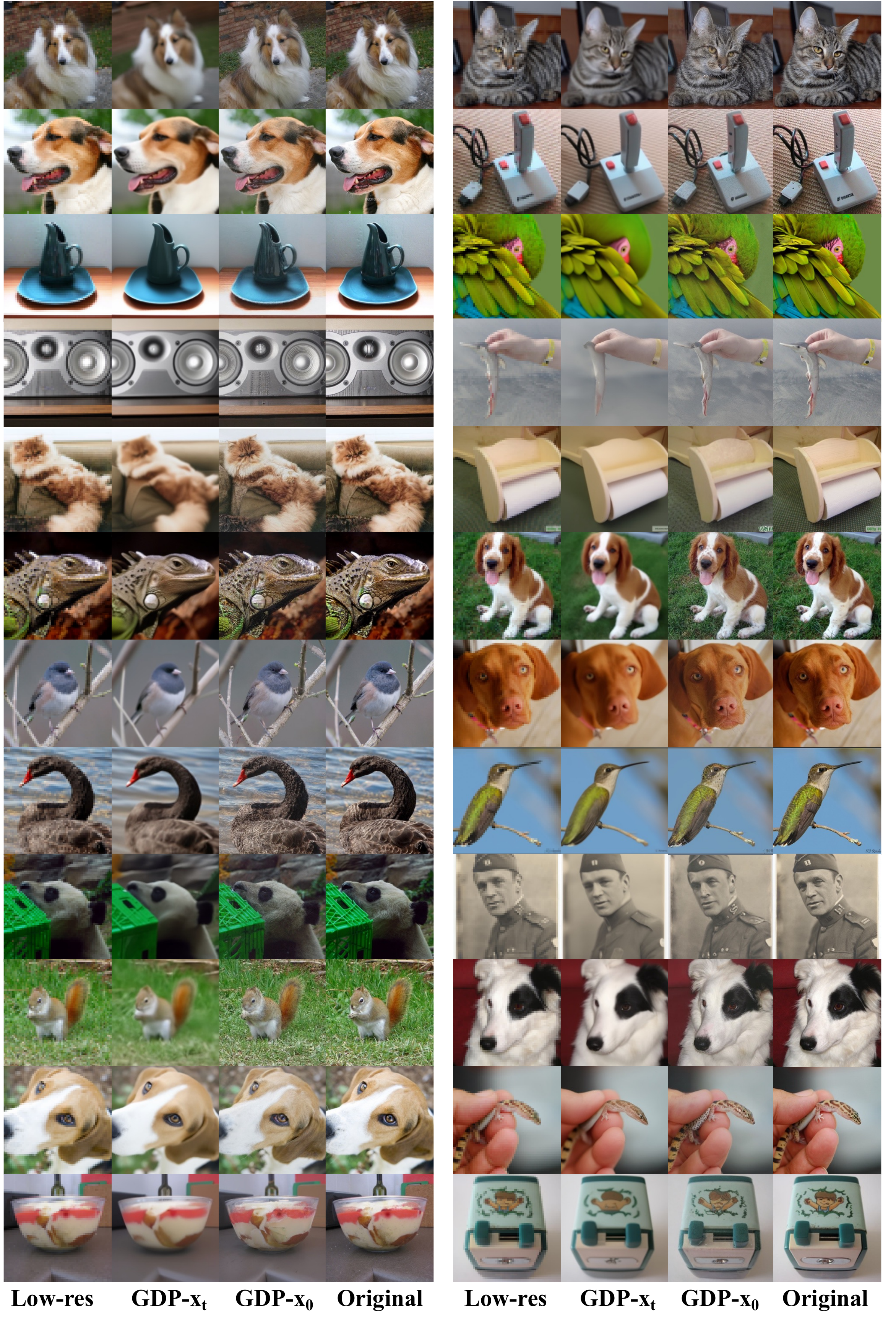}
   \caption{\textbf{Uncurated samples from the \underline{4 $\times$ super-resolution} task on 256 $\times$ 256 \underline{ImageNet 1K.}}}
% \vspace{-0.5cm}
\label{SM:figure-sr4-imagenet}
\end{figure*}

\begin{figure*}[htbp]
  \centering
   \includegraphics[width=24pc]{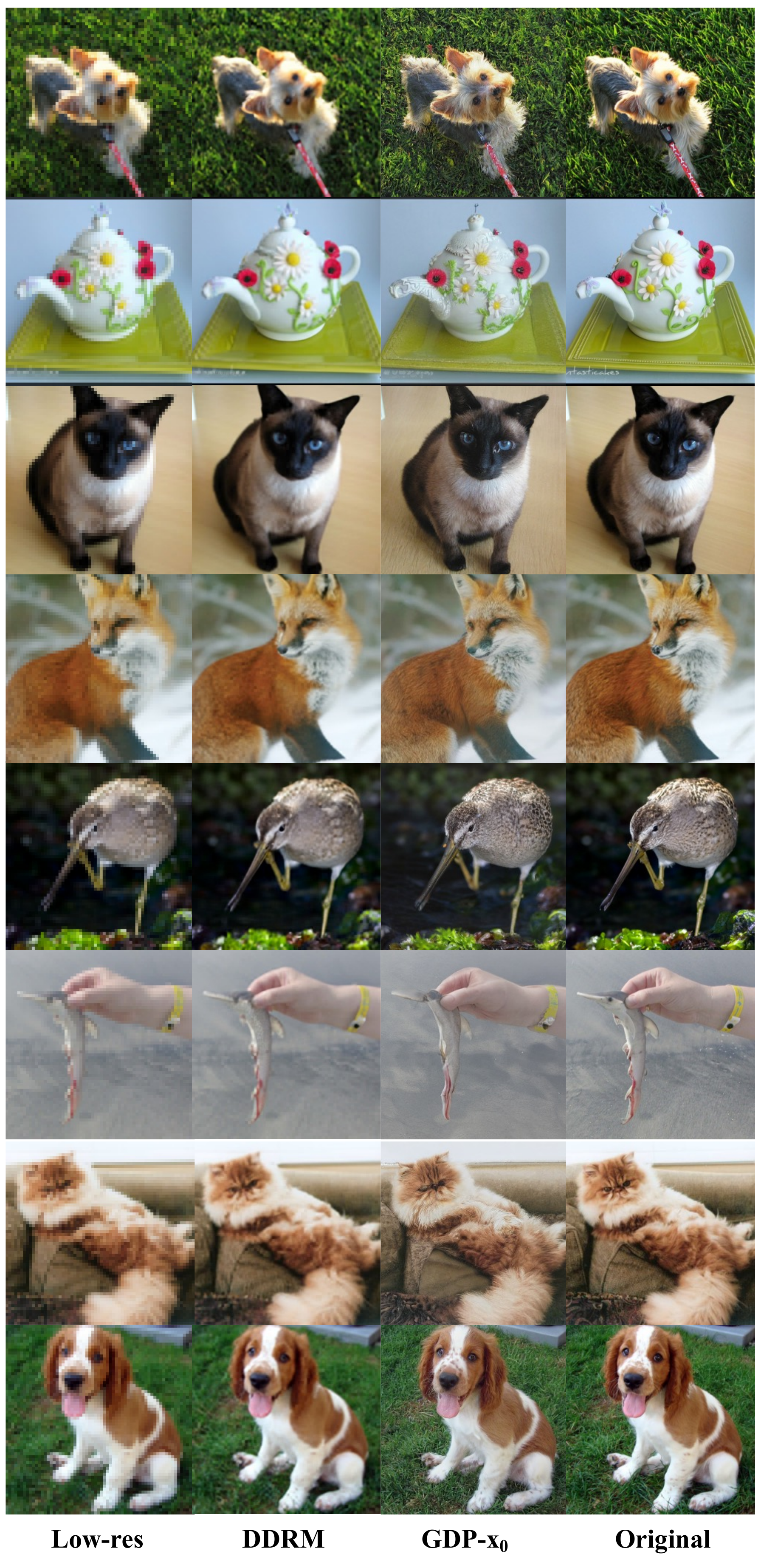}
   \caption{\textbf{More samples from the \underline{4 $\times$ super-resolution} task compare with DDRM on 256 $\times$ 256 \underline{ImageNet 1K.}} As we mentioned above, DDRM adds guidance on the $x_t$, leading to the less satisfactory results than our GDP-$x_0$.}
% \vspace{-0.5cm}
\label{SM:figure-sr4-imagenet-ddrm}
\end{figure*}

\begin{figure*}[htbp]
  \centering
   \includegraphics[width=34pc]{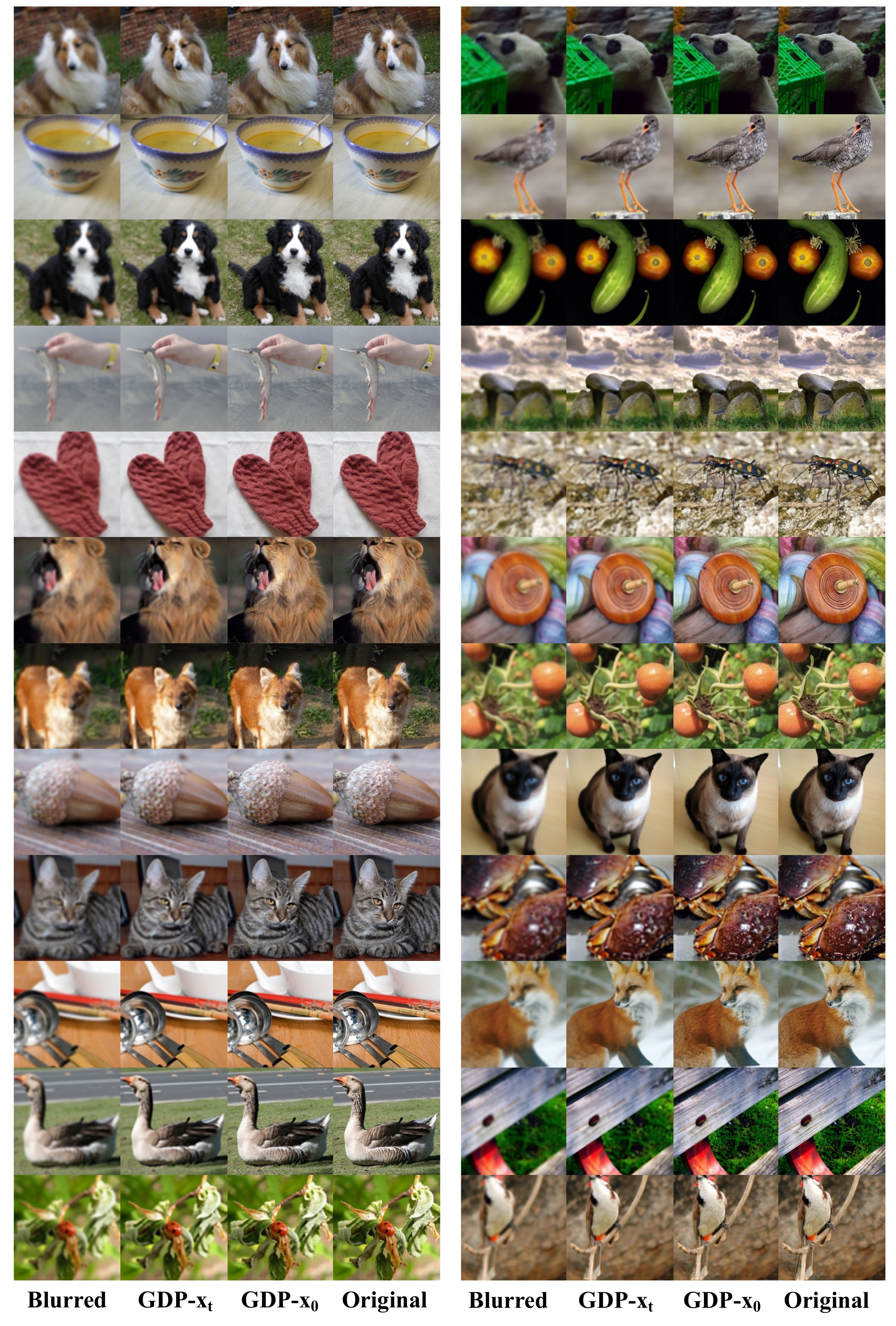}
   \caption{\textbf{Uncurated samples from the \underline{deblurring task} on 256 $\times$ 256 \underline{ImageNet 1K}.}}
% \vspace{-0.5cm}
\label{SM:figure-blur-imagenet}
\end{figure*}

\begin{figure*}[htbp]
  \centering
   \includegraphics[width=35pc]{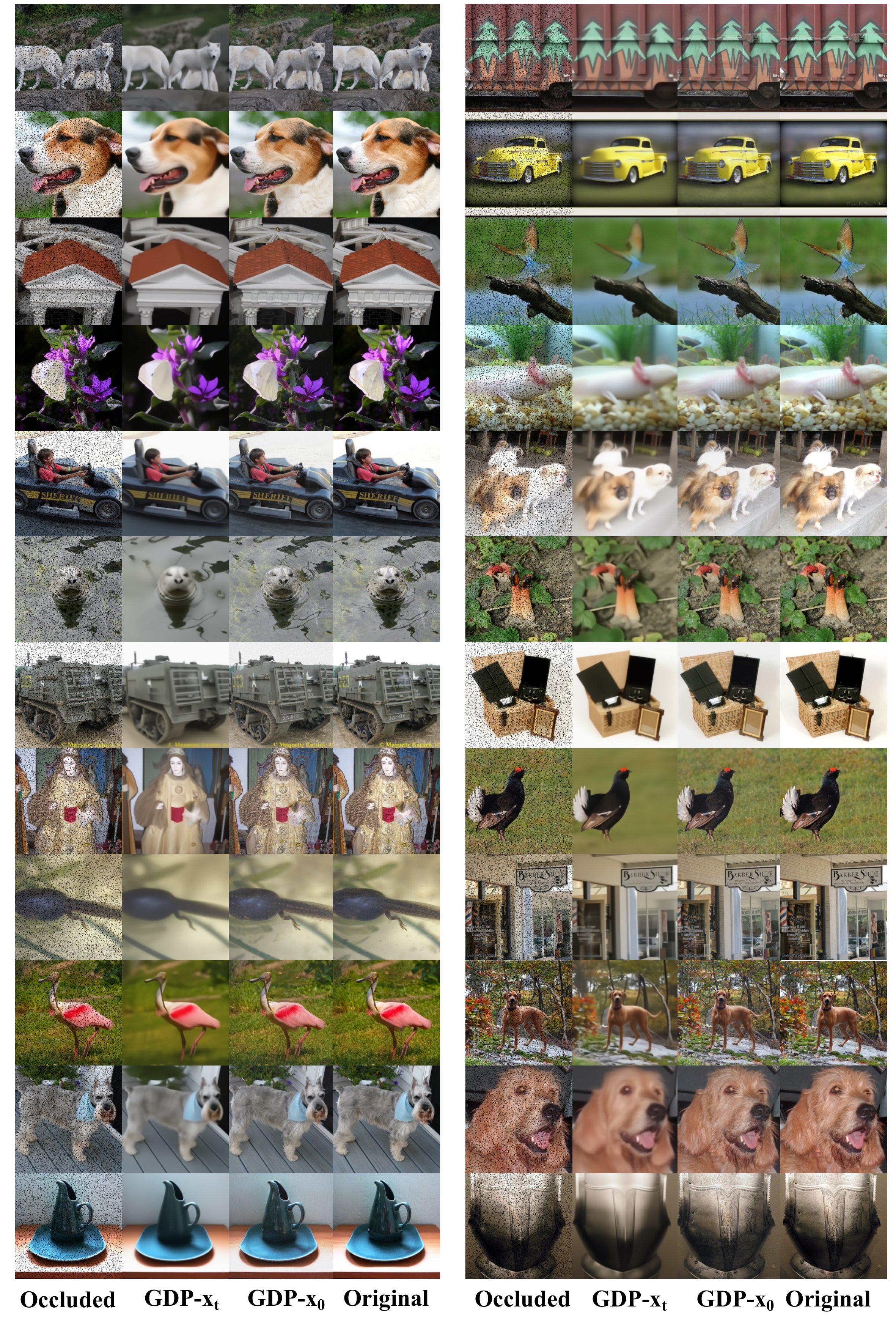}
   \caption{\textbf{Uncurated samples from the \underline{$10\%$ inpainting} task on 256 $\times$ 256 \underline{ImageNet 1K}.}}
% \vspace{-0.5cm}
\label{SM:figure-10inp-imagenet}
\end{figure*}

\begin{figure*}[htbp]
  \centering
   \includegraphics[width=35pc]{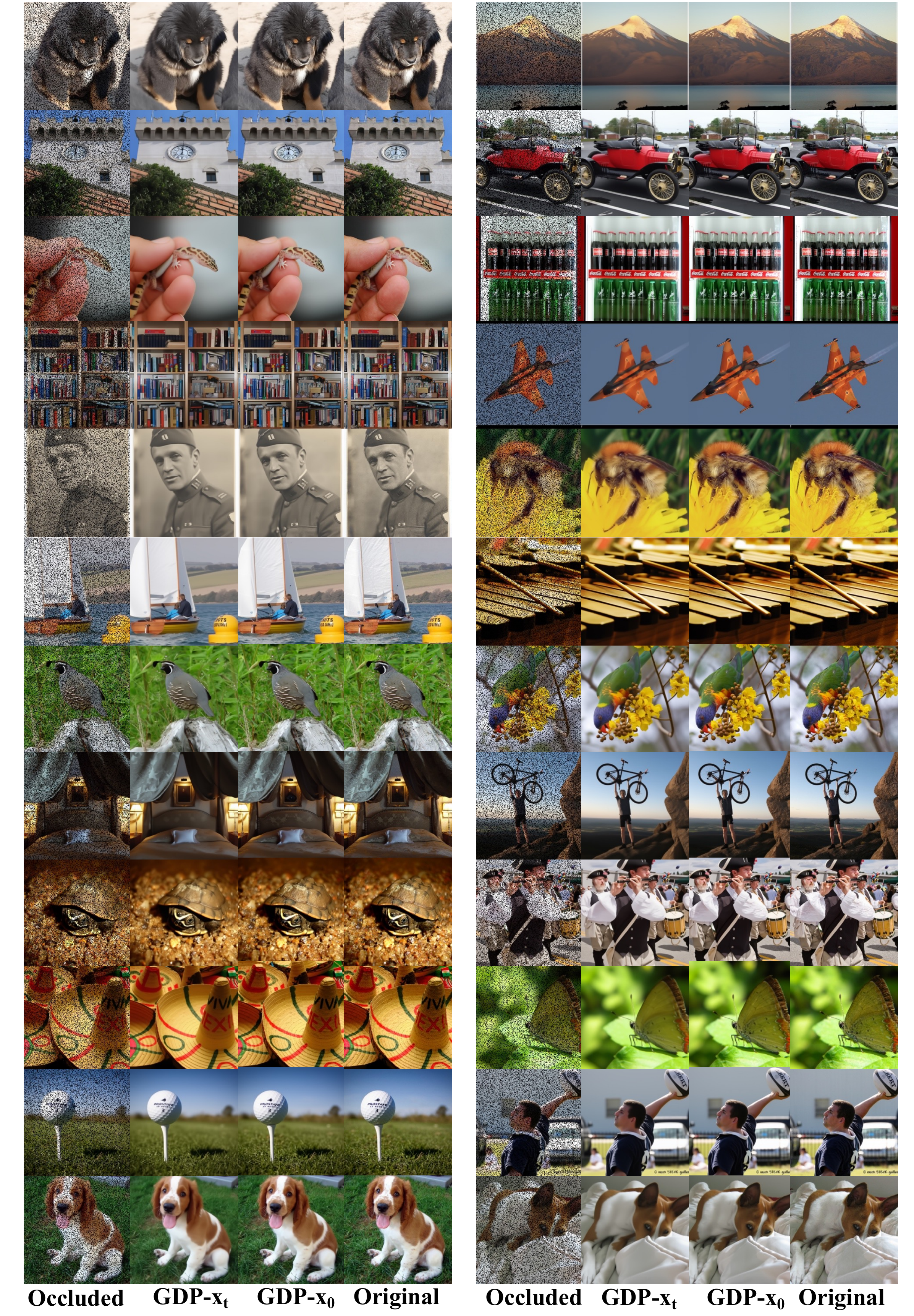}
   \caption{\textbf{Uncurated samples from the \underline{$25\%$ inpainting} task on 256 $\times$ 256 \underline{ImageNet 1K}.}}
% \vspace{-0.5cm}
\label{SM:figure-25inp-imagenet}
\end{figure*}

\begin{figure*}[htbp]
  \centering
   \includegraphics[width=35pc]{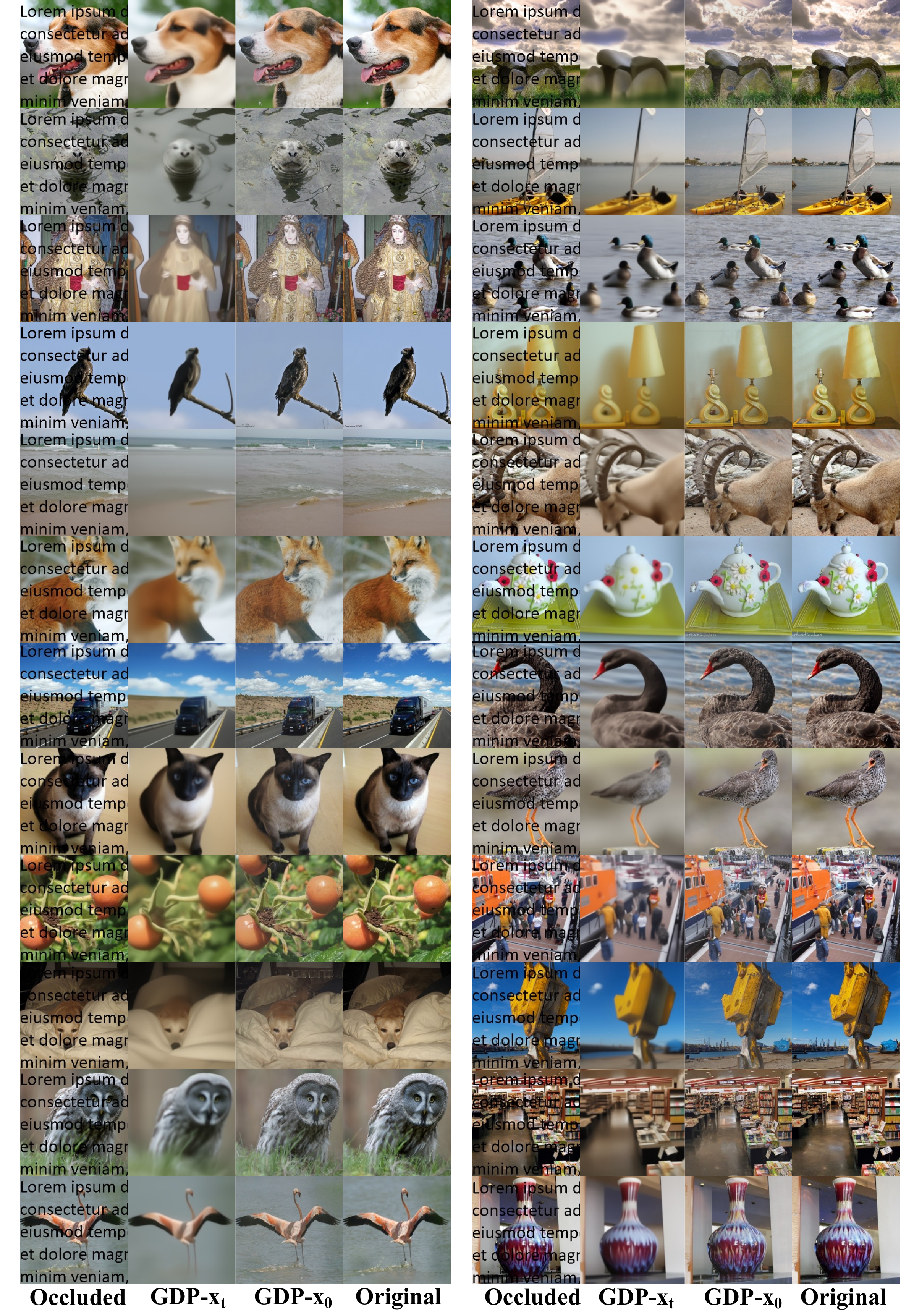}
   \caption{\textbf{Uncurated samples from the \underline{inpainting} task on 256 $\times$ 256 \underline{ImageNet 1K}.}}
% \vspace{-0.5cm}
\label{SM:figure-lorem-imagenet}
\end{figure*}

\begin{figure*}[htbp]
  \centering
   \includegraphics[width=35pc]{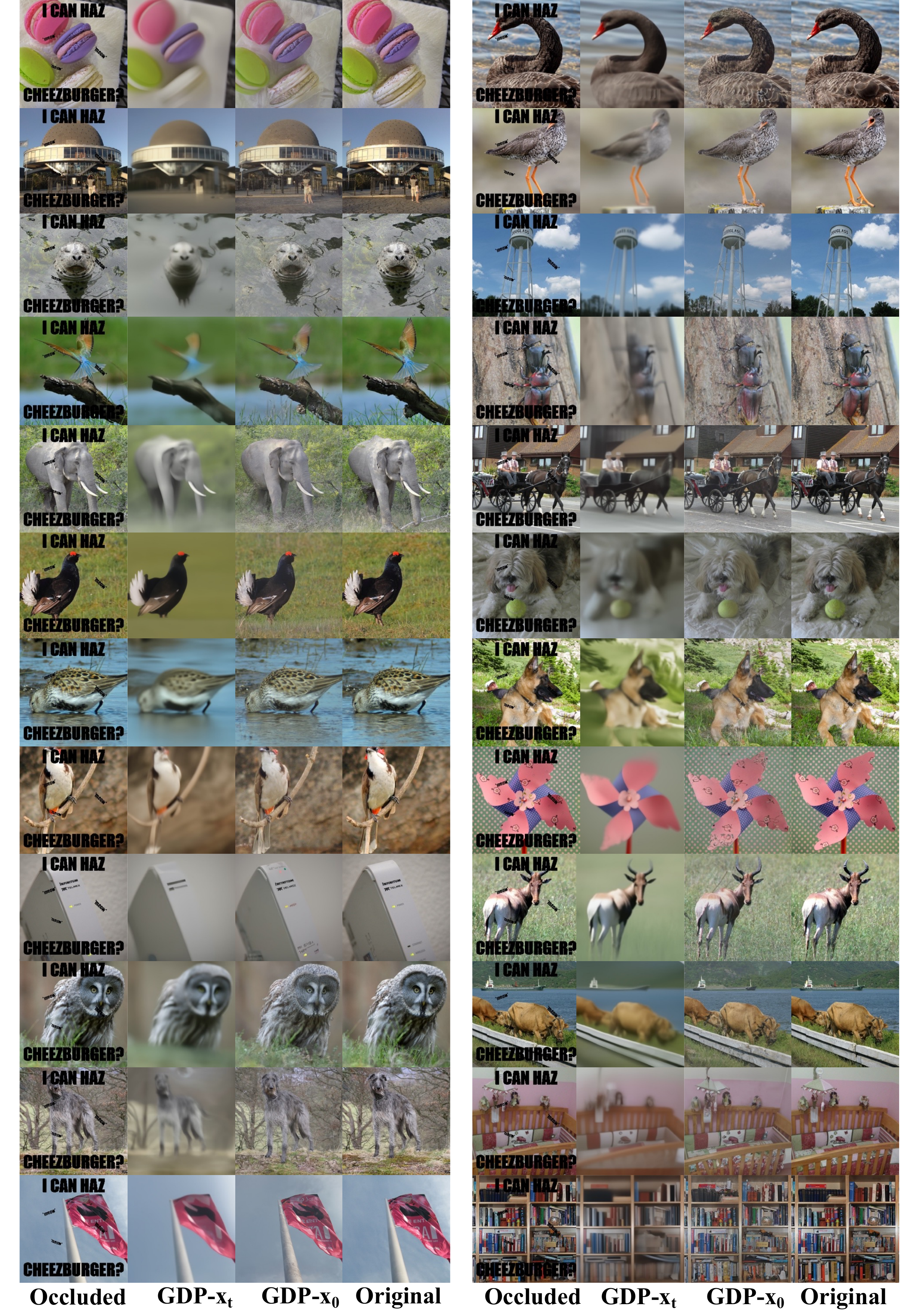}
   \caption{\textbf{Uncurated samples from the \underline{inpainting} task on 256 $\times$ 256 \underline{ImageNet 1K.}}}
% \vspace{-0.5cm}
\label{SM:figure-lolcat-imagenet}
\end{figure*}

\begin{figure*}[htbp]
  \centering
   \includegraphics[width=35pc]{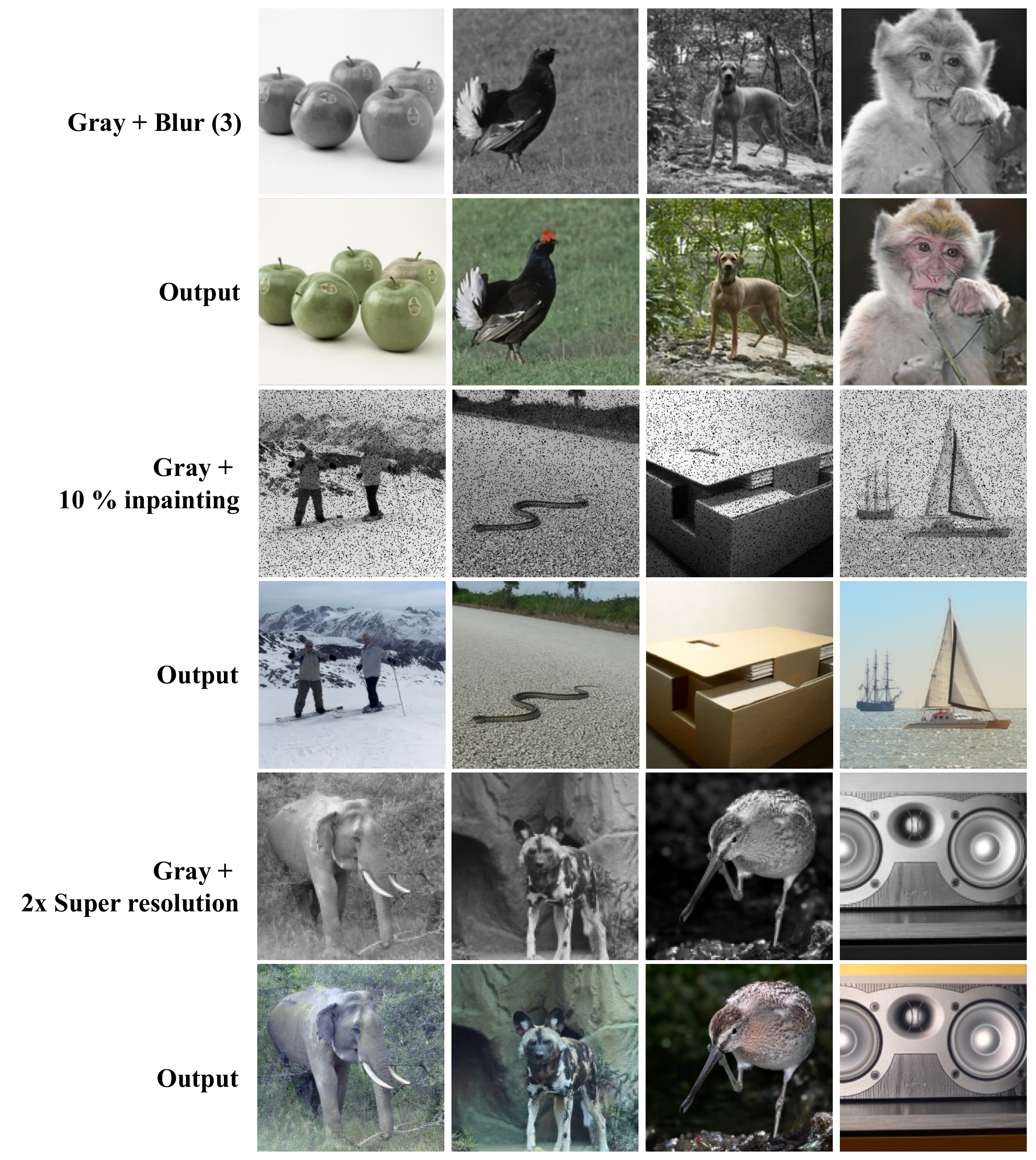}
   \caption{\textbf{Samples from the \underline{multi-degradation} tasks on 256 $\times$ 256 \underline{ImageNet 1K.}} It is shown that GDP can recover the corrupted images undergoing multiple degradations, such as gray + blur, gray + inpainting, and gray + down-sampling. It is noted that multi-linear degradation should be only one degradation model that will damage the contents of the images. In other words, the restoration will be more difficult if two content-damaged degradations occur at the same time, such as down-sampling + mask.}
% \vspace{-0.5cm}
\label{SM:figure-multidegra-imagenet}
\end{figure*}

We provide additional figures below showing GDP’s versatility across different datasets and linear inverse problems (Figures~\ref{SM:figure-sr4-celeba},~\ref{SM:figure-blur-bedroom},~\ref{SM:figure-3task-face},~\ref{SM:figure-4task-bedroom}), and~\ref{SM:figure-sr4-imagenet-ddrm}).  We present more uncurated samples from the ImageNet experiments in Figures~\ref{SM:figure-sr4-imagenet},~\ref{SM:figure-blur-imagenet},~\ref{SM:figure-10inp-imagenet},~\ref{SM:figure-25inp-imagenet},~\ref{SM:figure-lorem-imagenet}, and~\ref{SM:figure-lolcat-imagenet}. Moreover, our GDP is also able to recover the corrupted images that undergo multi-linear degradations, as shown in Fig.~\ref{SM:figure-multidegra-imagenet}
% Moreover, we further illustrate GDP’s advantage over previous unsupervised methods by evaluating two additional inverse problems: (i) 4$\times$ super-resolution with the popular bicubic downsampling kernel; and (ii) deblurring with a Gaussian blur kernel.

\section{Additional Results on Low-light Enhancement}

\begin{figure*}[htbp]
  \centering
   \includegraphics[width=\linewidth]{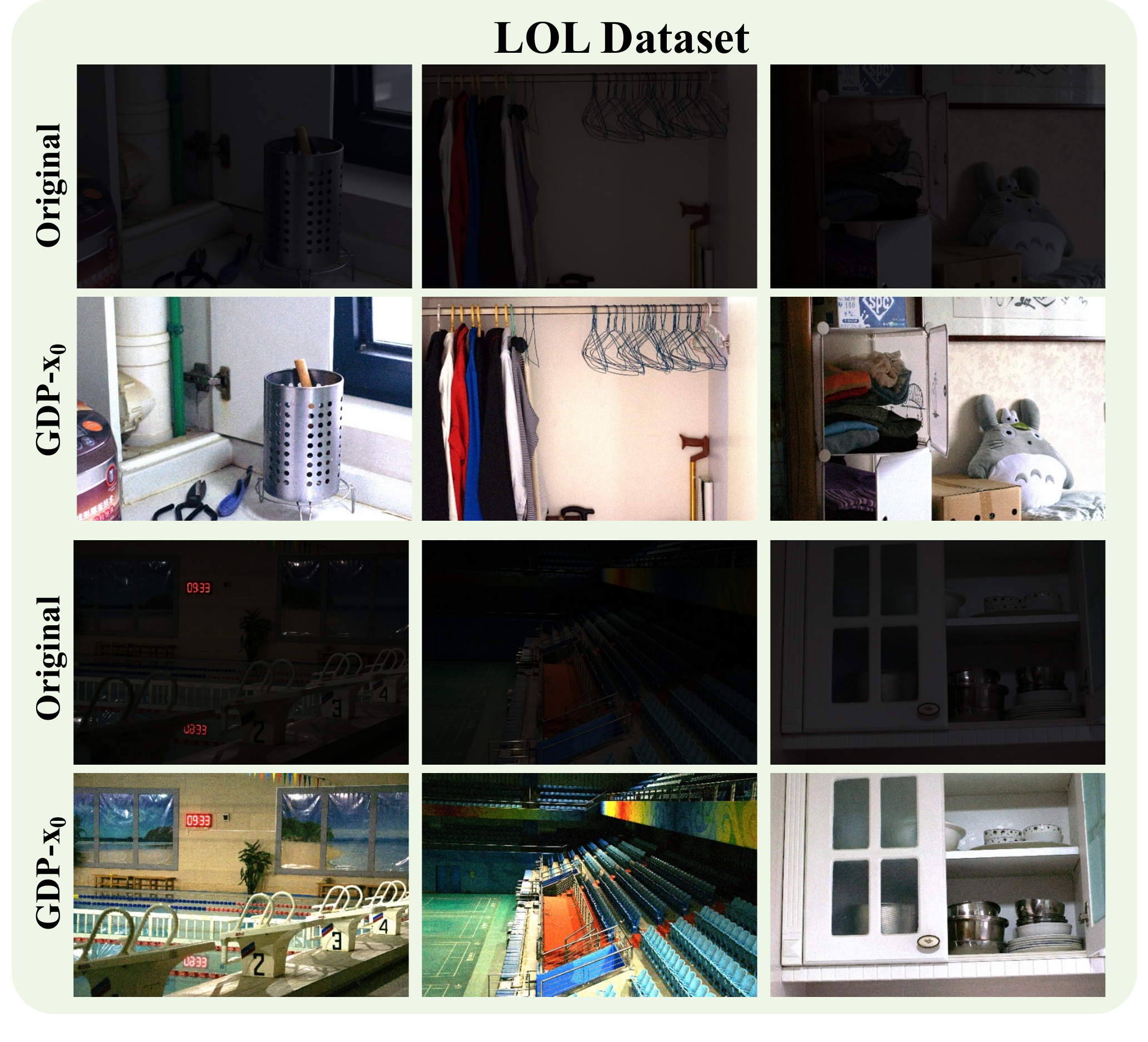}
   \caption{\textbf{Results of low-light image enhancement on \underline{LOL} dataset.}}
% \vspace{-0.5cm}
\label{SM:figure-lol}
\end{figure*}

\begin{figure*}[htbp]
  \centering
   \includegraphics[width=34pc]{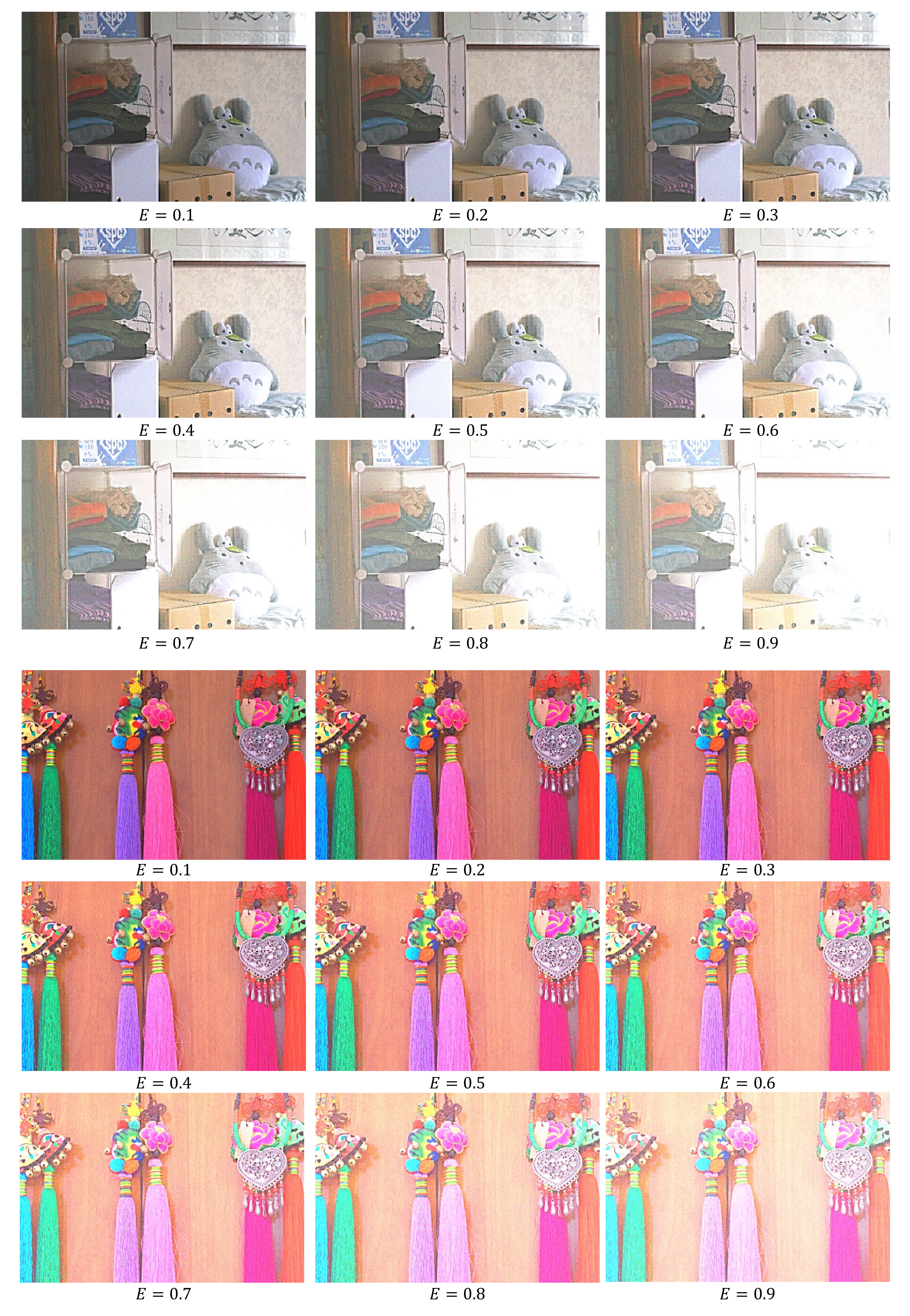}
   \caption{\textbf{Results of \underline{light control} on \underline{LOL} dataset.} We can adjust the brightness of the generated images with the help of Exposure Control Loss. Users can adjust the gray level $E$ in the RGB color space to obtain the images according to their needs.}
% \vspace{-0.5cm}
\label{SM:figure-lol-lightcontrol}
\end{figure*}

\begin{figure*}[htbp]
  \centering
   \includegraphics[width=\linewidth]{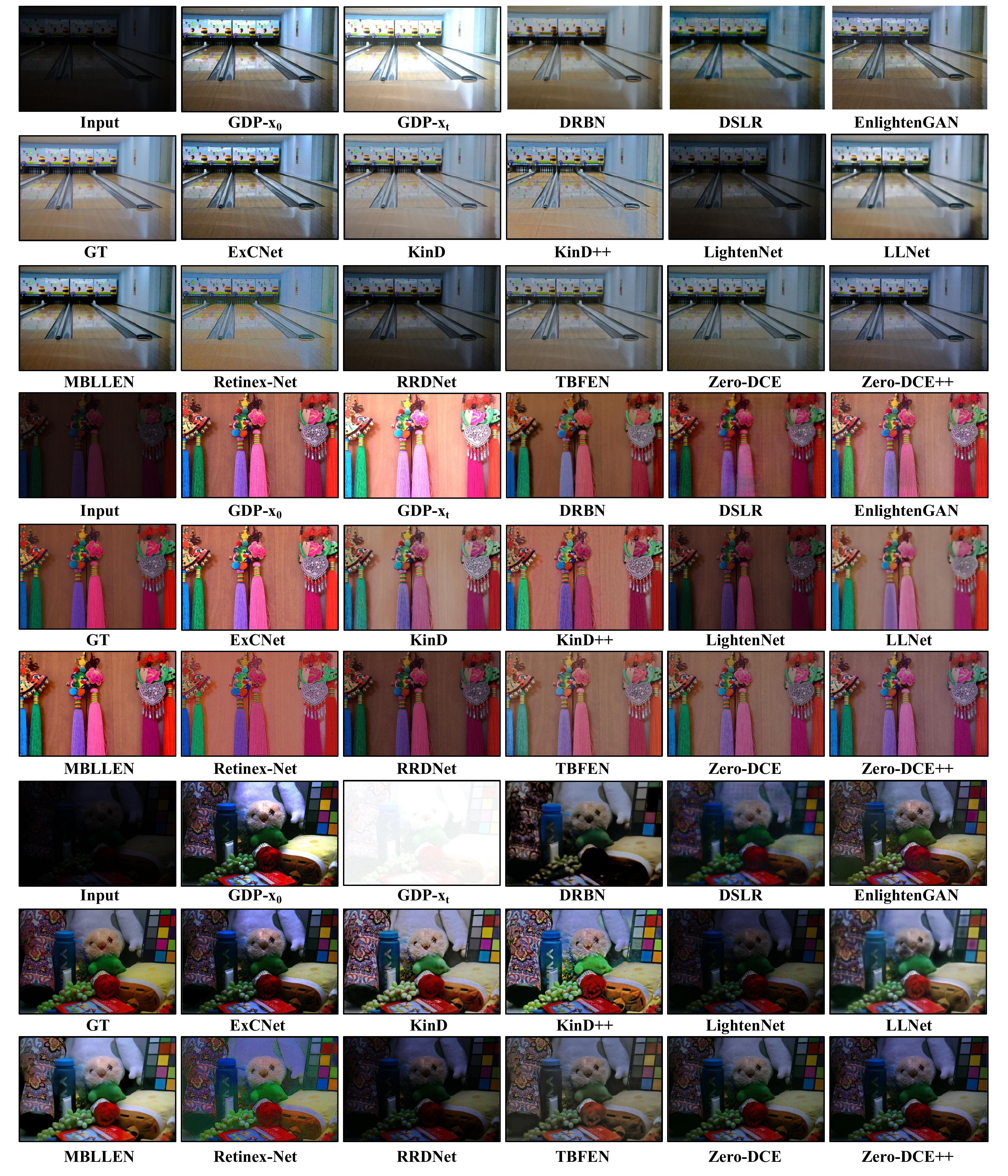}
   \caption{\textbf{The comparison of our GDP and other methods on the \underline{LOL} datasets towards low-light enhancement.}}
% \vspace{-0.5cm}
\label{SM:figure-lol-comparison}
\end{figure*}

\begin{figure*}[htbp]
  \centering
   \includegraphics[width=28pc]{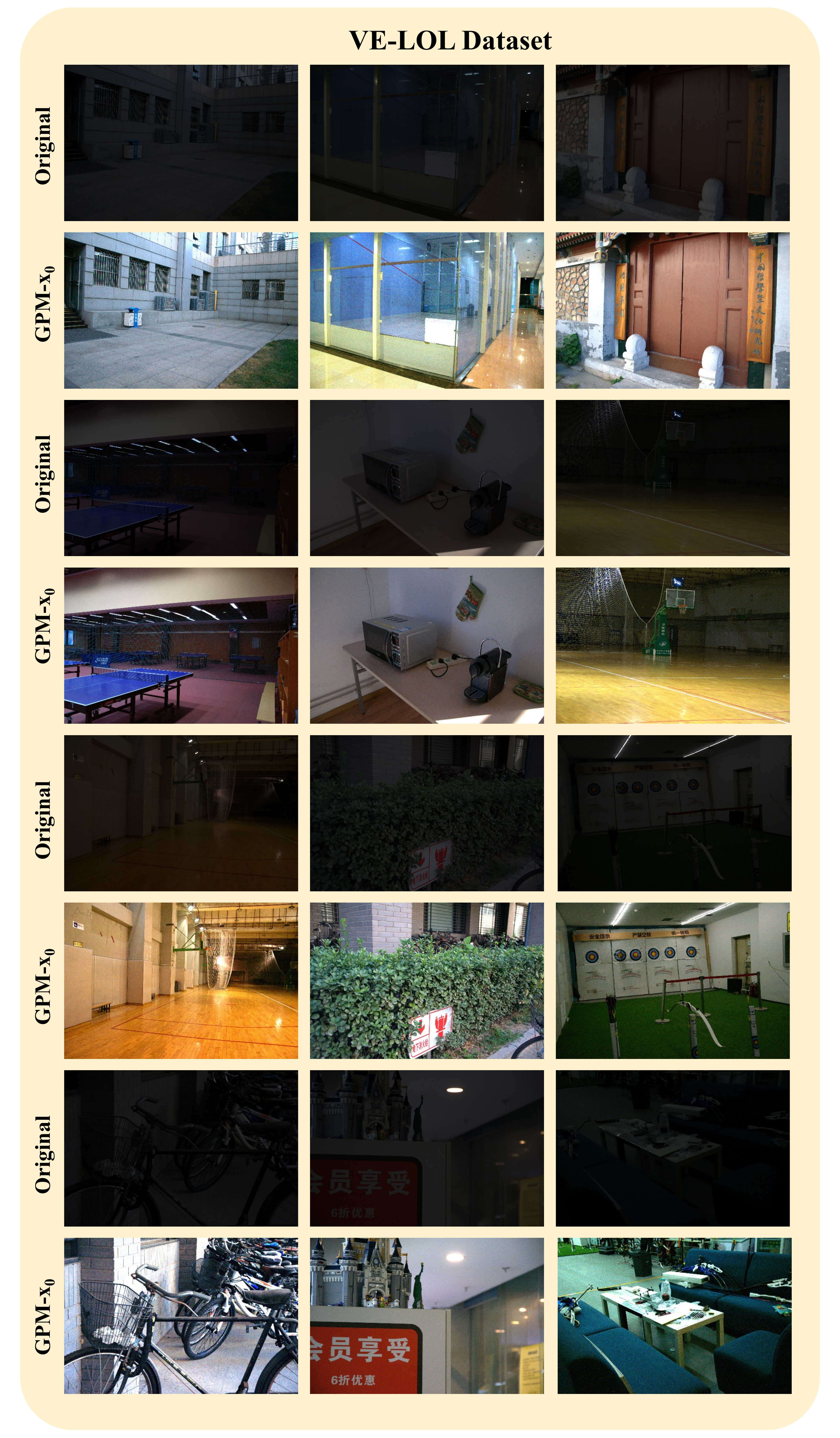}
   \caption{\textbf{Results of low-light image enhancement on \underline{VE-LOL-L} dataset.}}
% \vspace{-0.5cm}
\label{SM:figure-velol}
\end{figure*}

\begin{figure*}[htbp]
  \centering
   \includegraphics[width=\linewidth]{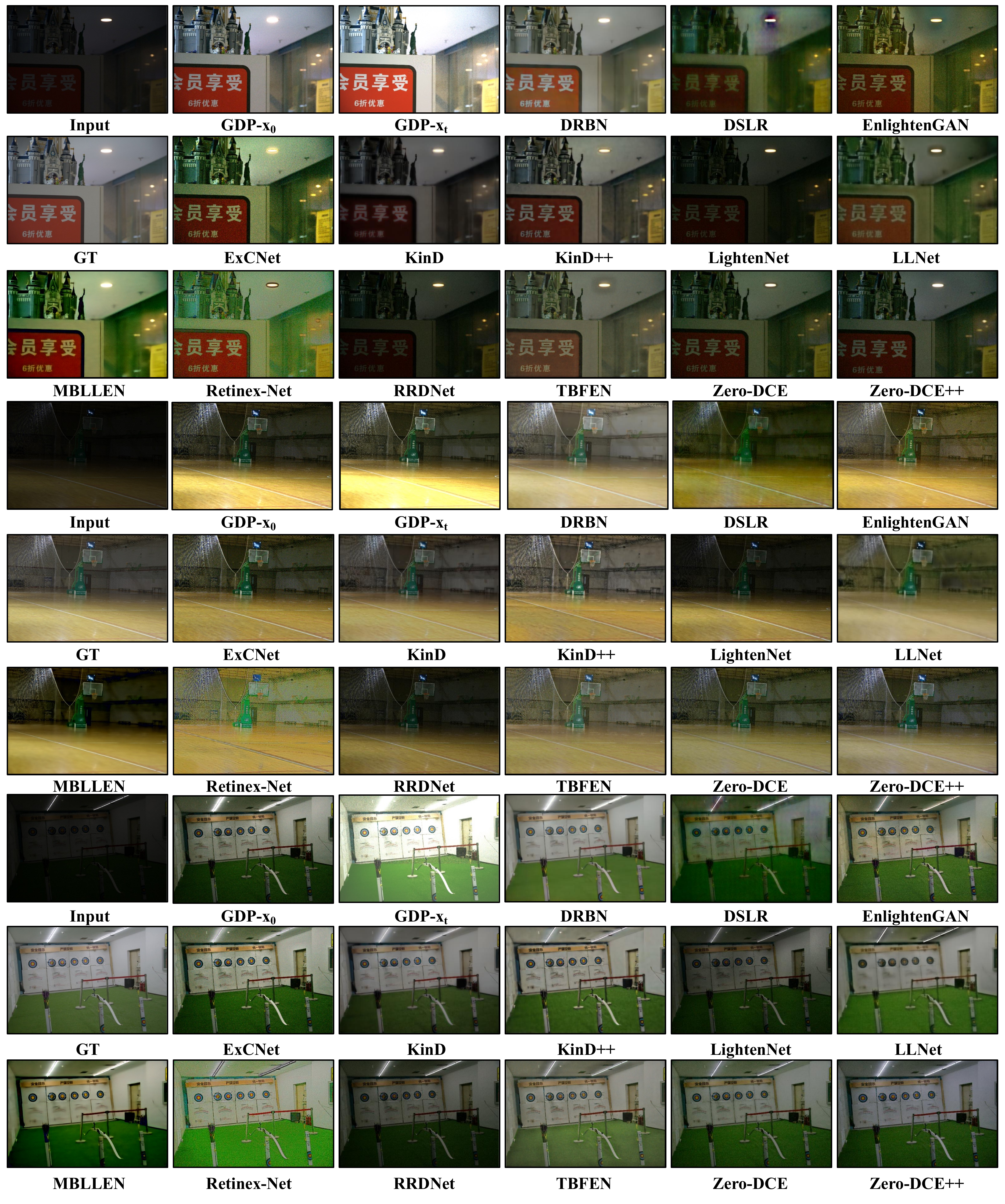}
   \caption{\textbf{The comparison of our GDP and other methods on the \underline{VE-LOL-L} datasets towards low-light enhancement.}}
% \vspace{-0.5cm}
\label{SM:figure-velol-comparison}
\end{figure*}

\begin{figure*}[htbp]
  \centering
   \includegraphics[width=0.7\linewidth]{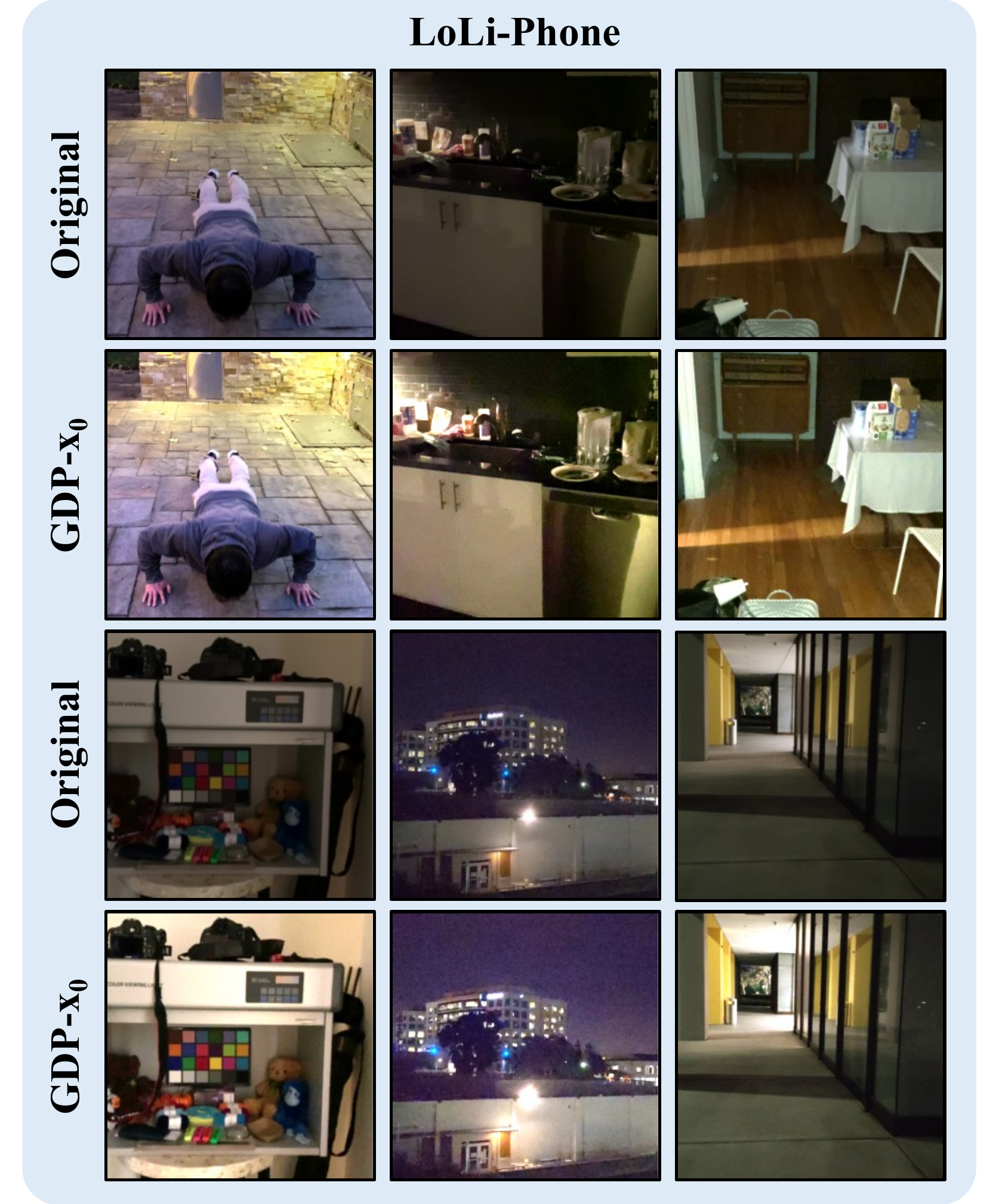}
   \caption{\textbf{Results of low-light image enhancement on \underline{LoLi-Phone} dataset.}}
\label{SM:figure-loliphone}
\end{figure*}

In addition to the linear inverse problems, we further show more samples on the blind and non-linear task of low-light enhancement. As shown in~\ref{SM:figure-lol},~\ref{SM:figure-velol}, and~\ref{SM:figure-loliphone}, our GDP performs well under the three datasets, including LOL, VE-LOL-L, and LoLi-phone, indicating the effectiveness of GDP under the different distributions of the images. 
Moreover, we also compare the GDP with other methods on the three datasets. As seen in~\ref{SM:figure-lol-comparison}, and~\ref{SM:figure-velol-comparison}, GDP-$x_0$ is able to generate more satisfactory images than other supervised learning, unsupervised learning, self-supervised, and zero-shot learning methods.
Note that GDP-$x_t$ tends to yield images lighter than the ones generated by GDP-$x_0$.
Furthermore, GDP can adjust the brightness of generated images by the Exposure Control Loss. As shown in~\ref{SM:figure-lol-lightcontrol}, users can change the gray level $E$ in the RGB color space to obtain the target images with specific brightness.

\begin{figure*}[htbp]
  \centering
   \includegraphics[width=\linewidth]{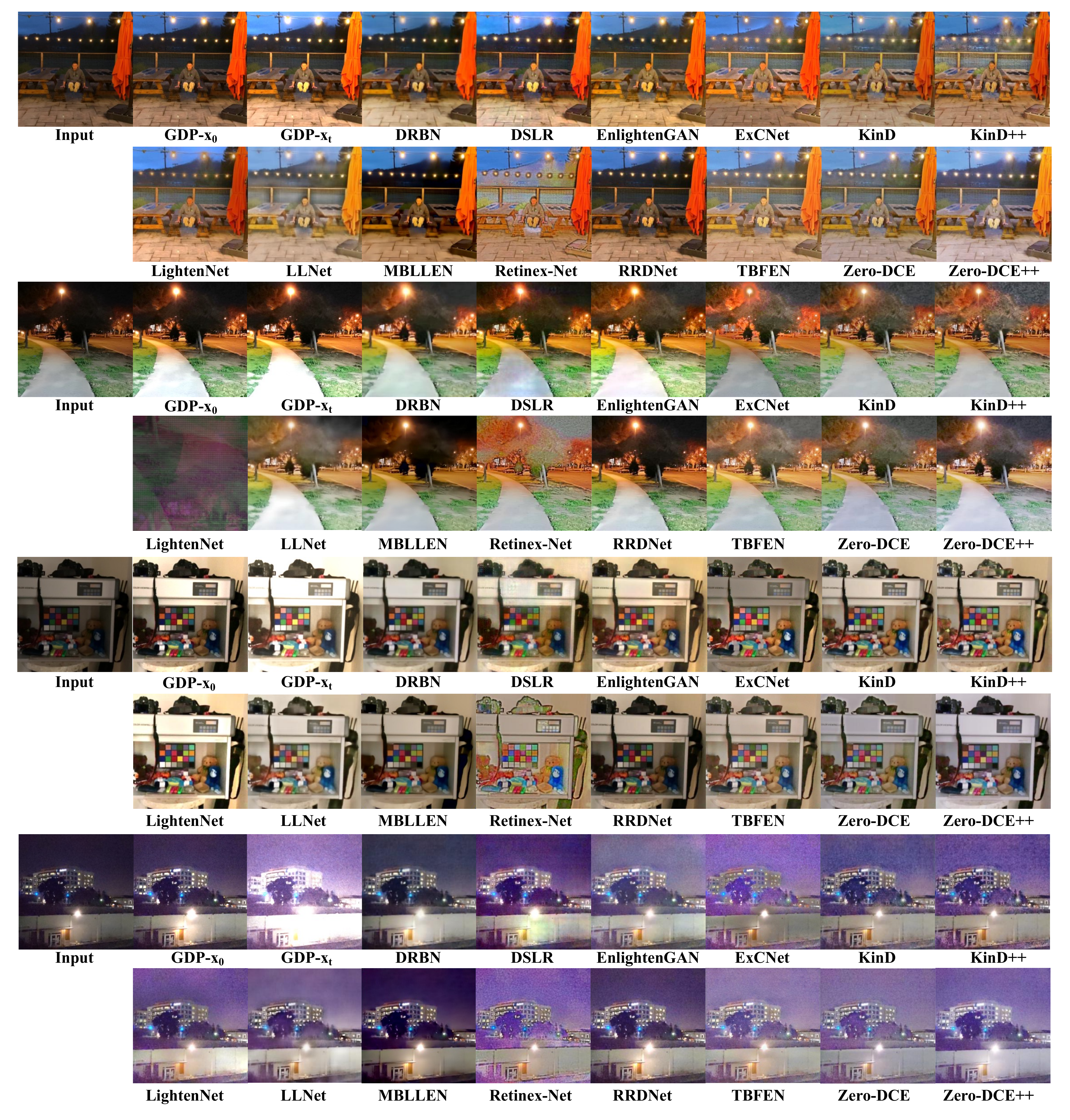}
   \caption{\textbf{The comparison of our GDP and other methods on the \underline{LoLi-phone} datasets towards low-light enhancement.}}
% \vspace{-0.5cm}
\label{SM:figure-loliphone-comparison}
\end{figure*}

\section{Additional Results on HDR Recovery}

\begin{figure*}[htbp]
  \centering
   \includegraphics[width=\linewidth]{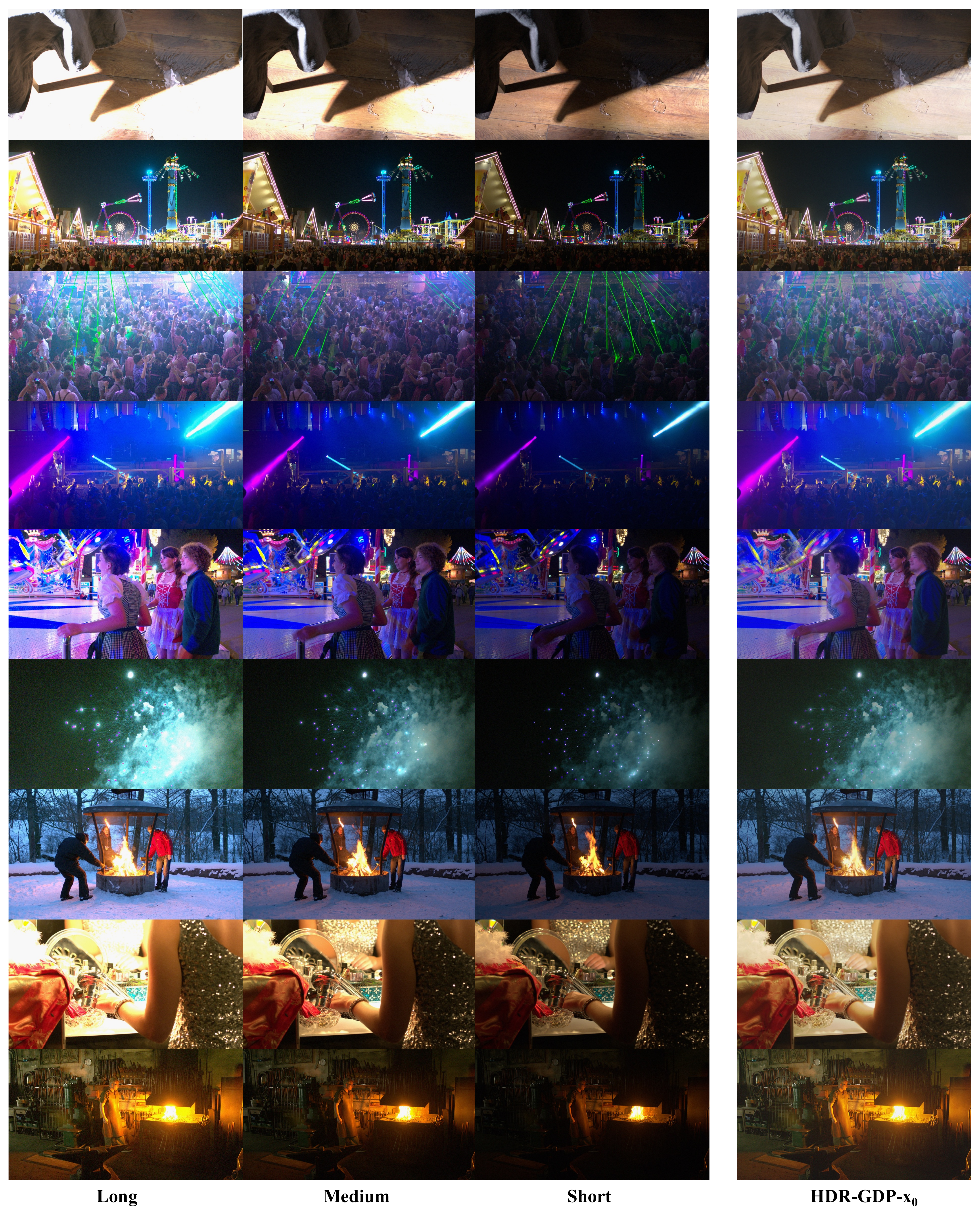}
   \caption{\textbf{Results of HDR image recovery on \underline{NTIRE2021} dataset.}}
\label{SM:figure-hdr}
\end{figure*}

\begin{figure*}[htbp]
  \centering
   \includegraphics[width=0.9\linewidth]{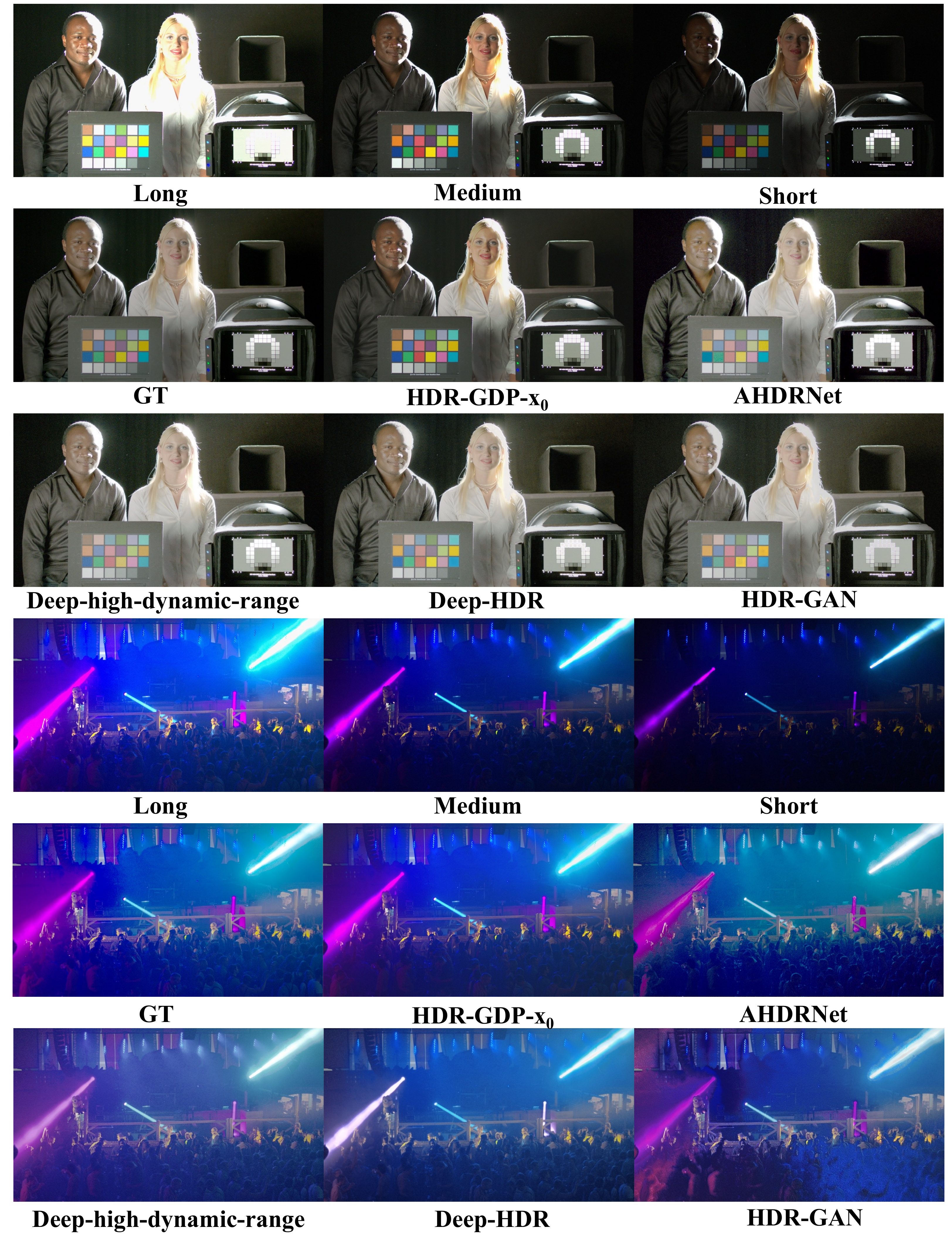}
   \caption{\textbf{The comparison of HDR image recovery on \underline{NTIRE2021} dataset.}}
\label{SM:figure-hdr-comparison}
\end{figure*}

As shown in~\ref{SM:figure-hdr}, our HDR-GDP-$x_0$ is capable of adjusting the over-exposed and under-exposed areas of the picture in various scenes. It is noted that since the model used by GDP is pre-trained on ImageNet, the tone of the generated picture will be slightly different from ground truth images.
Moreover, we also show more samples compared with the state-of-the-art methods, including AHDRNet~\cite{yan2019attention}, HDR-GAN~\cite{niu2021hdr}, DeepHDR~\cite{wu2018deep} and deep-high-dynamic-range~\cite{kalantari2017deep}. As seen in Fig.~\ref{SM:figure-hdr-comparison}, our HDR-GDP-$x_0$ can recover more realistic images with more details.

% \clearpage
% \newpage

\section{Additional Results on Ablation Study}

\begin{figure*}[htbp]
  \centering
   \includegraphics[width=\linewidth]{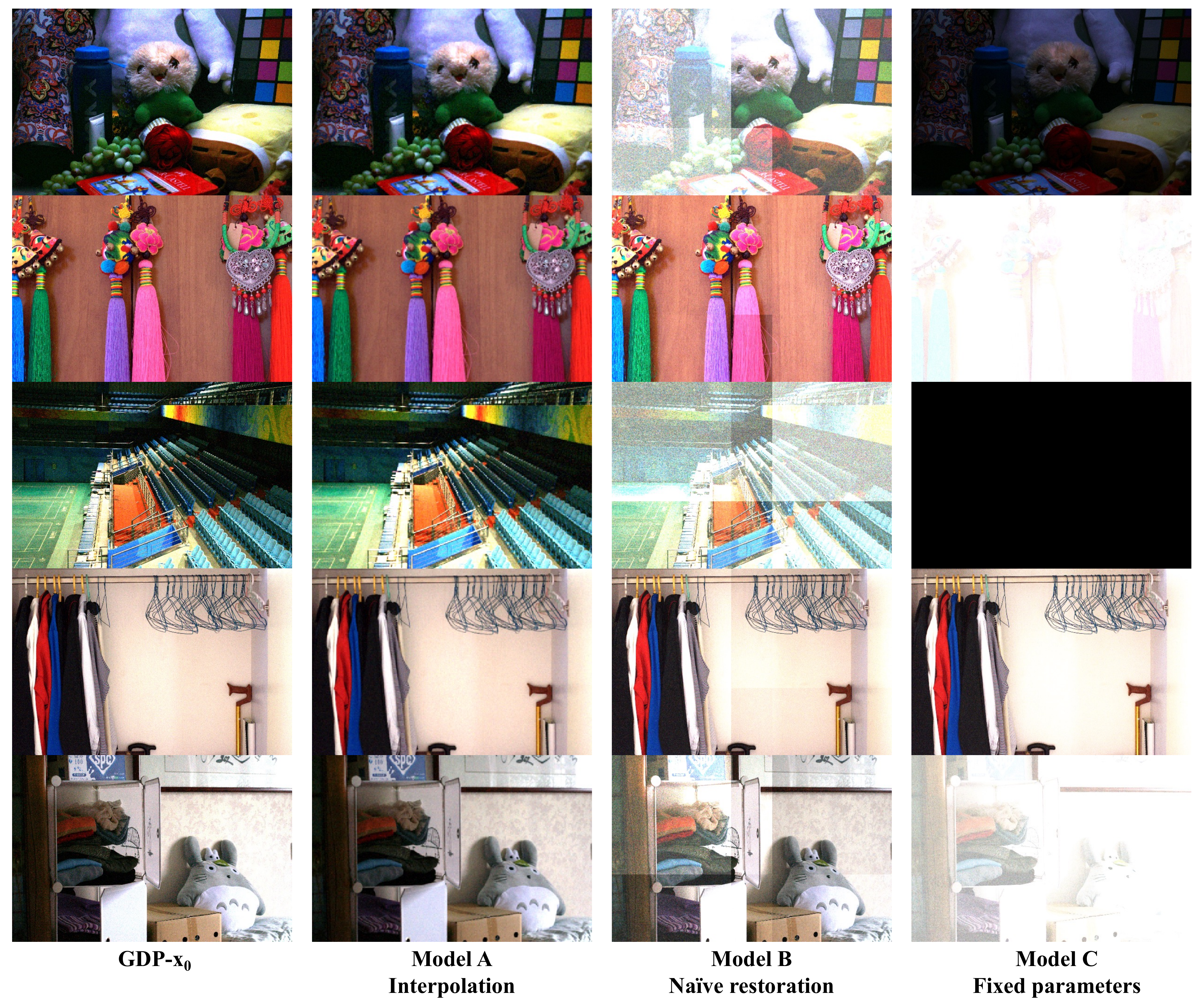}
   \caption{\textbf{Qualitative comparison of ablation study on LOL dataset.} Model A recovers the images in $256 \times N$ or $256 \times N$ sizes and is interpolated by the nearest neighbor to the original size. Model B is devised to naively restore the images from patches and patches where the parameters are not related. Model C is designed with fixed parameters for all patches in the images.}
\label{SM:ablation-low-light}
\end{figure*}

\begin{figure*}[htbp]
  \centering
   \includegraphics[width=\linewidth]{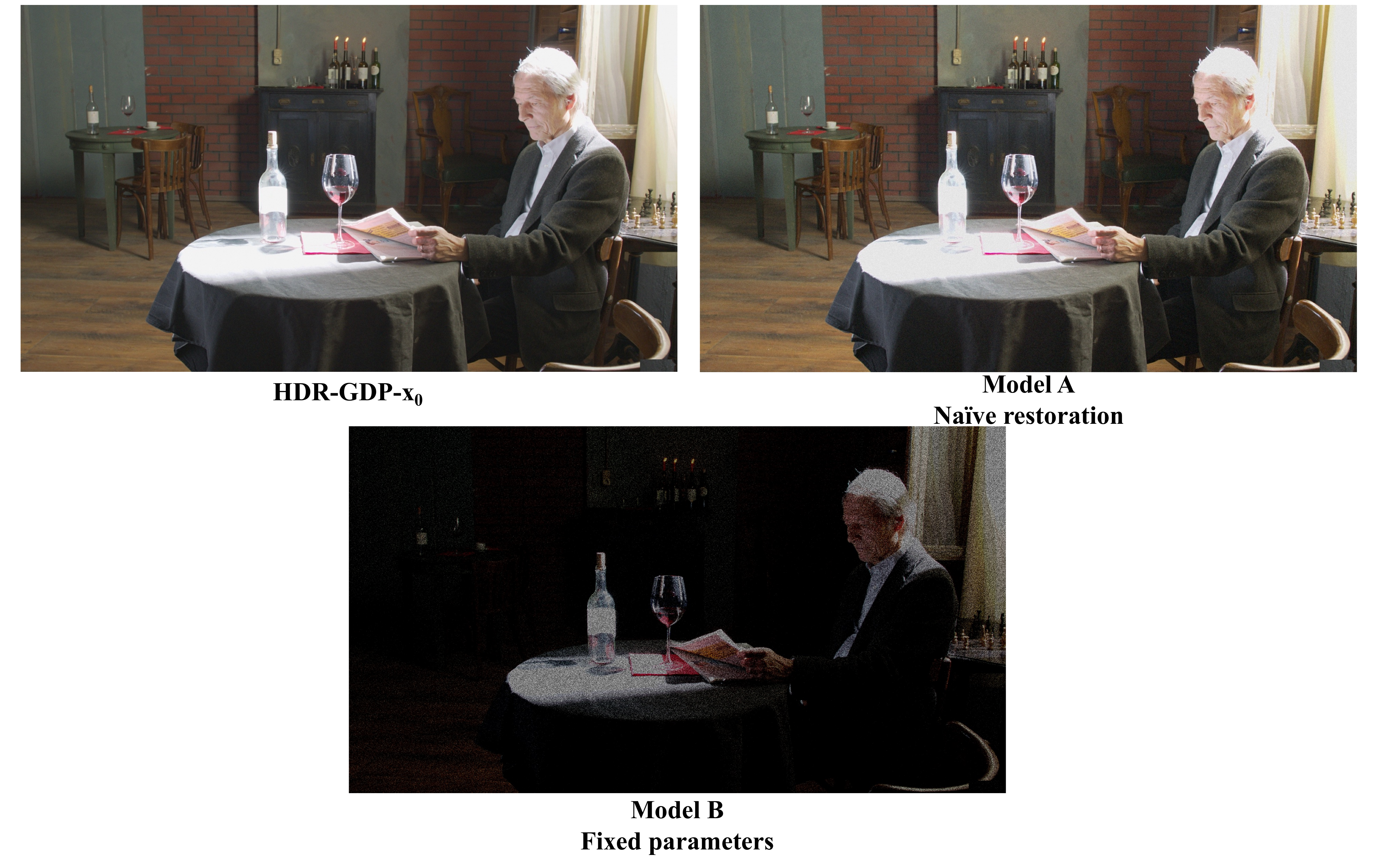}
   \caption{\textbf{Qualitative comparison of ablation study on \underline{NTIRE2021} dataset.}}
\label{SM:ablation-hdr}
\end{figure*}

The visualization comparisons of the ablation study on the trainable degradation and the patch-based tactic are shown in Figs.~\ref{SM:ablation-low-light} and~\ref{SM:ablation-hdr}. It is shown that Model A fails to generate high-quality images due to the interpolation operation, while Model B generates images with more artifacts because of the naive restoration. Model C predicts the outputs in an uncontrollable way thanks to the randomly initiated and fixed parameters.

% \clearpage
% \newpage

%%%%%%%%% REFERENCES
% \clearpage
{
% \small
% \clearpage
% \newpage
% \vfill
% \bibliographystyle{ieee_fullname}
% \bibliography{egbib}
% }

\end{document}